\documentclass[10pt,twocolumn,letterpaper]{article}

\usepackage{cvpr}
\usepackage{multirow}
\usepackage{makecell}
\usepackage{array}

\usepackage[dvipsnames]{xcolor}

\definecolor{cvprblue}{rgb}{0.21,0.49,0.74}
\usepackage[pagebackref,breaklinks,colorlinks,citecolor=cvprblue]{hyperref}

\def\paperID{7051} 
\def\confName{CVPR}
\def\confYear{2024}

\title{Gaussian Shading: Provable Performance-Lossless Image Watermarking for Diffusion Models}

\author{
Zijin Yang\textsuperscript{\rm 1},
Kai Zeng\textsuperscript{\rm 1},
Kejiang Chen\textsuperscript{\rm 1,}\thanks{Corresponding author},
Han Fang\textsuperscript{\rm 2},
Weiming Zhang\textsuperscript{\rm 1},
Nenghai Yu\textsuperscript{\rm 1},\\
\textsuperscript{\rm 1}Anhui Province Key Laboratory of Digital Security, University of Science and Technology of China \\
\textsuperscript{\rm 2}National University of Singapore  \\
{\tt\small \{bsmhmmlf@mail., zk0128@mail., chenkj@, zhangwm@, ynh@\}ustc.edu.cn}\quad
{\tt\small fanghan@nus.edu.sg}
}

\begin{document}
\maketitle
\begin{abstract}

Ethical concerns surrounding copyright protection and inappropriate content generation pose challenges for the practical implementation of diffusion models. One effective solution involves watermarking the generated images. However, existing methods often compromise the model performance or require additional training, which is undesirable for operators and users. To address this issue, we propose Gaussian Shading, a diffusion model watermarking technique that is both performance-lossless and training-free, while serving the dual purpose of copyright protection and tracing of offending content. Our watermark embedding is free of model parameter modifications and thus is plug-and-play. We map the watermark to latent representations following a standard Gaussian distribution, which is indistinguishable from latent representations obtained from the non-watermarked diffusion model. Therefore we can achieve watermark embedding with lossless performance, for which we also provide theoretical proof. Furthermore, since the watermark is intricately linked with image semantics, it exhibits resilience to lossy processing and erasure attempts. The watermark can be extracted by Denoising Diffusion Implicit Models (DDIM) inversion and inverse sampling. We evaluate Gaussian Shading on multiple versions of Stable Diffusion, and the results demonstrate that Gaussian Shading not only is performance-lossless but also outperforms existing methods in terms of robustness.

\end{abstract}    

\section{Introduction}
\label{sec:intro}

Diffusion models~\cite{sohl2015deep,song2019generative,song2020score,ho2020denoising,song2020denoising} signify a noteworthy leap forward in image generation. These well-trained diffusion models, especially commercial diffusion models like Stable Diffusion (SD)~\cite{rombach2022high}, Glide~\cite{nichol2021glide}, and Muse AI~\cite{rombach2022high}, enable individuals with diverse backgrounds to create high-quality images effortlessly. However, this raises concerns about intellectual property and whether diffusion models will be stolen or resold twice. 

\begin{figure}[t]
  \centering
\includegraphics[width=0.89\linewidth]{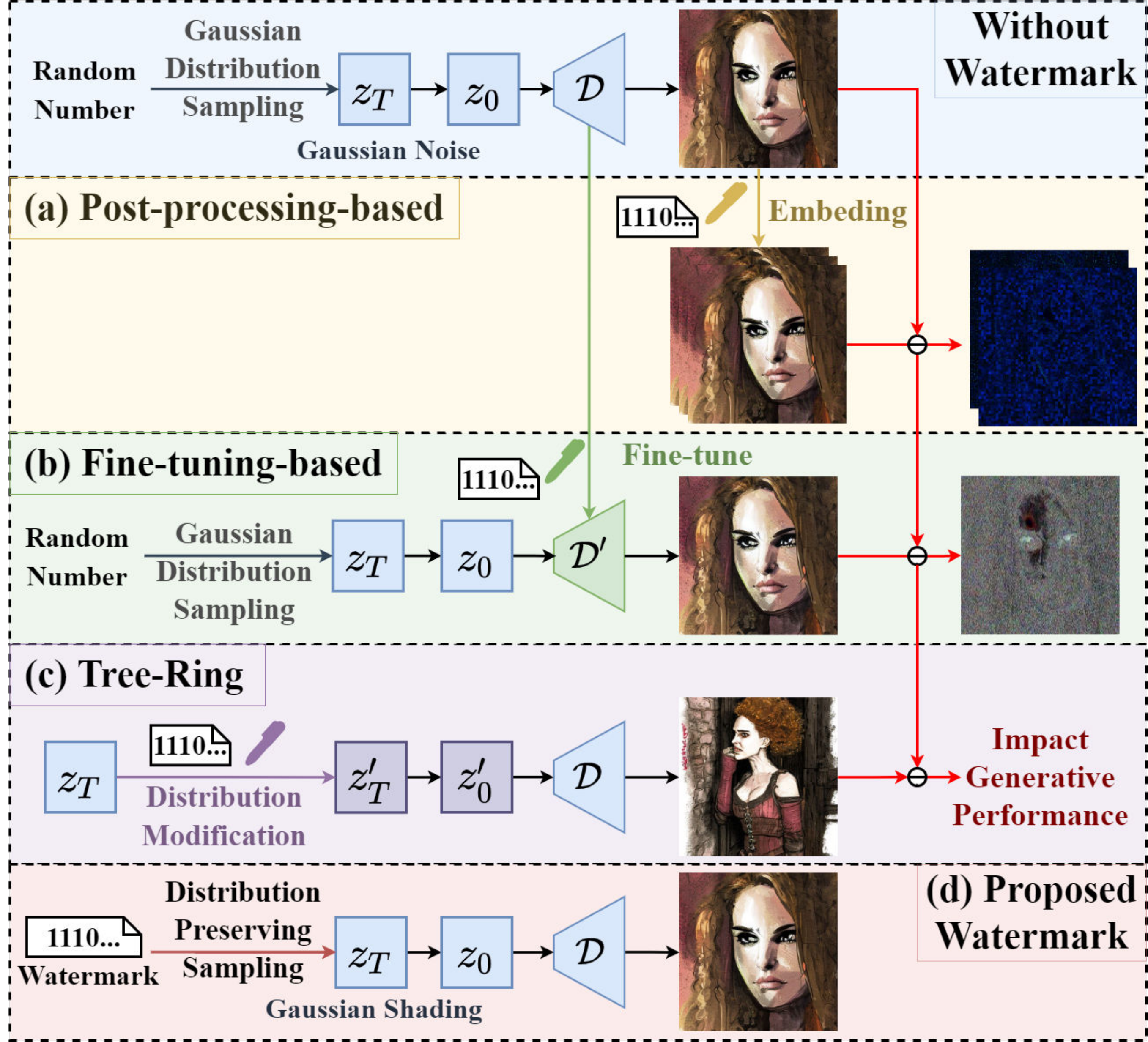}
\vspace{-0.3cm}
   \caption{Existing watermarking frameworks can be divided into three categories: post-processing-based, fine-tuning-based, and latent-representation-based Tree-Ring. Our method also relies on latent representations but achieves performance-lossless without altering the distribution.}
   \label{fig:watermark}
   \vspace{-0.6cm}
\end{figure}

On the other hand, the ease of generating realistic images raises concerns about potentially misleading content generation. For example, on May 23, 2023, a Twitter-verified user named Bloomberg Feed posted a tweet titled ``Large explosion near the Pentagon complex in Washington DC-initial report," along with a synthetic image. This tweet led to multiple authoritative media accounts sharing it, even causing a brief impact on the stock market\footnote{\href{https://www.cnn.com/2023/05/22/tech/twitter-fake-image-pentagon-explosion/index.html}{Fake image of Pentagon explosion on Twitter}}. On October 30, 2023, White House issued an executive order on AI security, emphasizing the need to protect Americans from AI-enabled fraud and deception by establishing standards and best practices for detecting AI-generated content and authenticating official content\footnote{\href{https://www.whitehouse.gov/briefing-room/statements-releases/2023/10/30/fact-sheet-president-biden-issues-executive-order-on-safe-secure-and-trustworthy-artificial-intelligence/}{FACT SHEET: President Biden Issues Executive Order on Safe, Secure, and Trustworthy Artificial Intelligence}}. The urgency of labeling generated content for copyright authentication and prevention of misuse is evident.

Watermarking is highlighted as a fundamental method for labeling generated content, as it embeds watermark information within the generated image, allowing for subsequent copyright authentication and the tracking of false content. Existing watermarking methods for the diffusion model can be divided into three categories, as shown in \cref{fig:watermark}. Post-processing-based watermarks~\cite{cox2007digital,zhang2019robust} adjust robust image features to embed watermarks, thereby directly altering the image and degrading its quality. 
To mitigate this concern, recent research endeavors propose fine-tuning-based methods~\cite{fernandez2023stable,zhao2023recipe,liu2023watermarking,cui2023diffusionshield,xiong2023flexible}, which amalgamate the watermark embedding process with the image generation process. Intuitively, these methods need to modify model parameters, introducing supplementary computational overhead. Recently, Wen et al.~\cite{wen2023tree} proposed the latent-representation-based Tree-Ring watermark, which conveys information by adapting the latent representations to match specific patterns. However, it restricts the randomness of sampling,  which impacts generative performance.

Through the above analysis, we can find that these methods compromise model performance to embed watermarks. In practical applications, model performance is paramount for both business interests and user experience. Substantial resource investment is often necessary to pursue enhanced model performance. This leads to a fundamental question: Can watermarks be embedded without compromising model performance?

We affirmatively address the question presented above. Succinctly, the generation process can be delineated into two key phases: latent representation sampling and decoding. Our goal is to align the distribution of the latent representation in watermarked images with that of the latent representation in normally generated images. By keeping the model unaltered, the distribution of watermarked images is naturally consistent with that of normally generated images, enabling the seamless embedding of watermarks without compromising model performance.

Building upon this insight, we propose a watermarking method named Gaussian Shading, designed to ensure no deterioration in model performance. The embedding process encompasses three primary elements: watermark diffuse, randomization, and distribution-preserving sampling. Watermark diffusion spreads the watermark information throughout the latent representation. During the generation process, the watermark information will be diffused to the whole semantics of the image, thus achieving excellent robustness. Watermark randomization and distribution-preserving sampling guarantee the congruity of the latent representation distribution with that of watermark-free latent representations, thereby achieving performance-lossless. In the extraction phase, the latent representations are acquired through Denoising Diffusion Implicit Model (DDIM) inversion~\cite{song2020denoising}, allowing for the retrieval of watermark information. Harnessing the extensive scope of the SD latent space, we can achieve a high-capacity watermark of 256 bits, surpassing prior methods.

To the best of our knowledge, ours is the first technique that tackles this challenging problem of performance-lossless watermarking for diffusion models, and we provide theoretical proof. Moreover, this technique leaves the architecture and parameters of SD unaltered, necessitating no supplementary training. It can seamlessly integrate as a plug-and-play module within the generation process. Model providers can easily replace watermarked models with non-watermarked ones without affecting usability experiences.

We conducted thorough experiments on SD. Under strong noise perturbation, the average true positive rate and bit accuracy can exceed 0.99 and 0.97, respectively, validating the superiority of Gaussian Shading in both detection and traceability tasks compared to prior methods. Additionally, experiments on visual quality and image-text similarity serve as indicators of performance preservation in our approach. Lastly, we deliberated on various watermark erasure attacks, affirming the steadfast performance of our watermark in the face of such adversities.

\section{Related Work}

\begin{figure*}[t]
  \centering
   \includegraphics[width=0.75\linewidth]{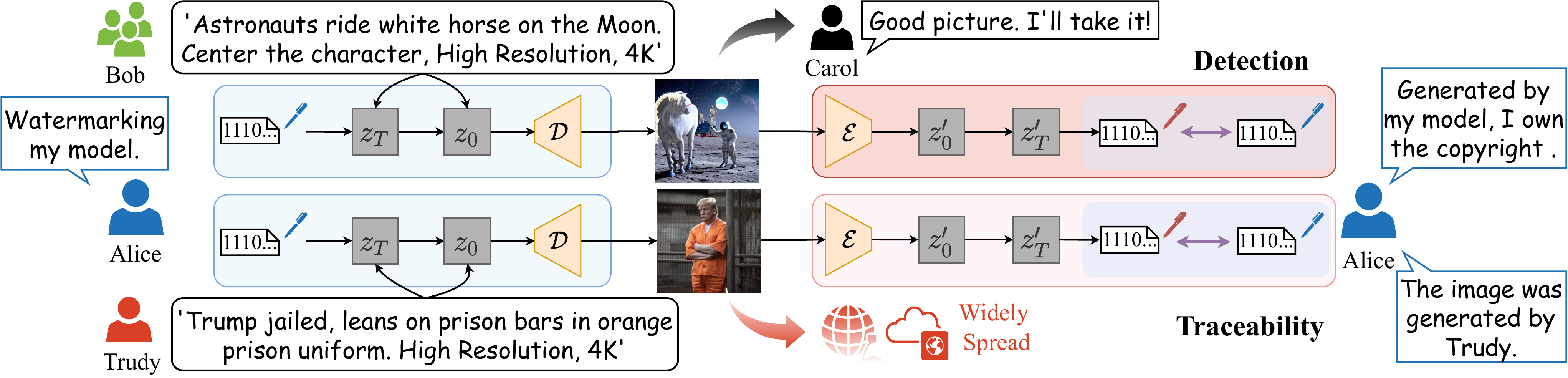}
   \vspace{-0.3cm}
   \caption{Application scenarios for Gaussian Shading.}
   \label{fig:sec}
   \vspace{-0.6cm}
\end{figure*}

\subsection{Diffusion Models}

Inspired by non-equilibrium thermodynamics, Ho et al.~\cite{ho2020denoising} introduced the Denoising Diffusion Probabilistic Model (DDPM). DDPM consists of two Markov chains used for adding and removing noise, and subsequent works~\cite{song2020denoising,dhariwal2021diffusion, nichol2021improved,lu2022dpm,gu2022vector,ho2022classifier,rombach2022high} have adopted this bidirectional chain framework. To reduce computational complexity and improve efficiency, the Latent Diffusion Model (LDM)~\cite{rombach2022high} was designed, in which the diffusion process occurs in a latent space $\mathcal{Z}$. To map an image $x \in \mathbb{R}^{H\times W\times 3}$ to the latent space, the LDM employs an encoder $\mathcal{E}$, such that $z_0 = \mathcal{E}(x) \in \mathbb{R}^{h\times w\times c}$. Similarly, to reconstruct an image from the latent space, a decoder $\mathcal{D}$ is used, such that $x = \mathcal{D}(z_0)$. A pretrained LDM can generate images without the encoder $\mathcal{E}$. Specifically, a latent representation $z_T$ is first sampled from a standard Gaussian distribution $ \mathcal{N}(0,I)$. Subsequently, through iterative denoising using methods like DDIM~\cite{song2020denoising}, $z_0$ is obtained, and an image can be generated using the decoder: $x = \mathcal{D}(z_0)$.

\subsection{Image Watermarking}

Digital watermarking~\cite{van1994digital} is an effective means to address copyright protection and content authentication by embedding copyright or traceable identification information within carrier data. Typically, the functionality of a watermark depends on its capacity. For example, a single-bit watermark can determine whether an image was generated by a particular diffusion model, i.e., copyright protection; a multi-bit watermark can further determine which user of the diffusion model generated the image, i.e., traceability.

Image watermarking is a method that employs images as carriers for watermarking. Initially, watermark embedding methods primarily focused on the spatial domain~\cite{van1994digital}, but later, to enhance robustness, transform domain watermarking techniques~\cite{kundur1997robust,tsai2000joint,guo2002digital,lee2007reversible,al2007combined,stankovic2009application,hamidi2018hybrid} were developed. In recent years, with the advancement of deep learning, researchers have turned their attention to neural networks~\cite{lecun1998gradient,vaswani2017attention}, harnessing their powerful learning capabilities to develop watermarking techniques~\cite{zhu2018hidden,luo2020distortion,zhang2020udh,kishore2021fixed, tancik2020stegastamp,zhong2020automated,jia2021mbrs}.

\subsection{Image Watermarking for Diffusion Models}

Existing Image watermarking methods for the diffusion model~\cite{cox2007digital,zhang2019robust,fernandez2023stable,zhao2023recipe,liu2023watermarking,cui2023diffusionshield,xiong2023flexible,wen2023tree} can be divided into three categories, as shown in \cref{fig:watermark}. The image watermarking methods described in the previous section can be applied directly to the images generated by the diffusion model, which is called post-processing-based watermarks~\cite{cox2007digital,zhang2019robust}. These methods directly modify the image, thus degrading image quality. Recent research endeavors have amalgamated the watermark embedding process with the image generation process to mitigate this issue. Stable Signature~\cite{fernandez2023stable} fine-tunes the LDM decoder using a pre-trained watermark extractor, facilitating watermark extraction from images produced by the fine-tuned model. Zhao et al.~\cite{zhao2023recipe} and Liu et al.~\cite{liu2023watermarking} suggest fine-tuning the diffusion model to implant a backdoor as a watermark, enabling watermark extraction by triggering. These fine-tuning-based approaches enhance the quality of watermarked images but introduce supplementary computational overhead and modify model parameters. Furthermore, Wen et al.~\cite{wen2023tree} introduced the Tree-Ring Watermark, which conveys copyright information by adapting the frequency domain of latent representations to match specific patterns. This method achieves an imperceptible watermark. However, it directly disrupts the Gaussian distribution of noise, limiting the randomness of sampling and resulting in affecting model performance.

\section{Methods}

\begin{figure*}[t]
  \centering
   \includegraphics[width=.8\linewidth]{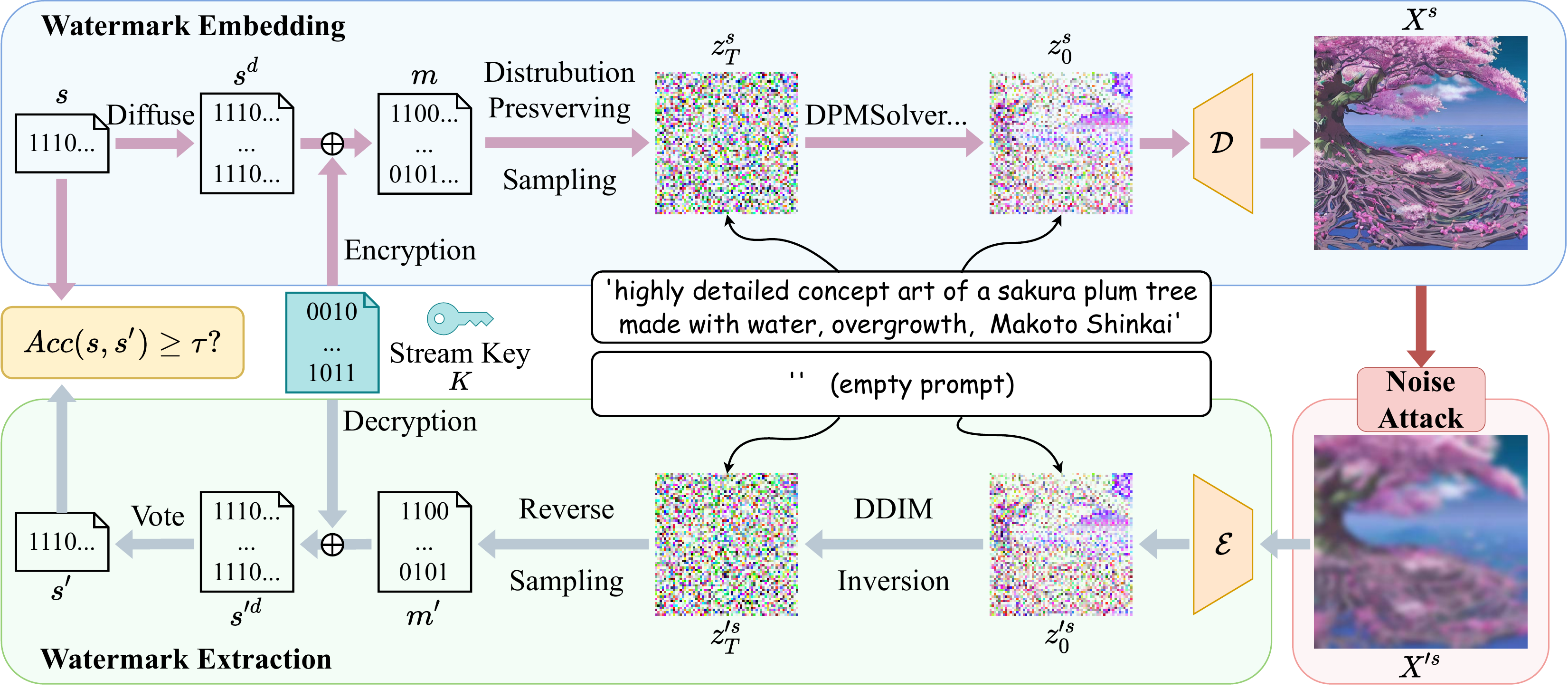}
   \vspace{-0.3cm}
   \caption{The framework of  Gaussian Shading. We utilize a $k$-bit binary sequence $s$ to represent the watermark. After diffusion and encryption, the watermark can be utilized to drive distribution-preserving sampling, followed by denoising to generate watermarked images $X^s$. For extraction, it is sufficient to introduce DDIM inversion and the inverse process of all the operations mentioned above.}
   \label{fig:framework}
   \vspace{-0.6cm}
\end{figure*}

In this section, we provide an overview of the application scenarios and functionalities in \cref{fig:sec}. We then proceed to detail the embedding and extraction processes shown in \cref{fig:framework}. Finally, we present a mathematical proof of the performance-lossless characteristic of the watermark.

\subsection{Application Scenarios}
\noindent
\textbf{Scenarios.}
See \cref{fig:sec}, the scenario involves the operator Alice, the thief Carol, and two types of users Bob and Trudy.

Alice is responsible for training the model, deploying it on the platform, and providing the corresponding API for users, but she does not open-source the code or model weights. Carol does not use Alice's services but steals images generated by her model, claiming ownership of the copyrights. Bob and Trudy, as community users, can utilize the API to generate and disseminate images. While Bob faithfully adheres to the community guidelines, Trudy aims to generate deep fake, and infringing content. To evade detection and traceability, Trudy can employ various data augmentation to modify illicit images.

\noindent
\textbf{Detection.} This scenario satisfies the detection (copyright protection) requirement. Alice embeds a single-bit watermark into each generated image. The successful extraction of the watermark from an image serves as evidence of Alice's rightful ownership of the copyright, while also indicating that the image is artificially generated (as opposed to natural images).

\noindent
\textbf{Traceability.} This scenario fulfills the traceability requirement.
Alice allocates a watermark to each user. By extracting the watermark from the illicit content, it enables tracing Trudy, through comparison with the watermark database.
Traceability is a higher pursuit than detection and can also achieve copyright protection for different users.

Details of the statistical tests in both scenarios are shown in Supplementary Material.

\subsection{Watermark Embedding}

\noindent
 \textbf{Watermark diffusion.} The dimensions of the latent representations are given by ${c\times h\times w}$, where each dimension can represent $l$ bits of the watermark. 
Therefore, the watermark capacity becomes ${l\times c\times h\times w}$ bits. To enhance the robustness of the watermark, we represent the watermark using $\frac{1}{f_{hw}}$ of the height and width, and $\frac{1}{f_c}$ of the channel, and replicate the watermark $f_c \cdot f^2_{hw}$ times. Thus, the watermark $s$ with dimensions $l\times \frac{c}{f_{c}}\times \frac{h}{f_{hw}}\times \frac{w}{f_{hw}}$  is expanded into a diffused watermark $s^d$ with dimensions $l\times c\times h\times w$. The actual watermark capacity is $k=\frac{{l\times c\times h\times w}}{f_c \cdot f^2_{hw}}$ bits.

\noindent
\textbf{Watermark randomization.} If we know the distribution of the diffused watermark $s^d$, we can directly utilize distribution-preserving sampling to obtain the corresponding latent representations $z^s_T$. However, in practical scenarios, its distribution is always unknown. Hence, we introduce a stream key $K$ to transform $s^d$ into a distribution-known randomized watermark $m$ through encryption. Considering the use of computationally secure stream cipher, such as ChaCha20~\cite{bernstein2008chacha}, $m$ follows a uniform distribution, i.e., $m$ is a random binary bit stream.

\noindent
\textbf{Distribution-preserving sampling driven by randomized watermark.}  When each dimension represents $l$-bit randomized watermark $m$, this $l$ bits can be regarded as an integer $y \in [0, 2^l-1]$. Since $m$ is a ciphertext, $y$ follows a discrete uniform distribution, i.e., $p(y) = \frac{1}{2^l}$ for $y = 0, 1, 2, \dots, 2^l-1$. Let $f(x)$ denote the probability density function of the Gaussian distribution $\mathcal{N}(0, I)$, and $ppf$ denotes the quantile function. We divide $f(x)$ into $2^l$ equal cumulative probability portions. When $y=i$, the watermarked latent representation $z^s_T$ falls into the $i$-th interval, which means $z^s_T$ should follow the conditional distribution:
\begin{equation}
      p(z^s_T|y=i)=\left\{\begin{array}{cl}		2^l\cdot f(z^s_T) & ppf(\frac{i}{2^l}) < z^s_T \leq ppf(\frac{i+1}{2^l})  \\ 	0 &otherwise\end{array} \right. .\label{eq:conditional_dis}
\end{equation}
The probability distribution of $z^s_T$ is given by:
\begin{equation}
    p(z^s_T) = \sum^{2^l-1}_{i=0} p(z^s_T|y=i)p(y=i)=f(z^s_T) .\label{eq:distribution}
\end{equation}
\cref{eq:distribution} indicates that $z^s_T$  follows the same distribution as the randomly sampled latent representation $z_T \sim \mathcal{N}(0, I)$. Next, we elaborate on how this sampling is implemented.

Let the cumulative distribution function of $f(x)$ be denoted as $cdf$. We can obtain the cumulative distribution function of  \cref{eq:conditional_dis} as follows,
\vspace{-0.1cm}
\begin{equation}
\begin{aligned}
     &F(z^s_T|y=i)\\
     &\quad =\left \{\begin{array}{cl}	0&z^s_T<ppf(\frac{i}{2^l})\\	2^l\cdot cdf(z^s_T) - i & ppf(\frac{i}{2^l}) \leq z^s_T \leq ppf(\frac{i+1}{2^l})  \\ 	1 &z^s_T>ppf(\frac{i+1}{2^l}) \end{array} \right..\label{eq:cdf_condition}
\end{aligned}
\end{equation}

Given $y=i$, we aim to perform random sampling of $z^s_T$ within the interval $[ppf(\frac{i}{2^l}),ppf(\frac{i+1}{2^l})]$. 
The commonly used method is rejection sampling~\cite{hopper2002provably,chen2021distribution, add}, which can be time-consuming as it requires repeated sampling until $z^s_T$ falls into the correct interval. Instead, we can utilize the cumulative probability density. When randomly sampling $F(z^s_T|y=i)$, the corresponding $z^s_T$ is naturally obtained through random sampling. Since $F(z^s_T|y=i)$ takes values in $[0,1]$, sampling from it is equivalent to sampling from a standard uniform distribution, denoted as $u=F(z^s_T|y=i) \sim \mathcal{U}(0,1)$. Shift the terms of \cref{eq:cdf_condition}, and take into account that $cdf$ and $ppf$ are inverse functions, we have
\vspace{-0.1cm}
\begin{equation}
\vspace{-0.1cm}
    z^s_T=ppf(\frac{u+i}{2^l}). \label{eq:z}
\end{equation}
\vspace{-0.1cm}
\cref{eq:z} represents the process of sampling the watermarked latent representation $z^s_T$ driven by the randomized watermark $m$. To extract the watermark, its inverse map is
\vspace{-0.1cm}
\begin{equation}
    i = \lfloor 2^l \cdot cdf(z^s_T) \rfloor .\label{eq:i}
\end{equation}
\vspace{-0.1cm}

\noindent
\textbf{Image generation.} After the sampling process, the watermark is embedded in the latent representation  $z^s_T$, and the subsequent generation process is no different from the regular generation process of SD. Here, we employ the DPMSolver~\cite{lu2022dpm} for iterative denoising of $z^s_T$.
In addition to DPMSolver~\cite{lu2022dpm}, other continuous-time samplers based on ordinary differential equation (ODE) solvers~\cite{song2020denoising}, such as DDIM~\cite{song2020denoising}, DEIS~\cite{zhang2022fast}, PNDM~\cite{liu2022pseudo}, and UniPC~\cite{zhao2023unipc}, can be used too. After obtaining denoised $z^s_0$, the watermarked image $X^s$ is generated using the decoder $\mathcal{D}$: $X^s=\mathcal{D}(z^s_0)$.

\subsection{Watermark Extraction}
\noindent
 \textbf{DDIM Inversion.} Using the SD encoder $\mathcal{E}$, we first restore $X'^s$ to the latent space $z'^s_0 = \mathcal{E}(X'^s)$. Then, we introduce the DDIM inversion~\cite{song2020denoising} to estimate the additive noise. 
 It can be considered that $z'^s_T \approx z^s_T$.
We also observe that although DDIM inversion is derived from DDIM, it can apply to other continuous-time samplers based on ODE solvers.

\noindent
\textbf{Watermark reduction from latent representations.} After obtaining $z'^s_T$, according to the inverse transformation defined in \cref{eq:i}, the tensor can be converted into a bit stream $m'$. Subsequently, $m'$ is decrypted using $K$ to obtain $s'^{d}$. Inverse diffusion of the watermark results in $f_c \cdot f^2_{hw}$ copies of the watermark. Similar to voting, if the bit is set to 1 in more than half of the copies, the corresponding watermark bit is set to $1$; otherwise, it is set to $0$. This process restores the true binary watermark sequence $s'$.

\subsection{Proof of Lossless Performance} 
In prior works, the incorporation of watermark embedding modules inevitably results in a decline in model performance, as typically evaluated using metrics such as Peak Signal-to-Noise Ratio (PSNR) and Fréchet Inception Distance (FID)~\cite{heusel2017gans}, which are more suitable for assessing post-processing methods. To assess methods that integrate the watermark embedding and generation processes, we propose a definition for the impact of watermark embedding on model performance, drawing on the complexity-theoretic definition of steganographic security~\cite{hopper2002provably}. This definition is based on a probabilistic game between a watermarked image $X^s$ and a normally generated image $X$. The tester $\mathcal{A}$ can use any watermark to drive the sampling process and generate $X^s$, similar to the \textit{chosen hidden text attacks} ~\cite{hopper2002provably}, which we refer to as \textit{chosen watermark tests}. The watermarking method is performance-lossless under \textit{chosen watermark tests}, if for any polynomial-time tester $\mathcal{A}$ and key $K\leftarrow \textsf{KeyGen}_{\mathcal{G}(1^{\rho})}$, it holds that
\vspace{-0.1cm}
\begin{equation}
\left|\Pr\left[\mathcal{A}\left({X^s}\right)=1\right]-\Pr\left[\mathcal{A}\left(X\right)=1\right]\right|<\textsf{negl}\left(\rho\right).
    \label{eq:definition}
\end{equation}
\vspace{-0.1cm}
 Here, $\rho$ represents the length of the security parameter, such as the key $K$, and $\textsf{negl}(\rho)$ is a negligible term relative to $\rho$.

We prove the statement using a proof by contradiction. First, assume that the watermarked image $X^s$ and the normally generated image $X$ are distinguishable, meaning
\vspace{-0.1cm}
\begin{equation}
\begin{aligned}
\left|\Pr\left[\mathcal{A}\left({X^s}\right)=1\right]-\Pr\left[\mathcal{A}\left(X\right)=1\right]\right| =\delta, \label{eq:distinguish}
\end{aligned}
\end{equation}
\vspace{-0.1cm}
where $\delta$ is non-negligible with respect to the key $K$. Let the iterative denoising process be denoted as $Q(\cdot)$, and substitute the LDM decoder $\mathcal{D}$ into \cref{eq:distinguish}, we have
\begin{equation}
\begin{aligned}
&\left|\Pr\left[\mathcal{A}\left(\mathcal{D}\left(Q\left(z_T^s\right)\right)\right)=1|m=E\left(K,s^d\right)\right]\right.\\
        &\left.\quad-\Pr\left[\mathcal{A}\left(\mathcal{D}\left(Q\left(z_T\right)\right)\right)=1|z_T\leftarrow \mathcal{N}\left(0,I\right)\right]\right| =\delta, \label{eq:distinguish_QE}
\end{aligned}
\end{equation}
where the randomized watermark $m$ is obtained by encrypting the diffused watermark $s^d$ using the encryption algorithm $E$ with key $K$. Note that \cref{eq:distribution} contains the fact that distribution-preserving sampling driven by randomized watermark and random sampling are equivalent. Therefore, we denote sequence-driven sampling as $S(\cdot)$. $z_T^s$ can naturally be obtained by sampling driven by $m$, i.e., $z^s_T = S(m)$. On the other hand, $z_T$ can be considered as obtained by sampling driven by a truly random sequence $r$ of the same length as $m$, i.e., $z_T = S(r)$. \cref{eq:distinguish_QE} can be written as
\vspace{-0.1cm}
\begin{equation}
\begin{aligned}
&\left|\Pr[\mathcal{A}\left(\mathcal{D}\left(Q\left(S\left(m\right)\right)\right)\right)=1|m=E(K,s^d)]\right.\\
    &\left.\quad\quad\quad\quad\quad\quad-\Pr\left[\mathcal{A}\left(\mathcal{D}\left(Q\left(S\left(r\right)\right)\right)\right)=1\right]\right| =\delta. \label{eq:distinguish_QES}
\end{aligned}
\end{equation}
\vspace{-0.1cm}
Sampling $S(\cdot)$, denoising $Q(\cdot)$, and decoder $\mathcal{D}$ can be considered as subroutines that the tester $\mathcal{A}_{\mathcal{D}, Q, S}$ can use. Thus, \cref{eq:distinguish_QES} can be simplified,
\vspace{-0.1cm}
\begin{equation}
\begin{aligned}
&\left|\Pr[\mathcal{A}_{\mathcal{D},Q,S}\left(m\right)=1|m=E\left(K,s^d\right)]\right.\\
    &\left.\quad\quad\quad\quad\quad\quad\quad\quad-\Pr\left[\mathcal{A}_{\mathcal{D},Q,S}\left(r\right)=1\right]\right| =\delta. \label{eq:distinguish_simply}
\end{aligned}
\end{equation}
\vspace{-0.1cm}

Note that $S(\cdot)$, $Q(\cdot)$, and $\mathcal{D}$ are all polynomial-time programs, so the time taken by the tester $\mathcal{A}_{\mathcal{D}, Q, S}$ to make the distinction is also polynomial. \cref{eq:distinguish_simply} essentially states that it is possible to distinguish between  $m$ and  $r$ in polynomial time. However, we have used the computationally secure stream cipher ChaCha20~\cite{bernstein2008chacha} in watermark randomization, which means that $m$ as a pseudorandom sequence cannot be distinguished from a truly random sequence in polynomial time. \cref{eq:distinguish_simply} contradicts the computational security property of ChaCha20~\cite{bernstein2008chacha}. Therefore, \cref{eq:distinguish_simply} is not valid, leading us back to our initial assumption that \cref{eq:distinguish} is also not valid. This implies that the watermarked image $X^s$ and the normally generated image $X$ are indistinguishable in polynomial time. Hence, Gaussian Shading is performance-lossless under \textit{chosen watermark tests}.

\section{Experiments}
This section focuses on experimental analysis, including details of the experimental setup, performance evaluation of Gaussian Shading, comparison with baseline methods, ablation experiments, and potential attacks.

\subsection{Implementation Details}
\noindent
\textbf{SD models.} In this paper, we focus on text-to-image LDM, hence we select SD~\cite{rombach2022high}  provided by huggingface. We evaluate Gaussian Shading as well as baseline methods, using three versions of SD: V1.4, V2.0, and V2.1. The size of the generated images is $512\times 512$, and the latent space dimension is $4 \times 64 \times 64$. During inference, we employ the prompt from Stable-Diffusion-Prompt\footnote{\href{https://huggingface.co/datasets/Gustavosta/Stable-Diffusion-Prompts}{
Stable-Diffusion-Prompts}}, with a guidance scale of 7.5. We sample 50 steps using DPMSolver~\cite{lu2022dpm}. Considering that users tend to propagate the generated images without retaining the corresponding prompts, we use an empty prompt for inversion, with a  scale of 1. We perform 50 steps of inversion using  DDIM inversion~\cite{song2020denoising}.

\noindent
\textbf{Watermarking methods.} In the main experiments, the settings for Gaussian Shading are $f_c=1, f_{hw}=8, l=1$, resulting in an actual capacity of 256 bits. We select five baseline methods: three officially used by SD, namely DwtDct~\cite{cox2007digital}, DwtDctSvd~\cite{cox2007digital}, and RivaGAN~\cite{zhang2019robust}, a multi-bit watermarking called Stable Signature~\cite{fernandez2023stable}, and a train-free invisible watermarking called Tree-Ring~\cite{wen2023tree}.

\begin{figure}[t]
    \centering
    \subfloat[]{\label{Fig:watermaeked}\includegraphics[width=.18\linewidth]{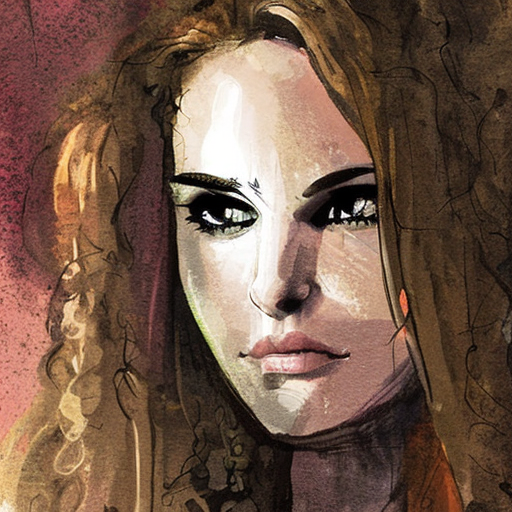}}\hspace{0.005\linewidth}
    \subfloat[]{\label{Fig:a}\includegraphics[width=.18\linewidth]{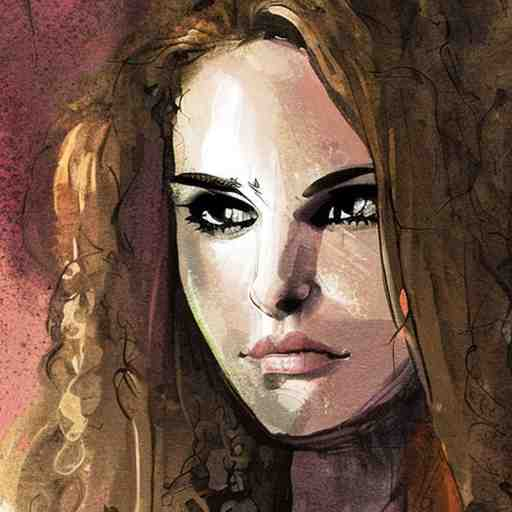}}\hspace{0.005\linewidth}
    \subfloat[]{\label{Fig:b}\includegraphics[width=.18\linewidth]{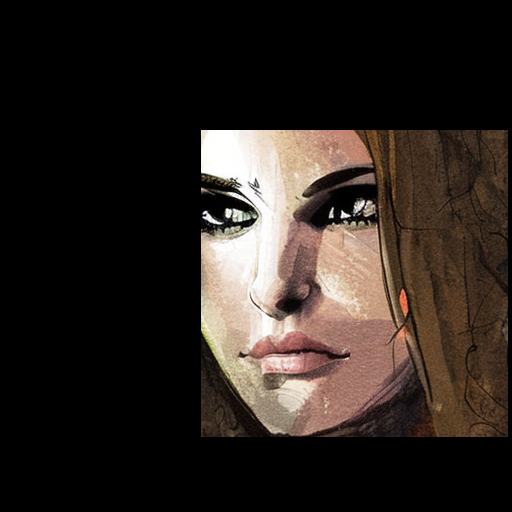}}\hspace{0.005\linewidth}
 \subfloat[]{\label{Fig:c}\includegraphics[width=.18\linewidth]{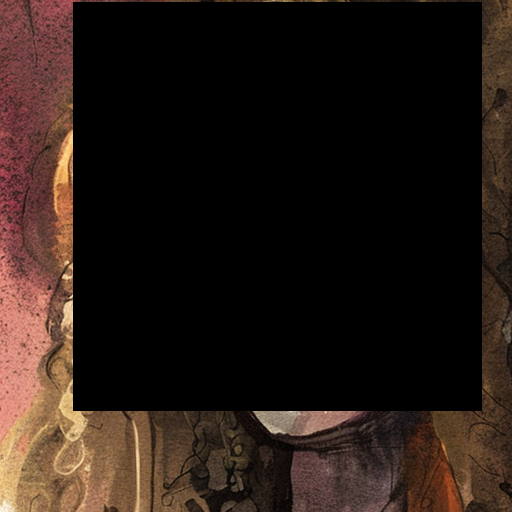}}\hspace{0.005\linewidth}
 \subfloat[ ]{\label{Fig:d}\includegraphics[width=.18\linewidth]{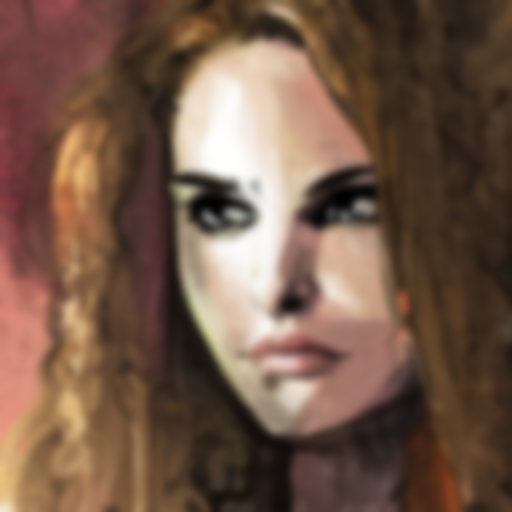}}\hspace{0.005\linewidth}
 \subfloat[]{\label{Fig:e}\includegraphics[width=.18\linewidth]{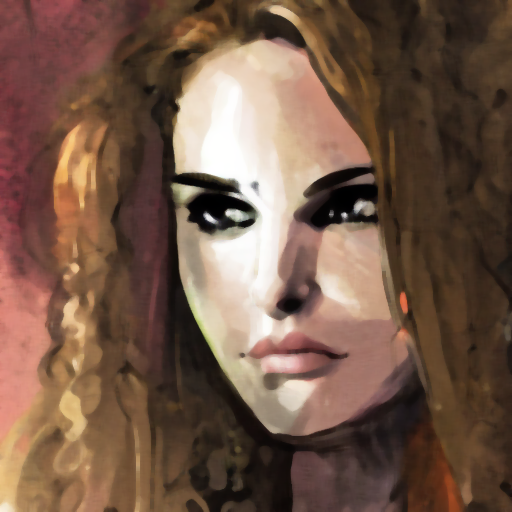}}\hspace{0.005\linewidth}
 \subfloat[]{\label{Fig:f}\includegraphics[width=.18\linewidth]{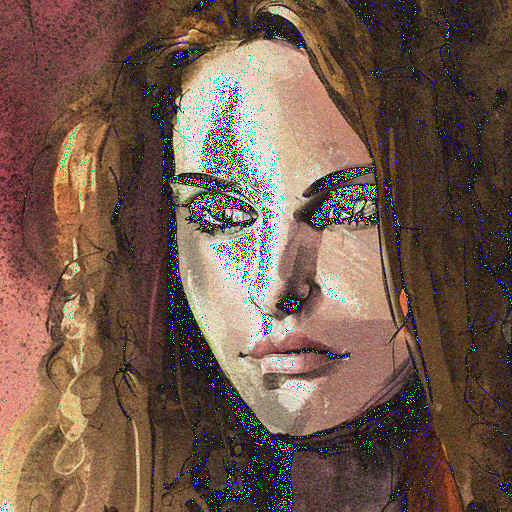}}\hspace{0.005\linewidth}
 \subfloat[]{\label{Fig:g}\includegraphics[width=.18\linewidth]{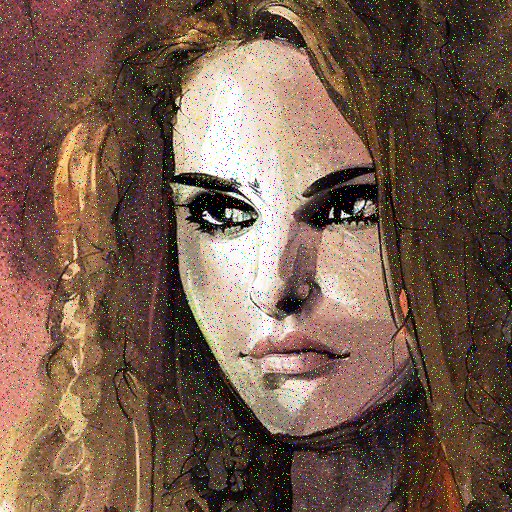}}\hspace{0.005\linewidth}
 \subfloat[]{\label{Fig:h}\includegraphics[width=.18\linewidth]{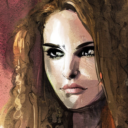}}\hspace{0.005\linewidth}
 \subfloat[]{\label{Fig:i}\includegraphics[width=.18\linewidth]{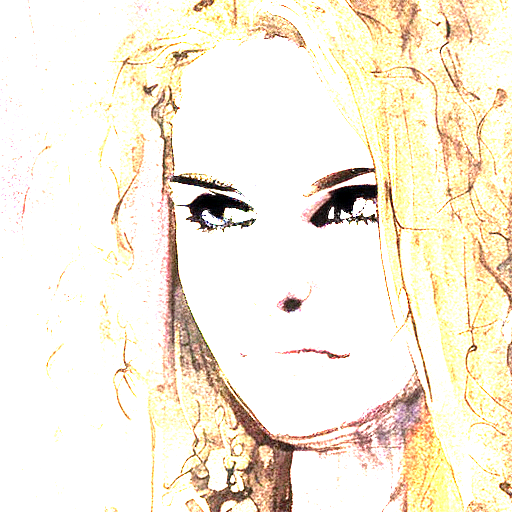}}\hspace{0.005\linewidth}
 \vspace{-0.3cm}
    \caption{Watermarked image is attacked by different noise. (a) Watermarked image. (b) JPEG, $QF=25$. (c) 60\% area Random Crop (RandCr).  (d) 80\% area Random Drop (RandDr). (e) Gaussian Blur, $r=4$ (GauBlur). (f) Median Filter, $k=7$ (MedFilter). (g) Gaussian Noise, $\mu= 0$, $\sigma = 0.05$ (GauNoise). (h) Salt and Pepper Noise, $p = 0.05$ (S\&PNoise). (i) 25\% Resize and restore (Resize). (j) Brightness, $factor=6$. }
    \label{fig:robustness}
\vspace{-0.7cm}
\end{figure}

\noindent
\textbf{Robustness evaluation} To evaluate the robustness, we select nine representative types of noise shown in \cref{fig:robustness}. We conduct experiments following the noise strength in \cref{fig:robustness}.

\noindent
\textbf{Evaluation metrics.} In the detection scenario, we calculate the true positive rate (TPR) corresponding to a fixed false positive rate (FPR). In the traceability scenario, we calculate the bit accuracy. To measure the bias in model performance, we compute the FID~\cite{heusel2017gans} and CLIP-Score~\cite{radford2021learning} for 10 batches of watermarked images and perform a $t$-test on the mean FID and CLIP-Score compared to that of watermark-free images.

All experiments are conducted using the PyTorch 1.13.0 framework, running on a single RTX 3090 GPU.

\subsection{Performance of Gaussian Shading}

\begin{figure}[t]
    \centering
    \subfloat[Detection results.]{\label{Fig:tpr}\includegraphics[width=0.4\linewidth]{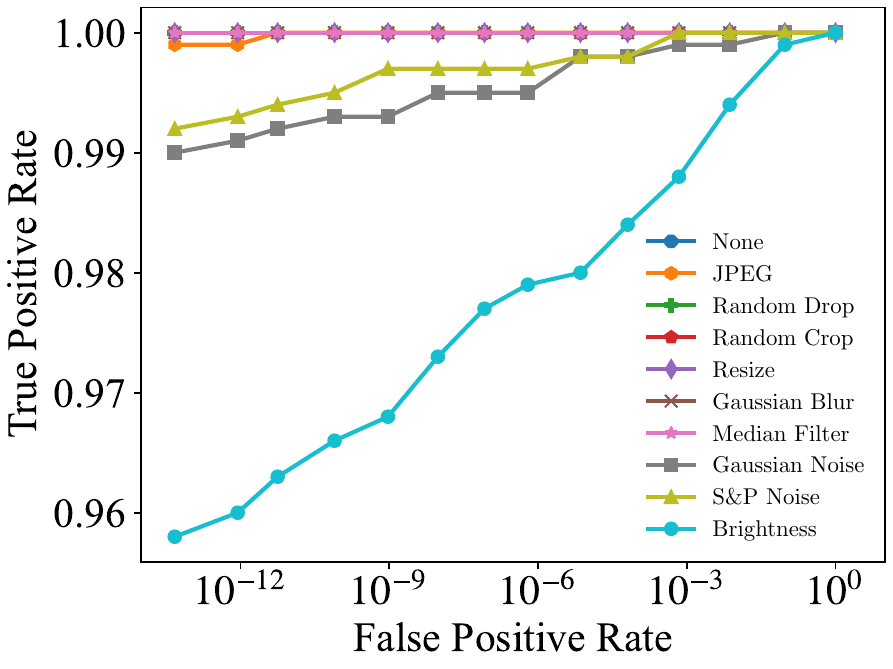}}\hspace{0.1\linewidth}
     \subfloat[Traceability results.]{\label{Fig:t_acc}\includegraphics[width=0.4\linewidth]{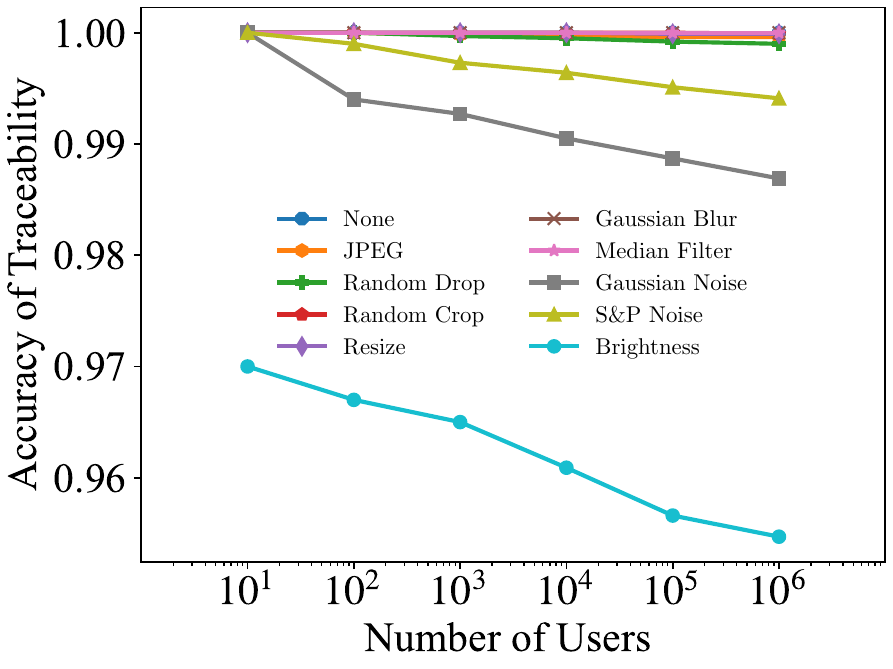}}\hspace{0.005\linewidth}
     \vspace{-0.3cm}
    \caption{Performance of Gaussian Shading. }
    \label{fig:performance}
    \vspace{-0.7cm}
\end{figure}

\noindent
 \textbf{Detection.} In the detection scenario, we consider Gaussian Shading as a single-bit watermark, with a fixed watermark $s$. We approximate the FPR to be controlled at $10^0,10^{-1},\dots,10^{-13}$, calculate the corresponding threshold $\tau$ 
 , and test the TPR on $1,000$ watermarked images  
 To mitigate the effects of randomness, we perform 5 trials with different $s$ and compute the average TPR. See \cref{Fig:tpr}, when the FPR is controlled at $10^{-13}$, the TPR remains at least $0.99$  for eight out of the nine cases. Although the TPR for Brightness is only $0.953$, it is still a promising result.

\noindent
\textbf{Traceability.} In this scenario, Gaussian Shading serves as a multi-bit watermark. Assuming Alice provides services to $N$ users, Alice needs to allocate one watermark for each user. In our experiments, we assume that $N' = 1,000$ users generate images, with each user generating $10$ images, resulting in a dataset of $10,000$ watermarked images.

During testing, we calculate the threshold $\tau$  
to control the FPR at $10^{-6}$. Note that when computing traceability accuracy, we need to consider two types of errors: false positives, where watermarked images are not detected, and traceability errors, where watermarked images are detected but attributed to the wrong user. Therefore, we first  
determine whether the image contains a watermark. If it does, we calculate the number of matching bits $Acc$ with all $N$ users on the platform. The user with the highest $Acc$ is considered the one who generated the image. Finally, we verify whether the correct user has been traced. When $N > N'$, it can be assumed that some users have been assigned a watermark but have not generated any images.

See \cref{Fig:t_acc}, 
when $N = 10^6$, Gaussian Shading exhibits almost perfect traceability in seven cases. Although the traceability accuracy for Brightness is only $95.47\%$, if a user generates two images, the probability of successfully tracing him is still no less than $99\%$.

\begin{table*}
  \centering
   \resizebox{\textwidth}{!}{
  \begin{tabular}{@{}ccccccc@{}}
    \toprule
    \multirow{2}{*}{Methods} &
		\multicolumn{6}{c}{Metrics} \\
		\cmidrule{2-7}
		&TPR (Clean) & TPR (Adversarial) &Bit Acc. (Clean)&Bit Acc. (Adversarial)& FID ($t$-value $\downarrow$)  & CLIP-Score ($t$-value $\downarrow$)\\
    \midrule
    Stable Diffusion&-&-&-&-&25.23$\pm$.18&0.3629$\pm$.0006\\
    \midrule
    DwtDct~\cite{cox2007digital} &0.825/0.881/0.866&0.172/0.178/0.173&0.8030/0.8059/0.8023&0.5696/0.5671/0.5622&24.97$\pm$.19 (3.026)&0.3617$\pm$.0007 (3.045)\\
    DwtDctSvd~\cite{cox2007digital} &\textbf{1.000}/\textbf{1.000}/\textbf{1.000}&0.597/0.594/0.599&0.9997/0.9987/0.9987&0.6920/0.6868/0.6905&24.45$\pm$.22 (8.253)&0.3609$\pm$.0009 (4.452)\\
    RivaGAN~\cite{zhang2019robust} &0.920/0.945/0.963 &0.697/0.697/0.706&0.9762/0.9877/0.9921&0.8986/0.9124/0.9019&24.24$\pm$.16 (12.29)&0.3611$\pm$.0009 (4.259)\\
    Tree-Ring~\cite{wen2023tree} &\textbf{1.000}/\textbf{1.000}/\textbf{1.000} &0.894/0.898/0.906&-&-&25.43$\pm$.13 (2.581)&0.3632$\pm$.0006 (0.8278)\\
    Stable Signature~\cite{fernandez2023stable} &\textbf{1.000}/\textbf{1.000}/\textbf{1.000}&0.502/0.505/0.496&0.9987/0.9978/0.9979&0.7520/0.7472/0.7500&25.45$\pm$.18 (2.477)&0.3622$\pm$.0027 (0.7066)\\
    \textbf{Ours} &\textbf{1.000}/\textbf{1.000}/\textbf{1.000}&\textbf{0.997}/\textbf{0.998}/\textbf{0.996}&\textbf{0.9999}/\textbf{0.9999}/\textbf{0.9999}&\textbf{0.9753}/\textbf{0.9749}/\textbf{0.9724}&25.20$\pm$.22 (\textbf{0.3567})&0.3631$\pm$.0005 (\textbf{0.6870}) \\
    \bottomrule
    
  \end{tabular}}
  \vspace{-0.3cm}
  \caption{Comparison results. We control the FPR at $10^{-6}$, and evaluate the TPR and bit accuracy for SD V1.4/V2.0/V2.1. To assess the bias in model performance, we conduct a $t$-test on SD V2.1. Adversarial here refers to the average performance of a series of noises. Additional results can be found in Supplementary Material. }
  \label{tab:compare}
  \vspace{-0.3cm}
\end{table*}

\begin{figure*}[t]
    \centering
    \subfloat[Guidance scales.]{\label{Fig:gs}\includegraphics[width=.18\linewidth]{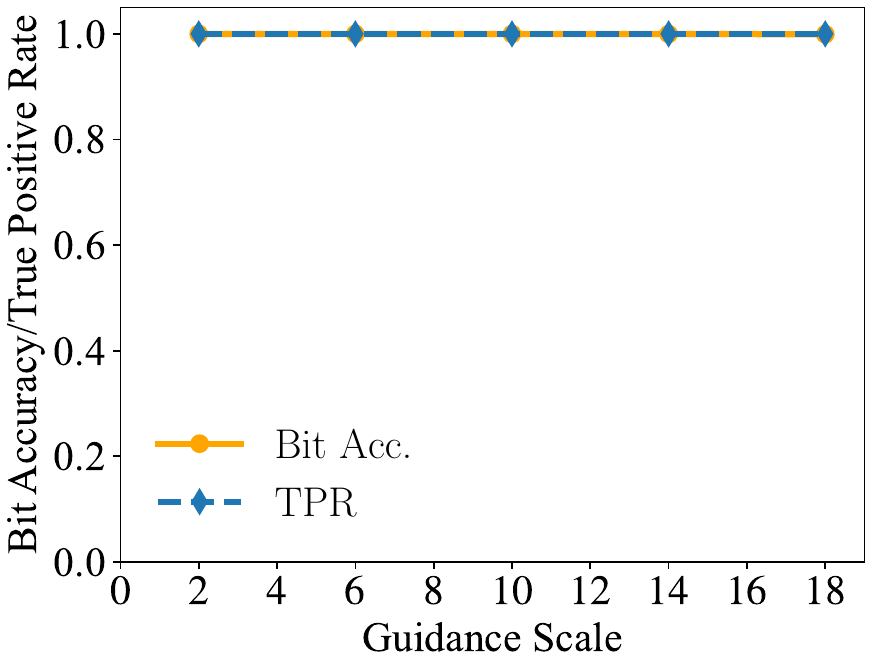}}\hspace{0.005\linewidth}
    \subfloat[JPEG]{\label{Fig:jp}\includegraphics[width=.18\linewidth]{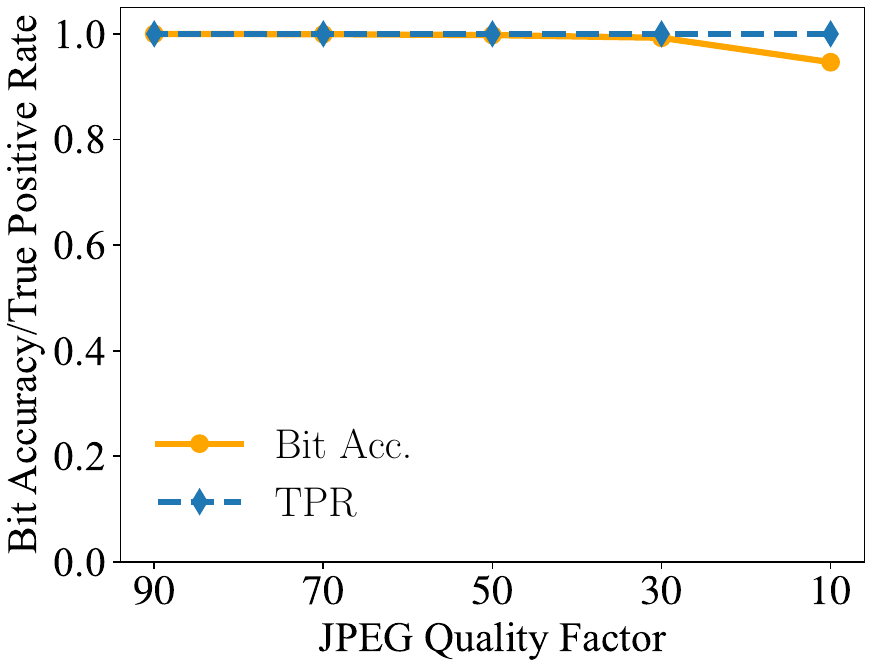}}\hspace{0.005\linewidth}
    \subfloat[Random Crop.]{\label{Fig:rc}\includegraphics[width=.18\linewidth]{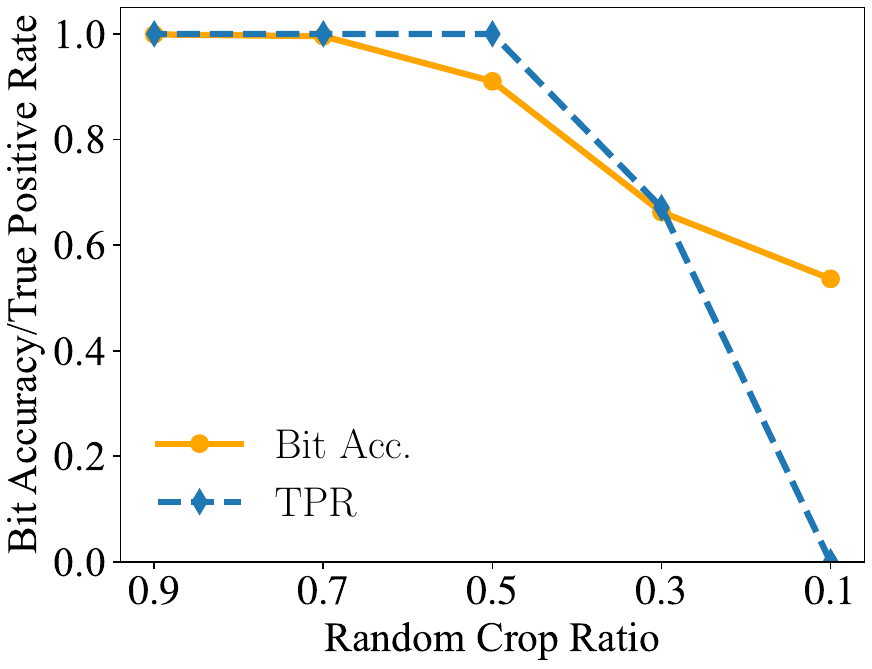}}\hspace{0.005\linewidth}
 \subfloat[Random Drop.]{\label{Fig:rd}\includegraphics[width=.18\linewidth]{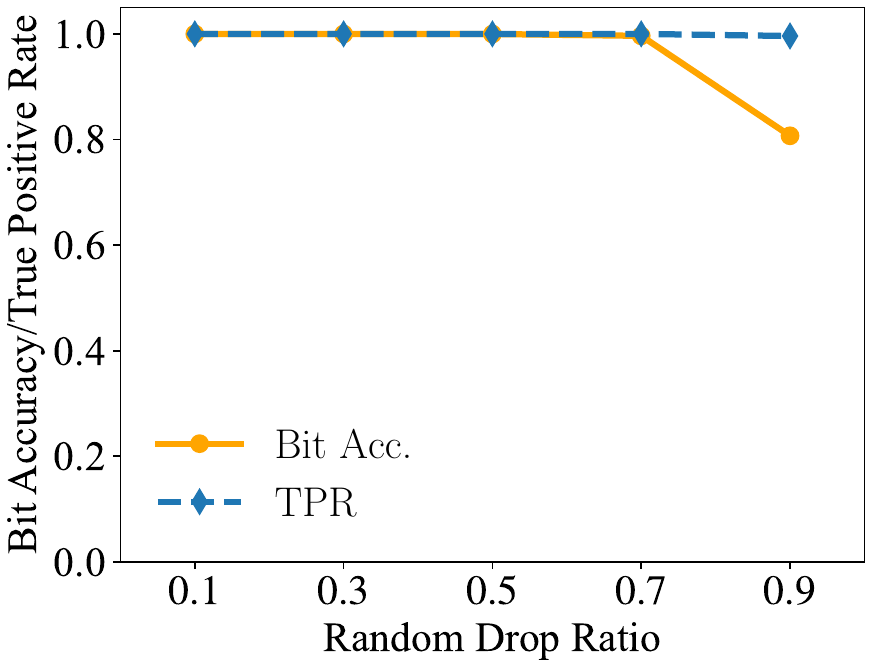}}\hspace{0.005\linewidth}
 \subfloat[Gaussian Blur. ]{\label{Fig:rs}\includegraphics[width=.18\linewidth]{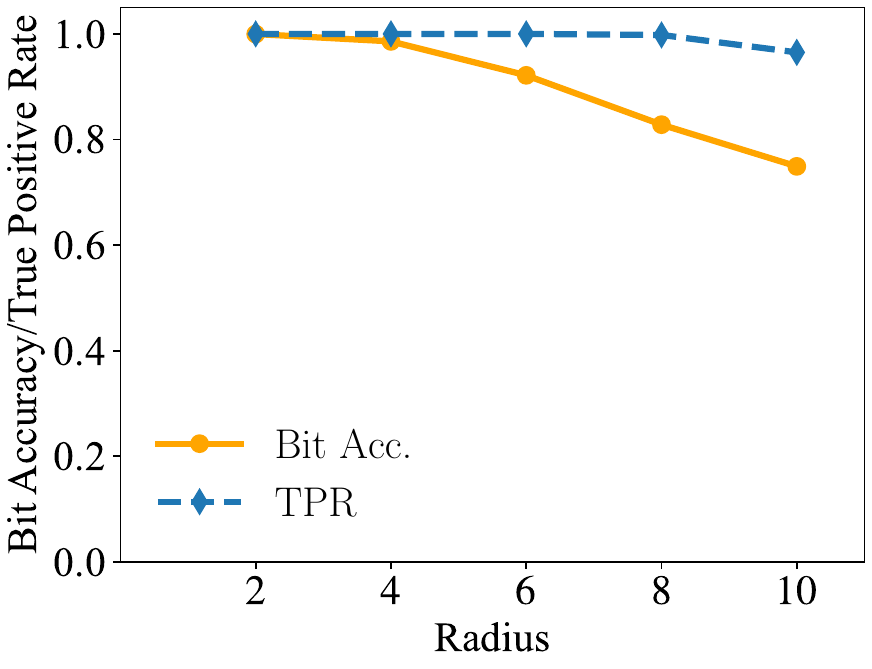}}\hspace{0.005\linewidth}
 \subfloat[Median Filter.]{\label{Fig:gb}\includegraphics[width=.18\linewidth]{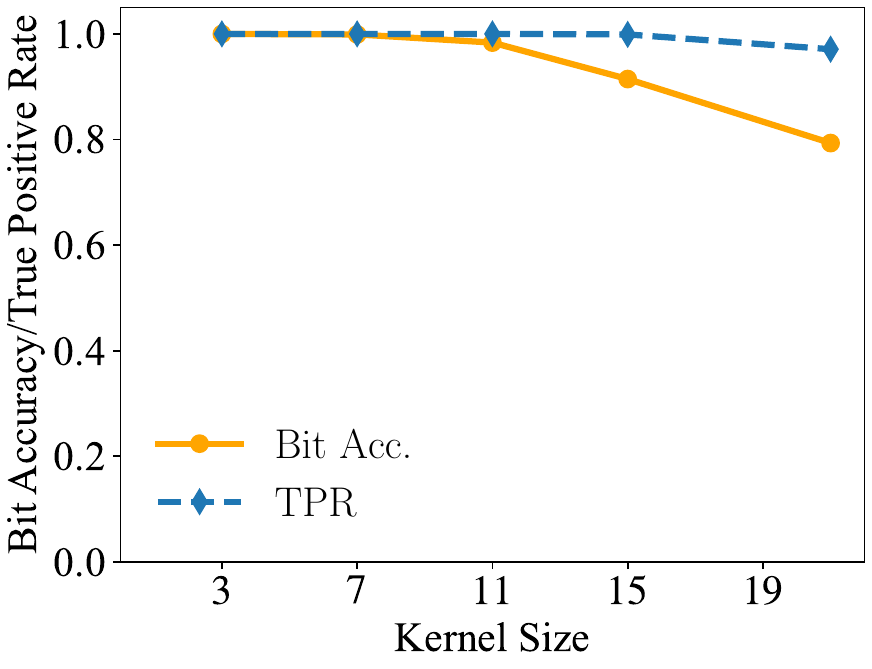}}\hspace{0.005\linewidth}
 \subfloat[Gaussian Noise.]{\label{Fig:mf}\includegraphics[width=.18\linewidth]{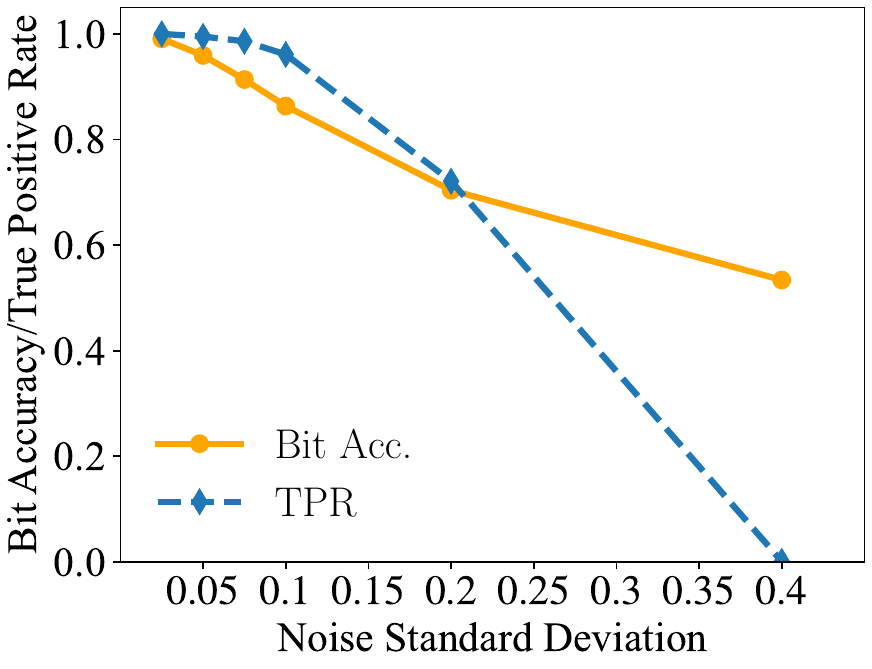}}\hspace{0.005\linewidth}
 \subfloat[Salt and Pepper Noise.]{\label{Fig:gn}\includegraphics[width=.18\linewidth]{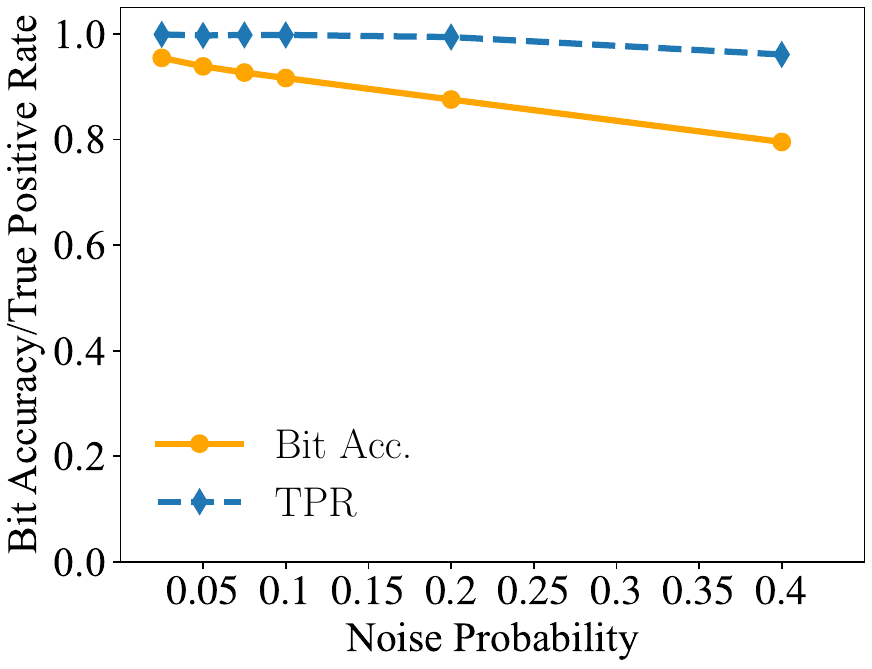}}\hspace{0.005\linewidth}
 \subfloat[Resize.]{\label{Fig:sp}\includegraphics[width=.18\linewidth]{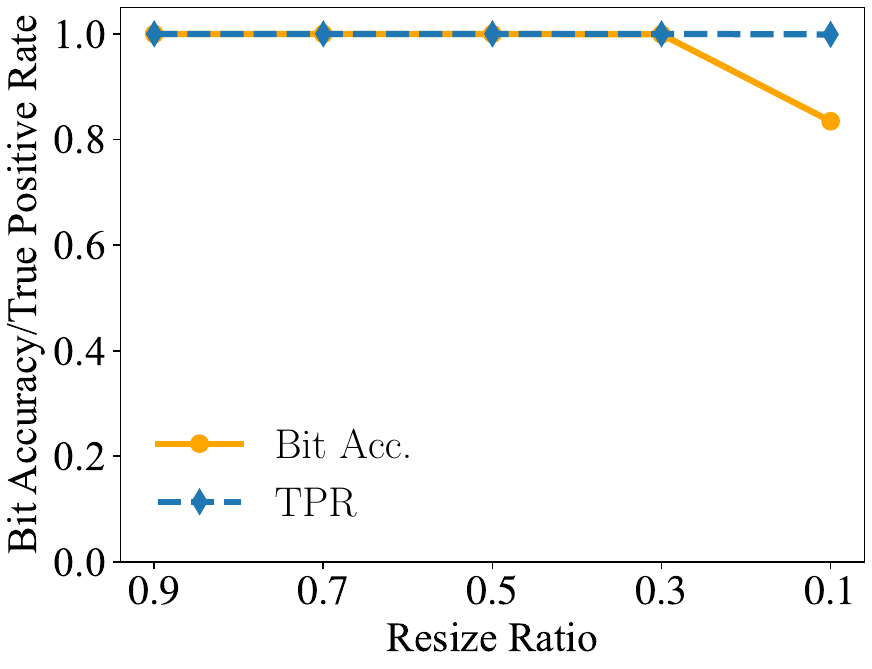}}\hspace{0.005\linewidth}
 \subfloat[Brightness.]{\label{Fig:br}\includegraphics[width=.18\linewidth]{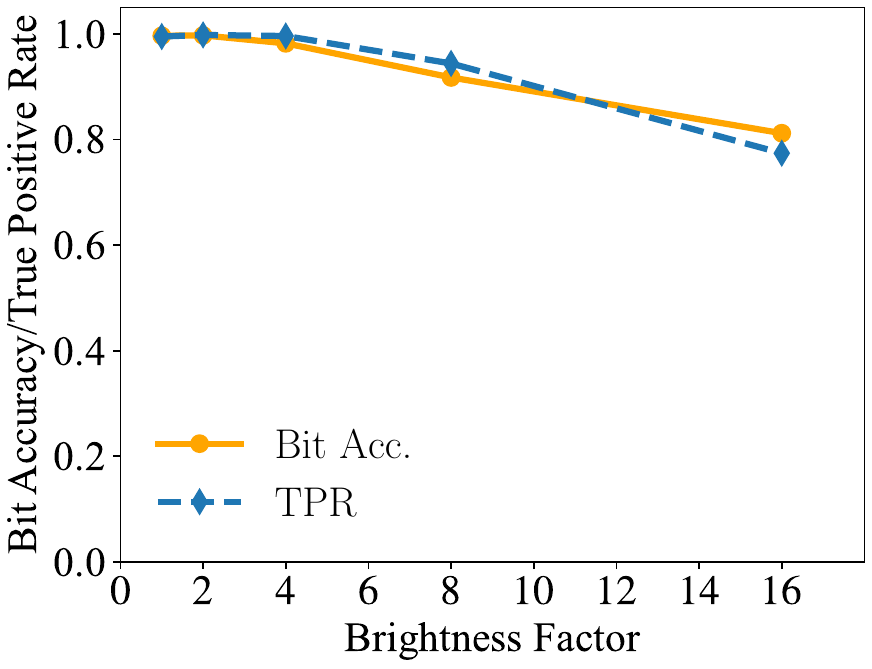}}\hspace{0.005\linewidth}
 \vspace{-0.3cm}
    \caption{Ablation studies.}
\vspace{-0.7cm}
    \label{fig:ablation}
\end{figure*}

\subsection{Comparison to Baselines}
In this section, we compare the performance of Gaussian Shading with baselines on SD V1.4, V2.0, and V2.1.  We use our implementations for each method, see details in Supplementary Material.

 We conduct tests on $1,000$ generated images for each method respectively. See \cref{tab:compare}.  Gaussian Shading exhibits strong robustness and significantly outperforms baselines in both scenarios.  In terms of bit accuracy, it surpasses the best-performing baseline by approximately $7\%$. This can be attributed to the extensive diffusion of the watermark throughout the entire latent space, establishing a profound binding between the watermark and the image semantics.

 To measure the performance bias introduced by the watermark embedding, we apply a $t$-test to evaluate. The hypotheses are $H_0 : \mu_s = \mu_0, H_1 : \mu_s \neq \mu_0$, where $\mu_s$ and $\mu_0$ represent the average FID~\cite{heusel2017gans} or CLIP-Score~\cite{radford2021learning} of multiple sets of watermarked and watermark-free images, respectively. A lower $t$-value indicates a higher
probability that $H_0$ holds. If the $t$-value is larger than a
threshold, $H_0$ is rejected, and model performance is considered to have been affected. See \cref{tab:compare}, Gaussian Shading achieves the smallest $t$-value, which indirectly reflects its performance-lossless characteristic. For a detailed analysis of the $t$-test, please refer to the Supplementary Material.

\begin{table*}
  \centering
  \footnotesize
  \begin{tabular}{@{}ccccccccc@{}}
    \toprule
    \multirow{2}{*}{Noise} &
		\multicolumn{8}{c}{$f_c$-$f_{hw}$  ($k$ bits)} \\
		\cmidrule{2-9}
		&1-2 (4096) & 4-1 (4096) &1-4 (1024)&4-2 (1024)&1-8 (256)&  4-4 (256) & 1-16 (64)& 4-8 (64)\\
    \midrule
    None &0.9413&0.9380&0.9985&0.9980&0.9999&0.9999&1.0000&1.0000\\
    Adversarial &0.7302&	0.7238&	0.8769&0.8614&	0.9724&	0.9671&	0.9959&	0.9953
 \\
    \bottomrule
    \end{tabular}
  \vspace{-0.3cm}
  \caption{Bit accuracy with different factors $f_c$ and $f_{hw}$ , where $l=1$. Additional results can be found in Supplementary Material.}
  \label{tab:capacity_chw}
  \vspace{-0.4cm}
\end{table*}

\subsection{Ablation Studies}
In this section, we conduct comprehensive ablation experiments on SD V2.1 to demonstrate hyperparameter selection. Unless specified, we generate $1,000$ images and test the TPR and the bit accuracy with a theoretical FPR of $10^{-6}$.

\noindent
\textbf{Watermark capacity.} The watermark capacity is determined by three parameters: channel diffusion factor $f_c$, height-width diffusion factor $f_{hw}$, and embedding rate $l$. See \cref{tab:capacity_chw}, to balance the capacity and robustness of Gaussian Shading, we chose $f_c=1$ and $f_{hw}=8$. After fixing $f_c$ and $f_{hw}$, we vary  $l$ to examine if it could enhance the capacity, and additional results can be found in Supplementary Material. Considering all factors, we determine that the optimal solution is $f_c=1$, $f_{hw}=8$, and $l=1$, resulting in a watermark capacity of $256$ bits.

\noindent
\textbf{Sampling methods.}
To validate the generalization, we select five commonly used sampling methods, all continuous-time samplers based on ODE solvers~\cite{song2020denoising}. See~\cref{tab:sampler}, all of them exhibit excellent performance with a bit accuracy of approximately $97\%$ against noises.

\begin{table}
  \centering
  \resizebox{0.45\textwidth}{!}{
  \footnotesize
  \begin{tabular}{@{}cccccc@{}}
    \toprule
    \multirow{2}{*}{Noise} &
		\multicolumn{5}{c}{Sampling Methods} \\
		\cmidrule{2-6}
		& \makecell[c]{DDIM \\ \cite{song2020denoising}}&  \makecell[c]{UniPC \\ \cite{zhao2023unipc}}&
  \makecell[c]{PNDM \\ \cite{liu2022pseudo}}&\makecell[c]{DEIS \\ \cite{zhang2022fast}}& \makecell[c]{DPMSolver \\ \cite{lu2022dpm}}\\
    \midrule
    None &0.9999&1.0000&1.0000&0.9999&0.9999\\
    Adversarial &  0.9706&0.9628&0.9721&0.9715&0.9724\\

    \bottomrule
    \end{tabular}}
  \vspace{-0.3cm}
  \caption{Bit accuracy with different sampling methods. Additional results can be found in Supplementary Material. These methods differ only in accuracy and order. DDIM is a first-order estimate of the ODE. Accordingly, DDIM inversion ensures a lower bound on the accuracy of the inversion process. Therefore, it can naturally be applied to higher-order and higher-accuracy methods.}
  \label{tab:sampler}
  \vspace{-0.3cm}
  
\end{table}

\noindent
\textbf{Impact of the inversion step.} In practice, the inference step is often unknown, which introduces a mismatch with the inversion step. See \cref{tab:stepmismatch}, such mismatch introduces minimal loss in accuracy. Considering the high efficiency of existing samplers, the inference step generally does not exceed $50$. Therefore, we set the inversion step to $50$.

\begin{table}
  \centering
  \footnotesize
  \begin{tabular}{@{}ccccc@{}}
    \toprule
    \multirow{2}{*}{\makecell[c]{Inference \\ Step}} &
		\multicolumn{4}{c}{Inversion Step} \\
		\cmidrule{2-5}
		&10 & 25 &50&100\\
    \midrule
    10& 0.9999&0.9999&0.9999&0.9999\\
    25 & 0.9998&0.9999&1.0000&1.0000\\
    50 &0.9995&0.9997&0.9999&0.9999\\
    100 & 0.9994&0.9996&0.9999&0.9999\\
    \bottomrule
    \end{tabular}
  \vspace{-0.3cm}
  \caption{Bit accuracy with different inference and inversion step. }
  \label{tab:stepmismatch}
  \vspace{-0.7cm}
\end{table}

\noindent
\textbf{Guidance scales.} Given diverse user preferences for image-prompt alignment, larger guidance scales ensure faithful adherence to prompts, while smaller scales grant the model greater creative freedom. In SD, the guidance scale is typically selected from the range of $[5, 15]$. Hence, experiments cover the range of $2$ to $18$. For the inversion, an empty prompt is used for guidance, and the guidance scale is fixed at 1, assuming unknown information during extraction. In \cref{Fig:gs}, the bit accuracy of Gaussian Shading surpasses $99.9\%$, showing its reliability in real-world-like scenes.

\noindent
\textbf{Noise intensities.} To further test the robustness, we conduct experiments using different intensities of noises. See \cref{Fig:jp,Fig:rc,Fig:rd,Fig:gb,Fig:mf,Fig:gn,Fig:sp,Fig:rs,Fig:br}, for Random Crop and Gaussian Noise, performance declines significantly with higher intensities. However, for the other seven types of noise, even at high intensities, the bit accuracy remains approximately $80\%$.

\subsection{Attacks against Gaussian Shading}
We consider two malicious attacks: compression attack, where the attacker employs a neural network to compress watermarked images, and inversion attack, assuming the attacker is aware of the watermark embedding method, enabling them to modify the image's latent representations.

\begin{figure}[t]
    \centering
    \subfloat[Detection results.]{\label{Fig:psnr_tpr}\includegraphics[width=.32\linewidth]{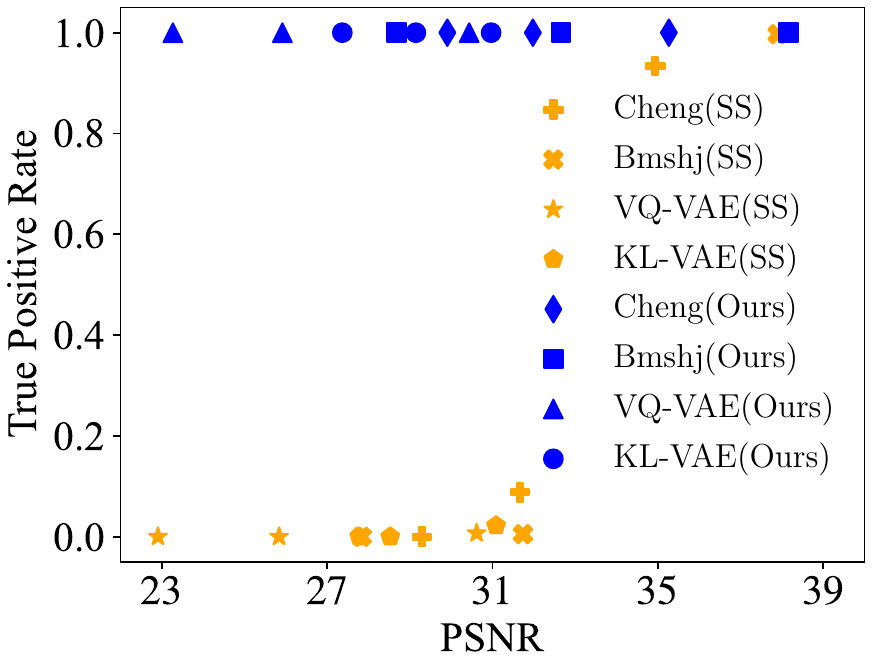}}\hspace{0.005\linewidth}
    \subfloat[Traceability results.]{\label{Fig:psnr_acc}\includegraphics[width=.32\linewidth]{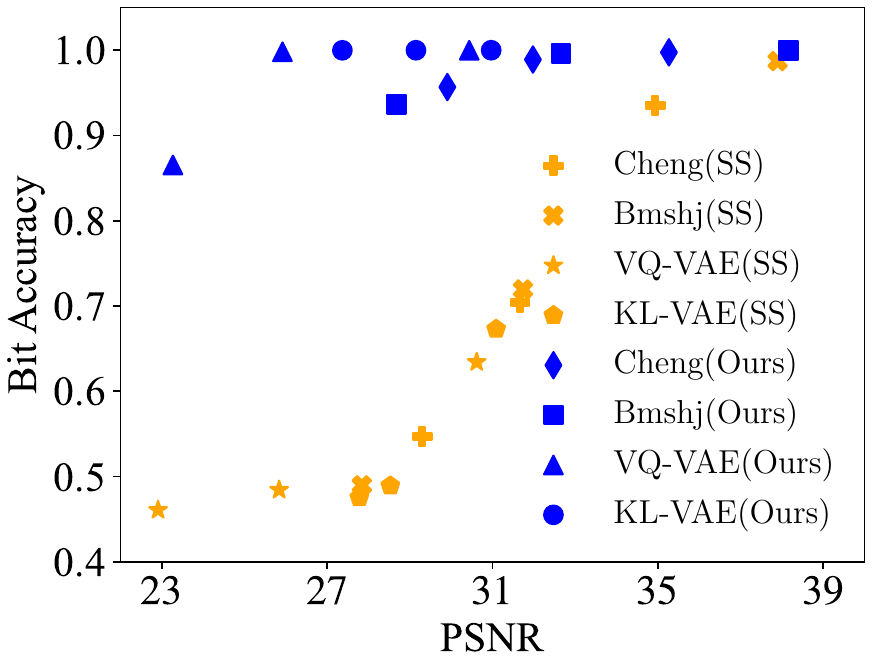}}\hspace{0.005\linewidth}
    \subfloat[Inversion attack.]{\label{Fig:flip}\includegraphics[width=.32\linewidth]{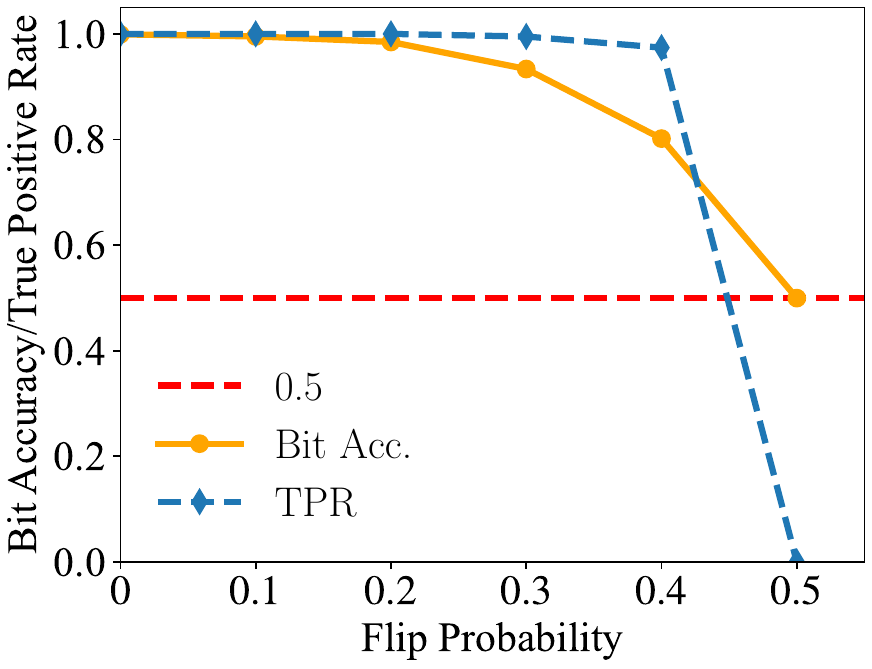}}\hspace{0.005\linewidth}
    \vspace{-0.3cm}
    \caption{Performance of Gaussian Shading under Malicious Attack, where (a) and (b) are under compression attack and (c) is under inversion attack. }
    \label{fig:attack}
    \vspace{-0.5cm}
\end{figure}

\noindent
\textbf{Compression attack.} We utilize popular auto-encoders~\cite{balle2018variational,cheng2020learned,esser2021taming,rombach2022high}  to compare Stable Signature (SS) with Gaussian Shading across various compression rates. Additionally, we assess the compression quality through the PSNR between the compressed and watermarked images. See \cref{Fig:psnr_tpr,Fig:psnr_acc}, Gaussian Shading significantly outperforms Stable Signature. This is because Gaussian Shading diffuses the watermark across the entire semantic space of images, while Stable Signature relies solely on the image texture.

\noindent
 \textbf{Inversion attack.} Assuming the attacker is aware of the embedding method, a more effective approach to erasing is through inversion to obtain latent representations and subsequently modify them. We validate the robustness against such attacks. Importantly, our experiments assume the strongest attacker capability of using the same model as Alice for precise inversion. In real-world scenarios, where the watermark embedding is not publicly available, the attacker's capabilities would be weaker.

Specifically, we perform inversion to obtain latent representations and randomly flip a certain rate of them. Using the flipped latent representations, we regenerate the images and extract the watermark. See \cref{Fig:flip}. the watermark can still be reliably extracted when the flipping rate (FR) is less than 0.4. At high FRs, significant changes in images are observed. Although the watermark cannot be accurately extracted, we consider the image transformed into a different one, resulting in the content not intended to be protected.

From another perspective, the attacker can launch a forgery attack by performing inversion on an innocuous image from Bob and generating harmful content using a different prompt. See \cref{Fig:flip}, when the FR is 0, Alice can accurately trace Bob based on the forgeries, enabling the attacker to successfully frame Bob. Therefore, protecting the model from leakage is crucial for operators.

\section{Limitations}

Despite extensive experimental validation of Gaussian Shading's superior performance, our work still has certain limitations. Firstly, the usage scenarios are restricted due to the reliance on DDIM inversion~\cite{song2020denoising}, which necessitates the utilization of continuous-time samplers based on ODE solvers~\cite{song2020denoising} like DPMSolver~\cite{lu2022dpm}. Secondly, Gaussian Shading employs stream ciphers,  necessitating proper key usage and management on the deployment platform. Additionally, we assume that the model is not publicly accessible, and only operators can verify the watermark, providing a certain level of protection against white-box attacks and ensuring security. However, if a legitimate third party requires watermark verification, cooperation from the operators becomes necessary. Lastly, Gaussian Shading is vulnerable to forgery attacks, emphasizing the importance for operators to safeguard the model parameters.

\section{Conclusion and Future Work}

We propose Gaussian Shading, a provably performance-lossless watermarking applied to diffusion models. Compared to baseline methods, Gaussian Shading offers simplicity and effectiveness by making a simple modification in the sampling process of the initial latent representation. Extensive experiments validate the superior performance in both detection and traceability scenarios. To our knowledge, we are the first to propose and implement a performance-lossless approach in image watermarking.

Regarding future work, we will introduce more efficient inversion methods~\cite{zhang2023robustness,wallace2023edict} and include a wider range of sampling methods. Additionally, careful consideration should be given to counteracting forgery attacks.

\vspace{0.4cm}
\noindent
 \textbf{Acknowledgement.} This work was supported in part by the Natural Science Foundation of China under Grant U2336206, 62102386, 62072421, 62372423, and 62121002.

\clearpage
{
    \small
    \bibliographystyle{ieeenat_fullname}
    \bibliography{main}
}

\clearpage
\setcounter{page}{1}
\maketitlesupplementary


\section{Details of Gaussian Shading}
\label{sec:dogs}

\subsection{Watermark Statistical Test}\label{dec_tra}

\noindent
\textbf{Detection.} Alice embeds a single-bit watermark, represented by $k$-bit binary watermark $s \in \{0,1\}^k$, into each generated image using Gaussian Shading. This watermark serves as an identifier for her model. Assuming the watermark $s'$ is extracted from image $X$, the detection test for the watermark can be represented by the number of matching bits between two watermark sequences, $Acc(s,s')$. When the threshold $\tau \in \{0,\dots,k\}$ is determined, if
\begin{equation}
 Acc(s,s') \geq \tau ,
  \label{smeq:acc}
\end{equation}
it is deemed that $X$ contains the watermark.

In previous works~\cite{yu2021artificial}, it is commonly assumed that the extracted watermark bits $s'_1,\dots,s'_k$ from the vanilla images are independently and identically distributed, with $s'_i$ following a Bernoulli distribution with parameter $0.5$. Thus, $Acc(s,s')$ follows a binomial distribution with parameters $(k,0.5)$. It is worth noting that if we extract from a vanilla image and decrypt it using a computationally secure stream key~\cite{bernstein2008chacha}, the resulting diffused watermark $s'^d$ should be a pseudorandom bit stream, and the corresponding watermark $s'$ would also be pseudorandom. In other words, the bits $s'_1,\dots,s'_k$ are independently and identically distributed, and each $s'_i$ follows a Bernoulli distribution with a parameter of $0.5$. This aligns perfectly with the above assumption. 

Once the distribution of $Acc(s,s')$ is determined, the false positive rate ($\operatorname{FPR}$) is defined as the probability that $Acc(s,s')$ of a vanilla image exceeds the threshold $\tau$. This probability can be further expressed using the regularized incomplete beta function $B_x(a;b)$~\cite{fernandez2023stable},
\begin{equation}
    \begin{aligned}
    \operatorname{FPR}(\tau) &=\mathbb{P}\left(Acc\left(s, s^{\prime}\right)>\tau \right) =\frac{1}{2^{k}}\sum_{i=\tau+1}^{k} \left(\begin{array}{c}k \\i\end{array}\right)\\
    &=B_{1 / 2}(\tau+1, k-\tau) .\label{smeq:detection}
\end{aligned}
\end{equation}

\noindent
\textbf{Traceability.} To enable traceability, Alice needs to assign a watermark $s^i \in \{0,1\}^k$ to each user, where $i=1,\dots,N$ and $N$ represents the number of users. During the traceability test, the bit matching count $Acc(s^1,s'),\dots, Acc(s^N,s')$ needs to be computed for all $N$ watermarks. If none of the $N$ tests exceed the threshold $\tau$, the image is considered not generated by Alice's model. However, if at least one test passes, the image is deemed to be generated by Alice's model, and the index with the maximum matching count is traced back to the corresponding user, i.e., $\operatorname*{argmax}_{i=1,\dots, N}Acc(s^i,s')$. When a threshold $\tau$ is given, the $\operatorname{FPR}$ can be expressed as follows~\cite{fernandez2023stable},
\begin{equation}
    \operatorname{FPR}(\tau, N) = 1 - (1-\operatorname{FPR}(\tau))^N \approx N \cdot \operatorname{FPR}(\tau) . \label{smeq:trace}
\end{equation}

\subsection{Details of Denoising and Inversion }

\noindent
\textbf{Markov chains of diffusion models.}
 DDPM~\cite{ho2020denoising} proposed that the diffusion model consists of two Markov chains used for adding and removing noise. The forward chain is pre-designed to transform the data distribution $q_0(x_0)$ into a simple Gaussian distribution $q_T(x_T) \approx \mathcal{N}(x_T|0, \sigma^2I)$ over a time interval of $T$. Here, $\sigma > 0$, and the transition probability $q(x_t|x_{t-1})$ is defined as $\mathcal{N}(x_t; \sqrt{\alpha_t}x_0, (1-\alpha_t)I)$, where $\alpha_t$ is a predetermined hyperparameter. By virtue of the Markov property, we have
 \begin{equation}
     q(x_t|x_0) = \mathcal{N}(x_t|\beta_tx_0, \sigma^2_tI) , \label{eq:forward}
 \end{equation}
  with $\beta_t = \sqrt{\overline{\alpha}_{t}}$, $\sigma^2_t = 1-\overline{\alpha}_t$, and ${\overline{\alpha}_{t}}= \prod_{i=0} ^t \alpha_i$. 
  
  The transition kernel of the reverse chain is learned by a neural network $\theta$ and aims to generate data from a Gaussian distribution with the transition probability distribution defined as 
  \begin{equation}
        p_{\theta}(x_{t-1}|x_t) = \mathcal{N}(x_{t-1};\mu_\theta(x_t,t),\Sigma(x_t,t)). \label{eq:backward}
  \end{equation}

  For LDM~\cite{rombach2022high}, since the diffusion process occurs in the latent space $\mathcal{Z}$, \cref{eq:forward} and \cref{eq:backward} should be rewritten for the latent representations $z$ of LDM as follows:
   \begin{equation}
     q(z_t|z_0) = \mathcal{N}(z_t|\beta_tz_0, \sigma^2_tI) , 
 \end{equation}
   \begin{equation}
        p_{\theta}(z_{t-1}|z_t) = \mathcal{N}(z_{t-1};\mu_\theta(z_t,t),\Sigma(z_t,t)).
  \end{equation}

   \begin{table*}
  \centering
  \resizebox{\textwidth}{!}{
  \begin{tabular}{@{}ccccccc@{}}
    \toprule
    \multirow{2}{*}{Noise} &
		\multicolumn{6}{c}{Methods} \\
		\cmidrule{2-7}
		&DwtDct~\cite{cox2007digital} & DwtDctSvd~\cite{cox2007digital} &RivaGAN~\cite{zhang2019robust}&Tree-Ring~\cite{wen2023tree}&Stable Signature~\cite{fernandez2023stable}& \textbf{Ours}\\
    \midrule
    None &0.825/0.881/0.866&\textbf{1.000}/\textbf{1.000}/\textbf{1.000}&0.920/0.945/0.963&\textbf{1.000}/\textbf{1.000}/\textbf{1.000}&\textbf{1.000}/\textbf{1.000}/\textbf{1.000}& \textbf{1.000}/\textbf{1.000}/\textbf{1.000} \\
    JPEG &  0/0/0&0.013/0.019/0.015&0.156/0.085/0.214&0.997/\textbf{1.000}/0.994&0.210/0.217/0.198&\textbf{0.999}/\textbf{1.000}/\textbf{0.997}\\
    RandCr &0.982/0.967/0.952&\textbf{1.000}/0.998/0.999&0.868/0.878/0.891&0.997/\textbf{1.000}/\textbf{1.000}&\textbf{1.000}/0.998/0.993&\textbf{1.000}/\textbf{1.000}/\textbf{1.000} \\
    RandDr &0/0/0&0/0/0&0.887/0.885/0.862&\textbf{1.000}/\textbf{1.000}/0.998&0.971/0.980/0.972&\textbf{1.000}/\textbf{1.000}/\textbf{1.000} \\
    GauBlur &0/0.001/0.002&0.430/0.419/0.432&0.328/0.331/0.316&\textbf{1.000}/\textbf{1.000}/0.997&0/0/0&\textbf{1.000}/\textbf{1.000}/\textbf{1.000} \\
    MedFilter & 0/0.001/0.001&0.996/0.999/\textbf{1.000}&0.863/0.832/0.873&\textbf{1.000}/\textbf{1.000}/\textbf{1.000}&0.001/0/0&\textbf{1.000}/\textbf{1.000}/\textbf{1.000}\\
    GauNoise & 0.354/0.353/0.364&0.842/0.862/0.884&0.441/0.457/0.535&0/0.006/0.077&0.424/0.406/0.404&\textbf{0.996}/\textbf{0.995}/\textbf{0.995}\\
    S\&PNoise & 0.089/0.160/0.102&0/0/0&0.477/0.411/0.431&0.972/0.986/0.994&0.072/0.078/0.052&\textbf{1.000}/\textbf{0.998}/\textbf{0.997}\\
    Resize & 0/0.005/0.008&0.985/0.977/0.983&0.850/0.886/0.887&\textbf{1.000}/\textbf{1.000}/\textbf{1.000}&0/0/0&\textbf{1.000}/\textbf{1.000}/\textbf{1.000}\\
    Brightness &0.126/0.114/0.124&0.110/0.072/0.074&0.480/0.404/0.386&0.084/0.089/0.092&0.843/0.862/0.849&\textbf{0.974}/\textbf{0.991}/\textbf{0.979} \\
    \makecell[c]{Average of \\Adversarial}&0.172/0.178/0.173&0.597/0.594/0.599&0.697/0.697/0.706&0.894/0.898/0.906&0.502/0.505/0.496&\textbf{0.997}/\textbf{0.998}/\textbf{0.996} \\
    \bottomrule
    \end{tabular}}
  \caption{The comparison in the detection scenario. Gaussian Shading demonstrates the best performance.}
  \label{tab:compare_detection}
\end{table*}

\begin{table*}
  \centering
   \resizebox{\textwidth}{!}{
   \small
  \begin{tabular}{@{}cccccc@{}}
    \toprule
    \multirow{2}{*}{Noise} &
		\multicolumn{5}{c}{Methods} \\
		\cmidrule{2-6}
		&DwtDct~\cite{cox2007digital} & DwtDctSvd~\cite{cox2007digital} &RivaGAN~\cite{zhang2019robust}&Stable Signature~\cite{fernandez2023stable}& \textbf{Ours}\\
    \midrule
    None &0.8030/0.8059/0.8023&0.9997/0.9987/0.9987&0.9762/0.9877/0.9921&0.9987/0.9978/0.9949&\textbf{0.9999}/\textbf{0.9999}/\textbf{0.9999}  \\
    JPEG &0.5018/0.5047/0.5046&0.5197/0.5216/0.5241&0.7943/0.7835/0.8181&0.7901/0.7839/0.7893&\textbf{0.9918}/\textbf{0.9905}/\textbf{0.9872}  \\
    RandCr & 0.7849/0.7691/0.7673&0.8309/0.7942/0.8151&0.9761/0.9723/0.9735&\textbf{0.9933}/\textbf{0.9903}/\textbf{0.9883}&0.9803/0.9747/0.9669\\
    RandDr & 0.5540/0.5431/0.5275&0.5814/0.5954/0.6035&0.9678/0.9720/0.9683&\textbf{0.9768}/\textbf{0.9747}/\textbf{0.9736}&0.9676/0.9687/0.9649\\
    GauBlur & 0.5000/0.5027/0.5039&0.6579/0.6466/0.6459&0.8323/0.8538/0.8368&0.4137/0.4110/0.4112&\textbf{0.9874}/\textbf{0.9846}/\textbf{0.9858}\\
    MedFilter & 0.5171/0.5243/0.5199&0.9208/0.9287/0.9208&0.9617/0.9585/0.9696&0.6374/0.6399/0.6587&\textbf{0.9987}/\textbf{0.9970}/\textbf{0.9990}\\
    GauNoise & 0.6502/0.6294/0.6203&0.7960/0.7950/0.8159&0.8404/0.9648/0.8776&0.7831/0.7766/0.7768&\textbf{0.9636}/\textbf{0.9556}/\textbf{0.9592}\\
    S\&PNoise &0.5784/0.6021/0.5845&0.5120/0.5267/0.5250&0.8881/0.8838/0.8634&0.7192/0.7170/0.7144&\textbf{0.9406}/\textbf{0.9433}/\textbf{0.9385} \\
    Resize &0.5067/0.5184/0.5135&0.8743/0.8498/0.8630&0.9602/0.9731/0.9733&0.5278/0.5051/0.5177&\textbf{0.9970}/\textbf{0.9975}/\textbf{0.9976} \\
    Brightness & 0.5336/0.5097/0.5175&0.5346/0.5234/0.5016&0.8666/0.8496/0.8369&0.9276/0.9267/0.9204&\textbf{0.9508}/\textbf{0.9623}/\textbf{0.9527}\\
    \makecell[c]{Average of \\Adversarial}&0.5696/0.5671/0.5622&0.6920/0.6868/0.6905&0.8986/0.9124/0.9019&0.7520/0.7472/0.7500&\textbf{0.9753}/\textbf{0.9749}/\textbf{0.9724} \\
    \bottomrule
    \end{tabular}}
  \caption{The comparison in the traceability scenario comparison. Although Gaussian Shading slightly underperforms Stable Signature in the presence of Random Crop and Random Drop, considering all the noise, Gaussian Shading still demonstrates the best overall performance.}
  \label{tab:compare_tracebility}
\end{table*}

\noindent
\textbf{Denoising method for Gaussian Shading.}
\label{sec:dpmsolver}
DPMSolver~\cite{lu2022dpm} is a higher-order ODE solver~\cite{song2020denoising}, and in this paper, we employ its second-order version during image generation, whose denoising process is as follows,
\begin{equation}
    \begin{array}{l}v_{t-1} = t_{\lambda}\left(\frac{\lambda_{{t-1}}+\lambda_{t}}{2}\right) \\
u_{t-1} = \frac{\beta_{v_{t-1}}}{\beta_{{t}}} z^s_{{t}}-\sigma_{v_{t-1}}\left(e^{\frac{h_{t-1}}{2}}-1\right) \epsilon_{\theta}\left(z^s_{{t}}, c,{t}\right) \\
z^s_{{t-1}} = \frac{\beta_{{t-1}}}{\beta_{{t}}} z^s_{{t}}-\sigma_{{t-1}}\left(e^{h_{t-1}}-1\right) \epsilon_{\theta}\left(u_{t-1},c, v_{t-1}\right)\end{array} , \label{eq:dpmsolver}
\end{equation}
where $\lambda_t = \lambda(t) = \log \left( \frac{\beta_t}{\sigma_t} \right)$, $t_\lambda(\cdot)$ represents the inverse function of $\lambda_{t}$, $h_{t-1}=\lambda_{t-1}-\lambda_{t}$, $t=1,2,\dots,T$, and $c$ indicates  the prompt used for text-to-image generation. 

\noindent
\textbf{Inversion method for Gaussian Shading.}
\label{sec:ddimiversion} We note that in DDIM~\cite{song2020denoising}, Song et al. proposed an inversion method where they used the Euler method to solve the ODE~\cite{song2020denoising} and obtained an approximate solution for the inverse process:
 \begin{equation}
 z'^s_{t+1} = \sqrt{\alpha_t}z'_{t} + \left(\sqrt{1-\overline{\alpha}_{t+1}}-\sqrt{\alpha_t - \overline{\alpha}_{t+1}}\right)\epsilon\left(z'_t, c,t\right) . \label{eq:ddim_inversion}
 \end{equation}
 According to \cref{eq:ddim_inversion} it is possible to estimate the noise to be added, which enables latent representation restoration.

\section{Experimental Details and Additional Experiments}\label{sec:ae}

 \subsection{Empirical check of the FPR}
\begin{figure}[t]
    \centering
    \subfloat[Theoretical FPR and measured FPR in detection scenario.]{\label{Fig:onebit}\includegraphics[width=.49\linewidth]{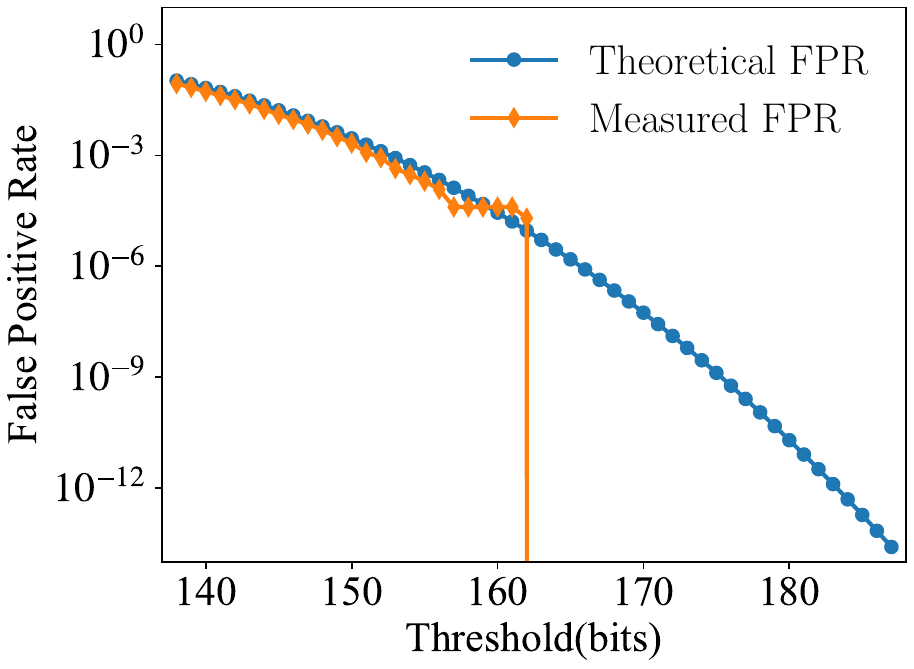}}\hspace{0.005\linewidth}
 \subfloat[Theoretical FPR and measured FPR in traceability scenario.]{\label{Fig:multibit}\includegraphics[width=.49\linewidth]{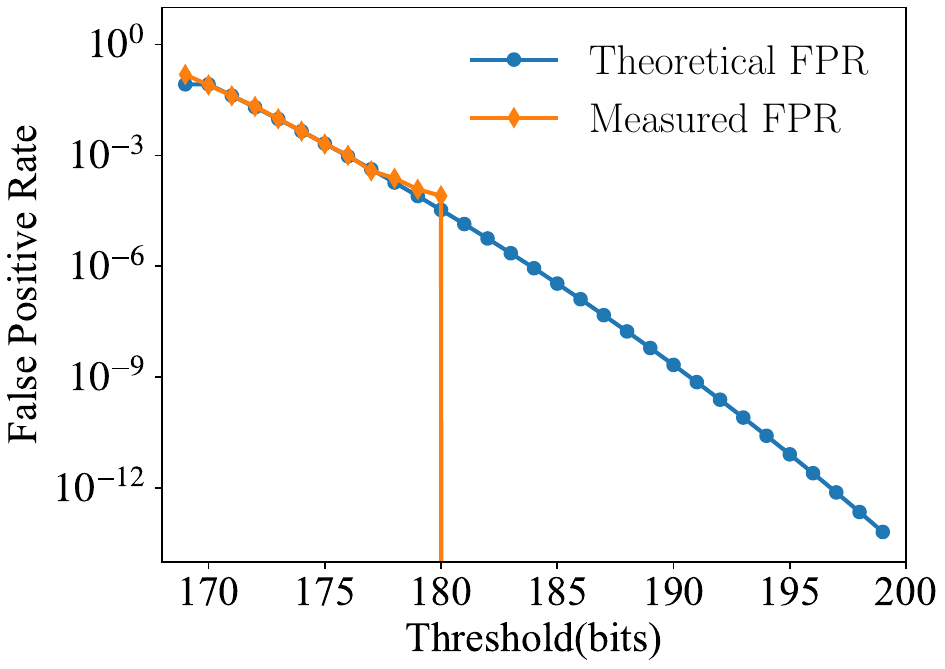}}\hspace{0.005\linewidth}
 \vspace{-0.2cm}
    \caption{Empirical check of the FPR. }
    \label{fig:performance_appendix}
    \vspace{-0.2cm}
\end{figure}

To test the actual FPR of Gaussian Shading, and to validate the accuracy of \cref{smeq:detection} and \cref{smeq:trace}, we performed watermark extraction on $50,000$ vanilla images from the ImageNet2014~\cite{deng2009imagenet} validation set. See \cref{fig:performance_appendix}, the theoretical and actual measured curves are very close, indicating that the theoretical thresholds derived from \cref{smeq:detection} and \cref{smeq:trace} can effectively guarantee the actual FPR.

\subsection{Details of Comparison Experiments}
\noindent
\textbf{Watermarking methods settings.} To ensure a fair comparison, we set the watermark capacity to 256 bits for DwtDct~\cite{cox2007digital} and DwtDctSvd~\cite{cox2007digital}. As RivaGAN~\cite{zhang2019robust} has a maximum capacity of only 32 bits, we retain this setting. The capacity and robustness of Stable Signature~\cite{fernandez2023stable} are determined by Hidden~\cite{zhu2018hidden}  trained in the first stage. However, in our experiments, we find that Hidden with a capacity of 256 bits did not converge during training. Additionally, if there are too many types of noise in the noise layer, Hidden does not converge either. As an alternative, we use the open-source model of Stable Signature with a capacity of 48 bits\footnote{\href{https://github.com/facebookresearch/stable_signature}{The GitHub Repository for Stable Signature}}. During fine-tuning, we utilize 400 images from the ImageNet2014~\cite{deng2009imagenet} validation set, with a batch size of 4 and 100 training steps. Tree-Ring~\cite{wen2023tree} is a single-bit watermark, and we only compare it in the detection scenario. Since its Rand mode is more closely aligned with the concept of performance-lossless, we adopt this setting. 

The specific experimental results in both scenarios are shown in \cref{tab:compare_detection} and \cref{tab:compare_tracebility}, respectively. In the detection scenario, the average TPR of Gaussian Shading remains above 0.995 in the presence of noise, surpassing the subpar performance of Tree-Ring by approximately 0.1. In the traceability scenario, the average bit accuracy of Gaussian Shading exceeds $97\%$ against noises, outperforming the second-best method, RivaGAN, by around $7\%$. In both scenarios, Gaussian Shading exhibits superior performance compared to baseline methods.

\noindent
\textbf{The $t$-test for model performance.} To measure the performance bias introduced by the watermark embedding, we apply a $t$-test to evaluate. 

We first generate 50,000/10,000 images using SD V2.1 for each watermarking method, divided into 10 groups of 5,000/1,000 images each. We then calculate the FID~\cite{heusel2017gans}/CLIP-Score~\cite{radford2021learning}   for each group and compute the average value $\mu_s$. Similarly, we generate 50,000/10,000 watermark-free images using SD V2.1, test the FID/CLIP-Score for 10 groups, and calculate the average value $\mu_0$. 
For the FID, we randomly select 5000 images from MS-COCO-2017~\cite{lin2014microsoft} validation set and calculate the scores using the aforementioned groups. For the CLIP-Score, we utilize OpenCLIP-ViT-G~\cite{cherti2023reproducible} to compute the image-text relevance.

If the model performance is maintained, then $\mu_s$ and $\mu_0$ should be statistically close to each other. Therefore, the hypotheses are
\begin{equation}
    H_0 : \mu_s = \mu_0, H_1 : \mu_s \neq \mu_0.
\end{equation}
The statistic $t\text{-}v$ is calculated as follows:
\begin{equation}
    t\text{-}v = \frac{|\mu_s - \mu_0|}{\sqrt{S^*\cdot (\frac{1}{n_s}+\frac{1}{n_0})}},
\end{equation}
 where 
 \begin{equation}
    S^*=\frac{1}{n_s+n_0-2}\left[(n_s-1)S_s^2+(n_0-1)S_0^2\right],
\end{equation}
$n_s$ and $n_0$ represent the number of testing times, which are both set to 10 in the experiments, and $S_s$ and $S_0$ represent the standard deviations of the FID/CLIP-Score for watermarked and watermark-free images, respectively.

A lower $t$-value indicates a higher
probability that $H_0$ holds. If the $t$-value is larger than a
threshold, $H_0$ is rejected, and model performance is considered to have been affected. The significance level for the test is set to $t\text{-}v_{0.05}(n_s+n_0-2)=t\text{-}v_{0.05}(18)\approx 2.101$. In terms of the FID, the $t$-values of the baseline methods, as depicted in \cref{tab:ttest}, are all greater than the critical value $t\text{-}v_{0.05}(18)\approx 2.101$, except for Gaussian Shading. Regarding the CLIP-Score, Tree-Ring, Stable Signature, and Gaussian Shading all exhibit competitive results. Note that the CLIP-Score tends to measure the alignment between generated images and prompts, while the FID is solely used to assess image quality. In summary, these baseline methods demonstrate a noticeable impact on the model's performance in a statistically significant manner. On the other hand, Gaussian Shading achieved the smallest $t$-value, which indirectly confirms its performance-lossless characteristic.

\begin{table}
  \centering
   \resizebox{0.47\textwidth}{!}{
  \begin{tabular}{@{}ccc@{}}
    \toprule
    \multirow{2}{*}{Methods} &
		\multicolumn{2}{c}{Metrics} \\
		\cmidrule{2-3}
		&FID ($t$-value $\downarrow$)  & CLIP-Score ($t$-value $\downarrow$)\\
    \midrule
    Stable Diffusion&25.23$\pm$.18&0.3629$\pm$.0006\\
    \midrule
    DwtDct~\cite{cox2007digital} &24.97$\pm$.19 (3.026)&0.3617$\pm$.0007 (3.045)\\
    DwtDctSvd~\cite{cox2007digital} &24.45$\pm$.22 (8.253)&0.3609$\pm$.0009 (4.452)\\
    RivaGAN~\cite{zhang2019robust} &24.24$\pm$.16 (12.29)&0.3611$\pm$.0009 (4.259)\\
    Tree-Ring~\cite{wen2023tree} &25.43$\pm$.13 (2.581)&0.3632$\pm$.0006 (0.8278)\\
    Stable Signature~\cite{fernandez2023stable} &25.45$\pm$.18 (2.477)&0.3622$\pm$.0027 (0.7066)\\
    \textbf{Ours} &25.20$\pm$.22 (\textbf{0.3567})&0.3631$\pm$.0005 (\textbf{0.6870}) \\
    \bottomrule
    
  \end{tabular}}
  \caption{Experimental results of $t$-test. }
  \label{tab:ttest}
\end{table}

\begin{table*}[t]
  \centering
  \small
  \begin{tabular}{@{}ccccccccc@{}}
    \toprule
    \multirow{2}{*}{Noise} &
		\multicolumn{8}{c}{$f_c$ - $f_{hw}$  ($k$ bits)} \\
		\cmidrule{2-9}
		&1-2 (4096) & 4-1 (4096) &1-4 (1024)&4-2 (1024)&1-8 (256)&  4-4 (256) & 1-16 (64)& 4-8 (64)\\
    \midrule
    None &0.9413&0.9380&0.9985&0.9980&0.9999&0.9999&1.0000&1.0000\\
    JPEG & 0.7685&0.7588&0.9204&0.9087&0.9872&0.9866&0.9973&0.9989 \\
    RandCr& 0.6735&0.6554&0.8177&0.7852&0.9669&0.9457&0.9981&0.9963\\
    RandDr& 0.6707&0.6785&0.8239&0.7754&0.9649&0.9444&0.9993&0.9985\\
    GauBlur& 0.7217&0.7205&0.8846&0.8832&0.9858&0.9881&0.9996&0.9998\\
    MedFilter &0.8151&0.8104&0.9637&0.9589&0.9990&0.9987&0.9999&1.0000 \\
    GauNoise & 0.7051&0.6933&0.8502&0.8366&0.9592&0.9539&0.9932&0.9933\\
    S\&PNoise & 0.6711&0.6661&0.8100&0.7987&0.9385&0.9366&0.9933&0.9914\\
    Resize & 0.7904&0.7861&0.9478&0.9438&0.9976&0.9976&0.9999&0.9999\\
    Brightness & 0.7558&0.7455&0.8737&0.8619&0.9527&0.9526&0.9829&0.9796\\
     \makecell[c]{Average of \\Adversarial} &0.7302&	0.7238&	0.8769&0.8614&	0.9724&	0.9671&	0.9959&	0.9953
 \\
    \bottomrule
    \end{tabular}
  \caption{Bit accuracy of Gaussian Shading with different factors $f_c$ and $f_{hw}$, where $l=1$.}
  \label{tab:capacity_chw_appendix}
\end{table*}

\subsection{Details of Ablation Studies}

\noindent
\textbf{Watermark capacity.} The watermark capacity is determined by three parameters: channel diffusion factor $f_c$, height-width diffusion factor $f_{hw}$, and embedding rate $l$. To investigate the impact of these hyperparameters on watermark performance, we first fix $l$ to find an optimal value for $f_c$ and $f_{hw}$. Experimental results are shown in \cref{tab:capacity_chw_appendix}. Subsequently, we fix $f_c$ and $f_{hw}$ to search for the highest possible $l$, and the corresponding experimental results are presented in \cref{tab:capacity_l_appendix}. 
\begin{table}[h]
  \centering
  \small
  \begin{tabular}{@{}ccccc@{}}
    \toprule
    \multirow{2}{*}{Noise} &
		\multicolumn{4}{c}{$l$  ($k$ bits)} \\
		\cmidrule{2-5}
		&2 (512) & 3 (768) &4 (1024)&5 (1280)\\
    \midrule
    None &0.9918&0.9502&0.8807&0.8188\\
    JPEG & 0.9112&0.8301&0.7635&0.7165 \\
    RandCr&0.7766&0.7343&0.6937&0.6586 \\
    RandDr& 0.8111&0.7545&0.7047&0.6708\\
    GauBlur& 0.8730&0.7820&0.7188&0.6783\\
    MedFilter & 0.9381&0.8534&0.7823&0.7311\\
    GauNoise & 0.8572&0.7854&0.7192&0.6750\\
    S\&PNoise & 0.8261&0.7478&0.6978&0.6546\\
    Resize & 0.9243&0.8397&0.7740&0.7188\\
    Brightness & 0.8656&0.8190&0.7480&0.7128\\
    \makecell[c]{Average of \\Adversarial} &  0.8648&0.7940&0.7332&0.6907\\
    \bottomrule
    \end{tabular}
  \caption{Bit accuracy of Gaussian Shading with different embedding rates $l$, where $f_c=1$ and $f_{hw}=8$.}
  \label{tab:capacity_l_appendix}
\end{table}

\begin{table}[h]
  \centering
  
  \small
  \begin{tabular}{@{}cccccc@{}}
    \toprule
    \multirow{2}{*}{Noise} &
		\multicolumn{5}{c}{Sampling Methods} \\
		\cmidrule{2-6}
		& \makecell[c]{DDIM \\ \cite{song2020denoising}}&  \makecell[c]{UniPC \\ \cite{zhao2023unipc}}&
  \makecell[c]{PNDM \\ \cite{liu2022pseudo}}&\makecell[c]{DEIS \\ \cite{zhang2022fast}}& \makecell[c]{DPMSolver \\ \cite{lu2022dpm}}\\
    \midrule
    None &0.9999&1.0000&1.0000&0.9999&0.9999\\
    JPEG &  0.9864&0.9797&0.9840&0.9849&0.9872\\
    RandCr& 0.9758&0.9395&0.9713&0.9507&0.9669\\
    RandDr& 0.9778&0.9642&0.9641&0.9990&0.9649\\
    GauBlur& 0.9854&0.9818&0.9886&0.9840&0.9858\\
    MedFilter & 0.9990&0.9983&0.9991&0.9991&0.9990\\
    GauNoise &0.9710&0.9264&0.9621&0.9518&0.9592\\
    S\&PNoise & 0.9302&0.9366&0.9363&0.9424&0.9385\\
    Resize & 0.9954&0.9952&0.9980&0.9977&0.9976\\
    Brightness & 0.9141&0.9431&0.9452&0.9338&0.9527\\
    \makecell[c]{Average of \\Adversarial}&  0.9706&0.9628&0.9721&0.9715&0.9724\\
    \bottomrule
    \end{tabular}

  \caption{Bit accuracy of Gaussian Shading with different sampling methods.}
  \label{tab:sampler_appendix}
\end{table}

Considering all factors, we determine that the optimal solution is $f_c=1$, $f_{hw}=8$, and $l=1$, resulting in a watermark capacity of $256$ bits.

\noindent
\textbf{Sampling methods.}
Experimental results about sampling methods under different noises are shown in \cref{tab:sampler_appendix}, and all of them exhibit excellent performance with an average bit accuracy of approximately $97\%$ against noises.

\subsection{Additional Visual Results}

\begin{figure*}[]
    \centering
    \begin{tabular}{m{1.5cm}<{\centering} m{3cm}<{\centering} m{3cm}<{\centering} m{3cm}<{\centering} m{3cm}<{\centering}}
    \toprule
       \makecell[c]{ Original} & 
        \includegraphics[width=0.16\textwidth]{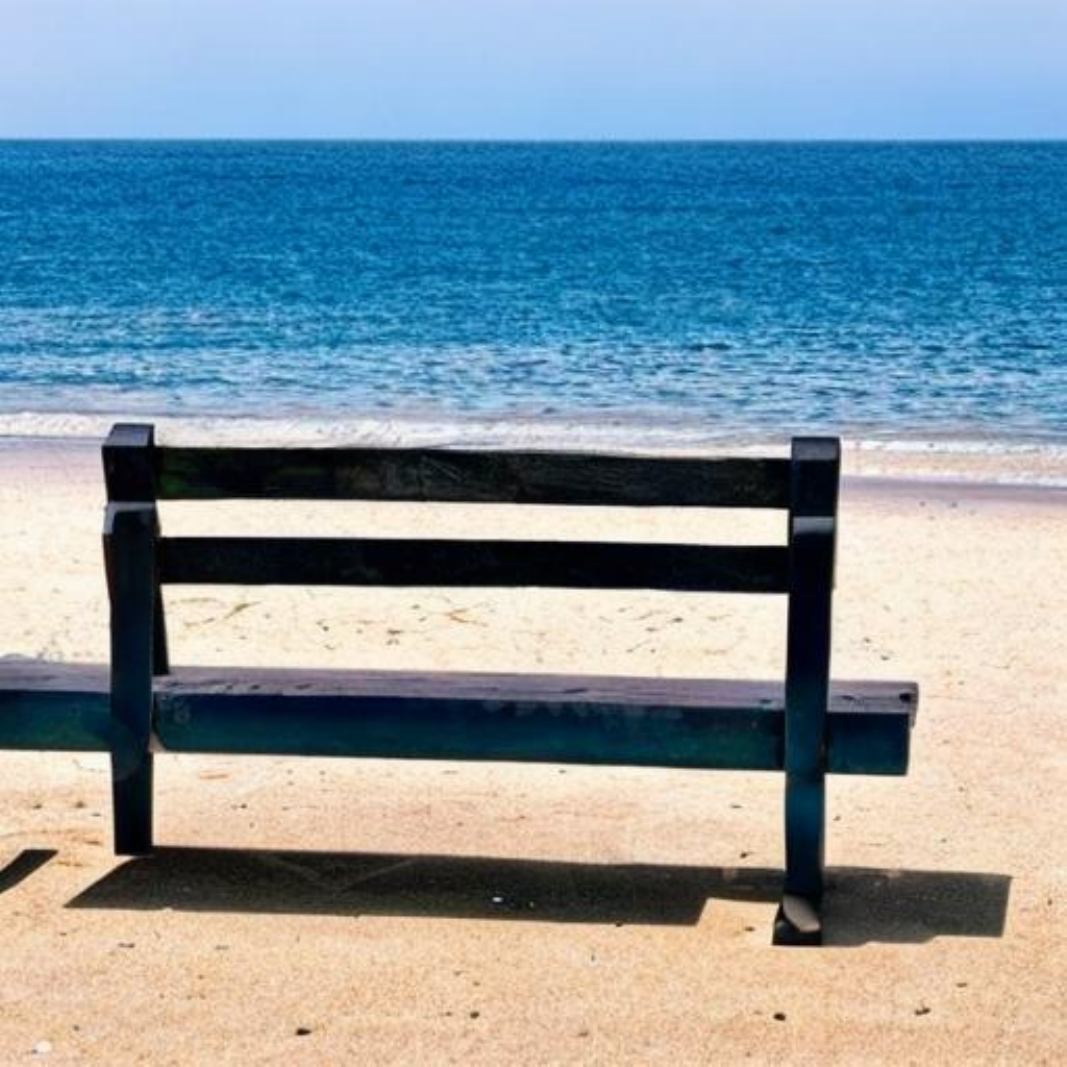} &
         &
        \includegraphics[width=0.16\textwidth]{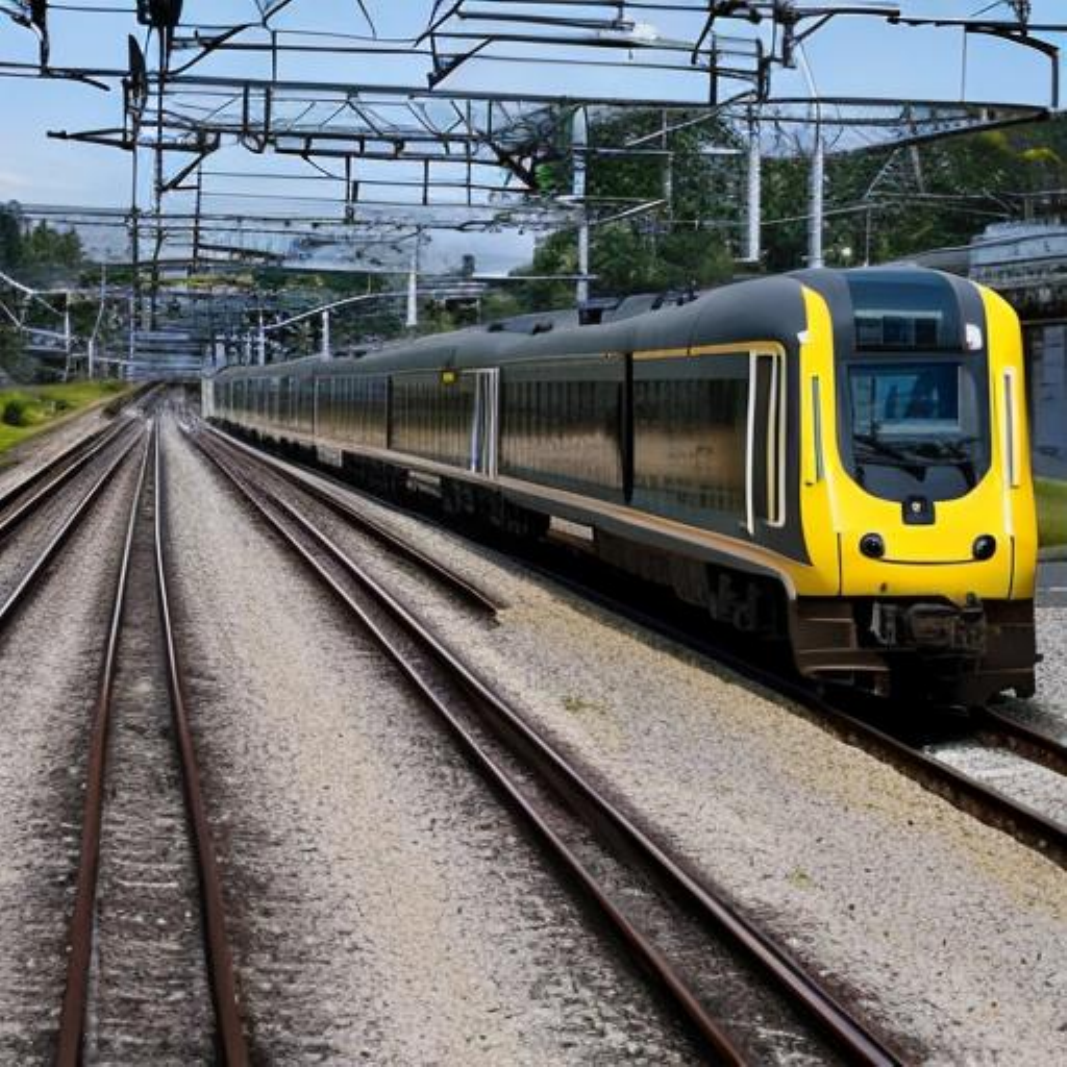} &
       \\
    \midrule
    \makecell[c]{DwtDct \\ \cite{cox2007digital}}&
        \includegraphics[width=0.16\textwidth]{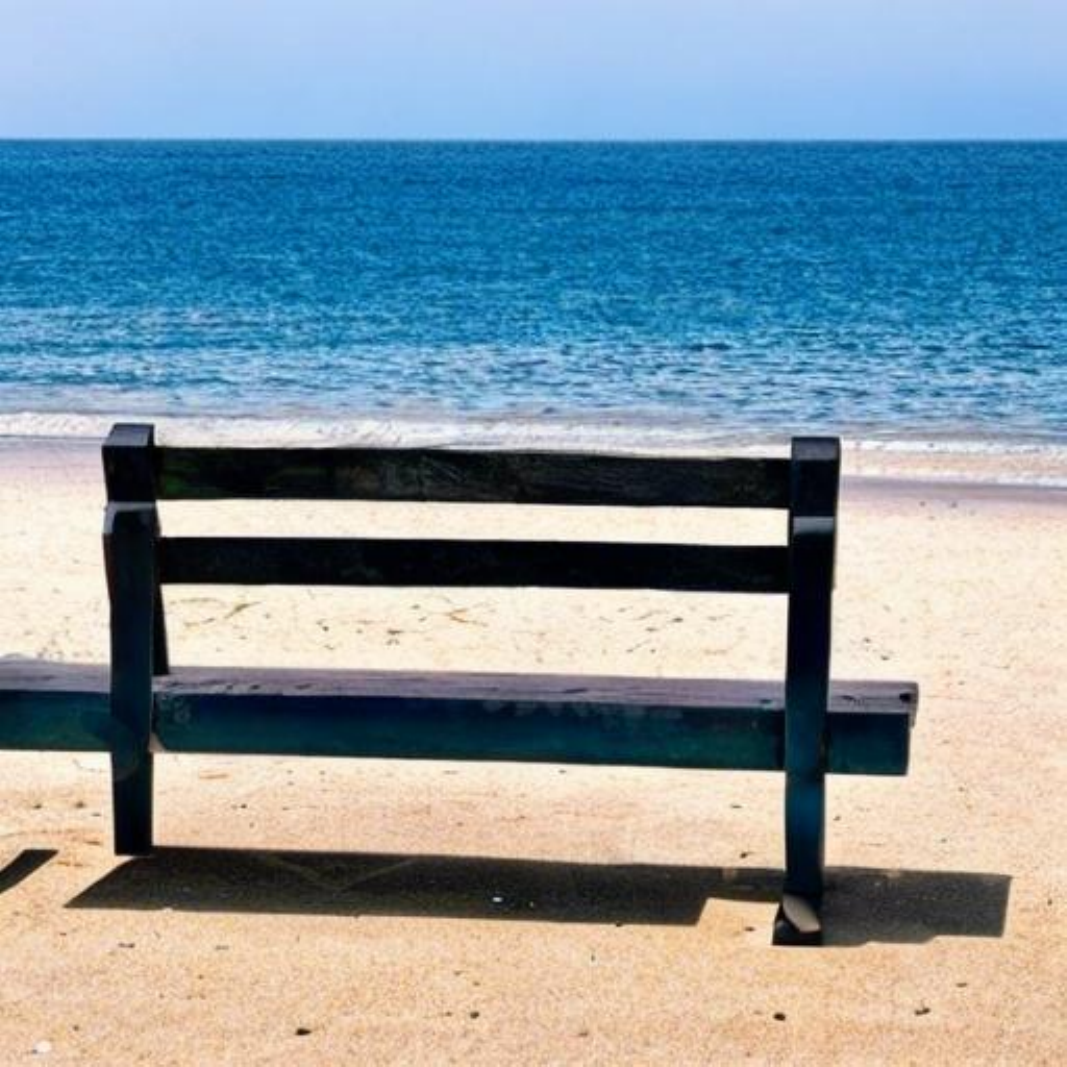} &
        \includegraphics[width=0.16\textwidth]{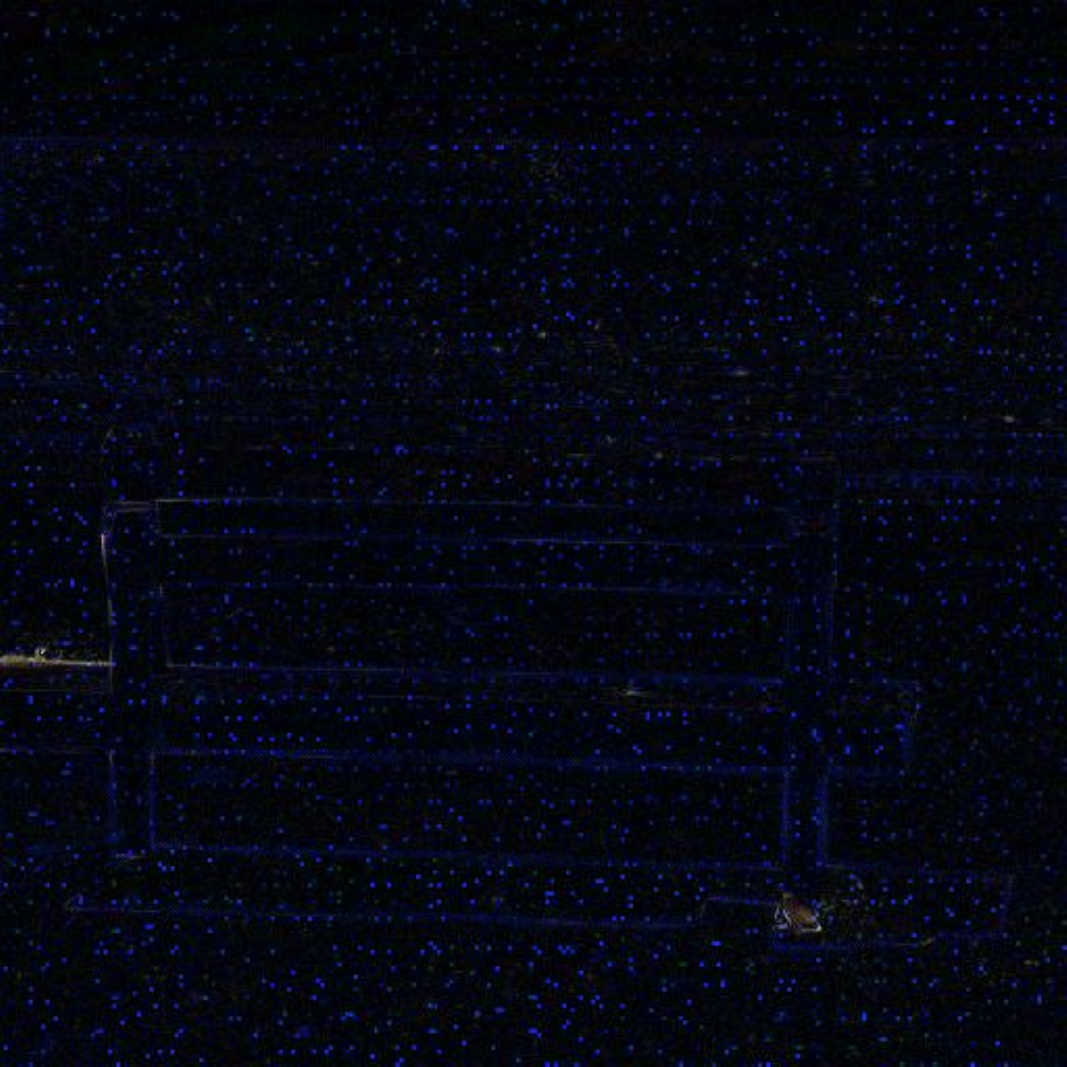} &
        \includegraphics[width=0.16\textwidth]{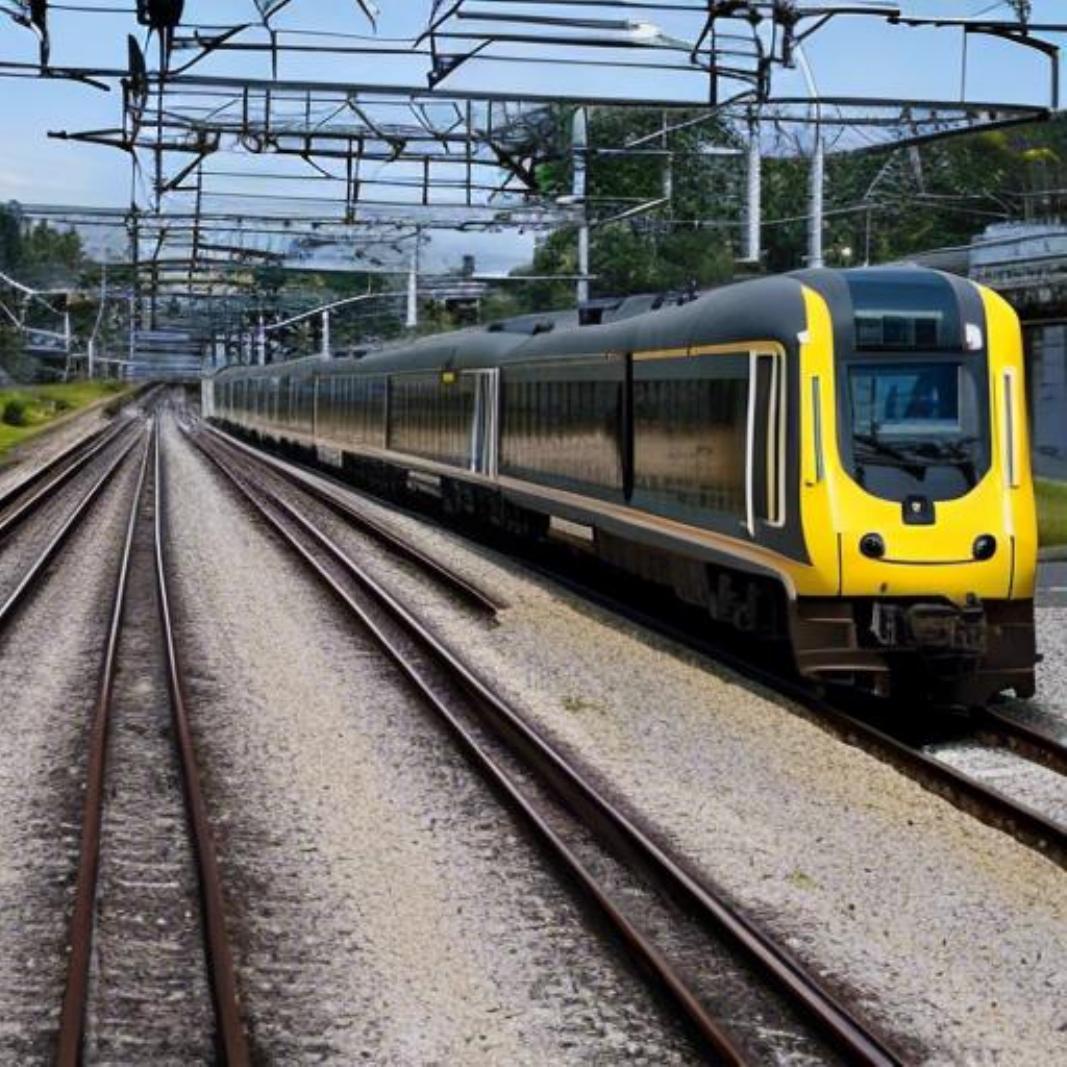} &
        \includegraphics[width=0.16\textwidth]{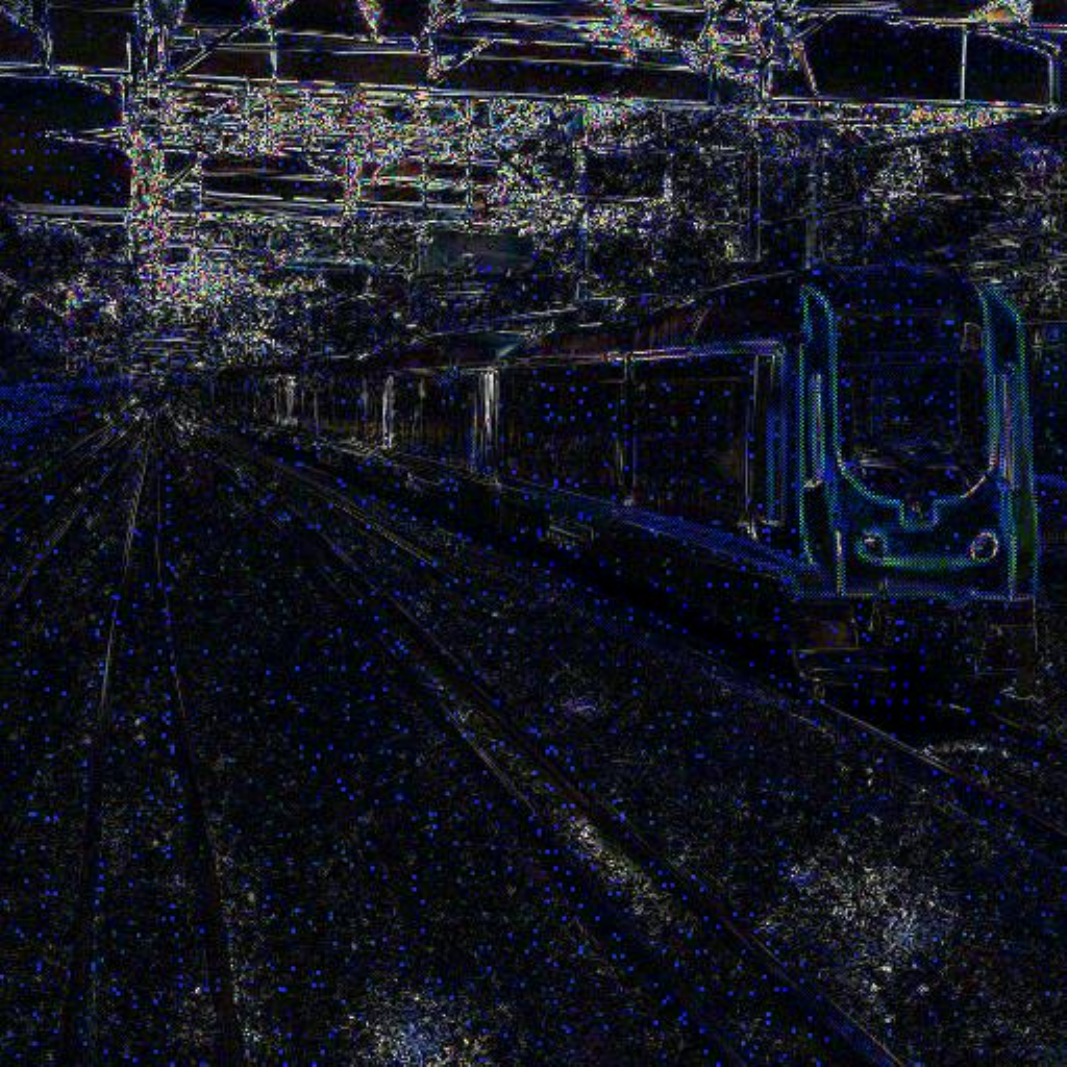}\\
    \midrule
        \makecell[c]{DwtDctSvd \\ \cite{cox2007digital}} &
        \includegraphics[width=0.16\textwidth]{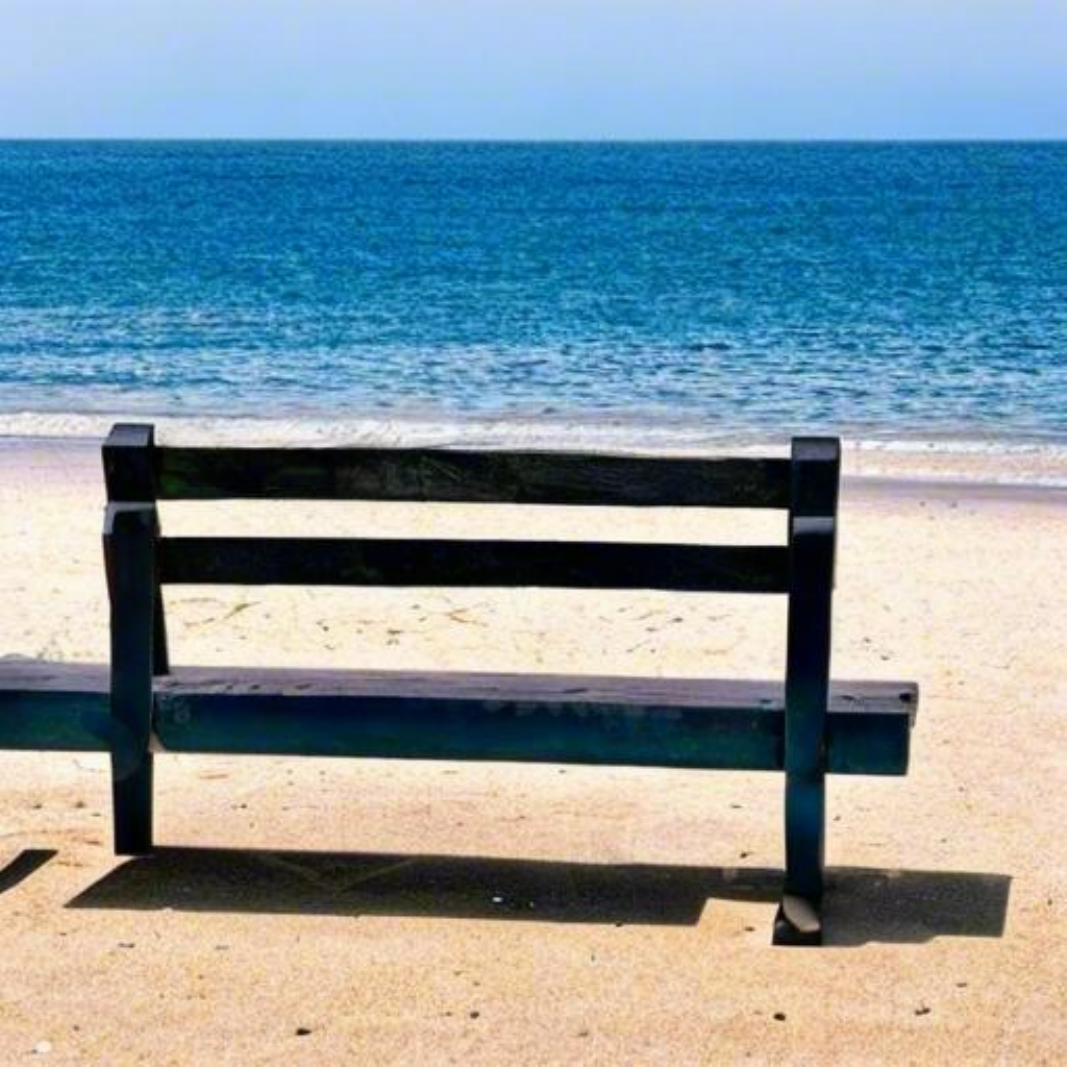} &
        \includegraphics[width=0.16\textwidth]{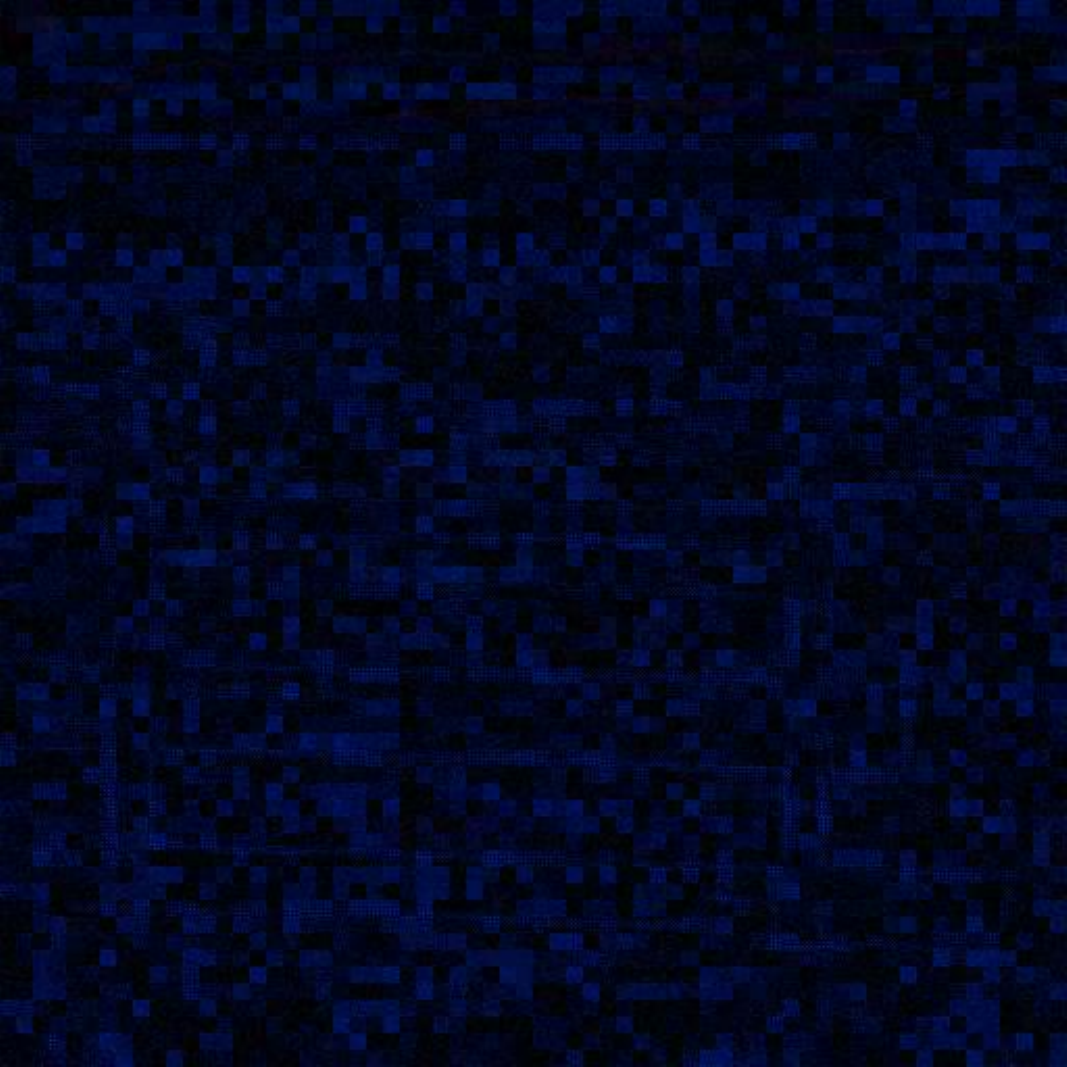} &
        \includegraphics[width=0.16\textwidth]{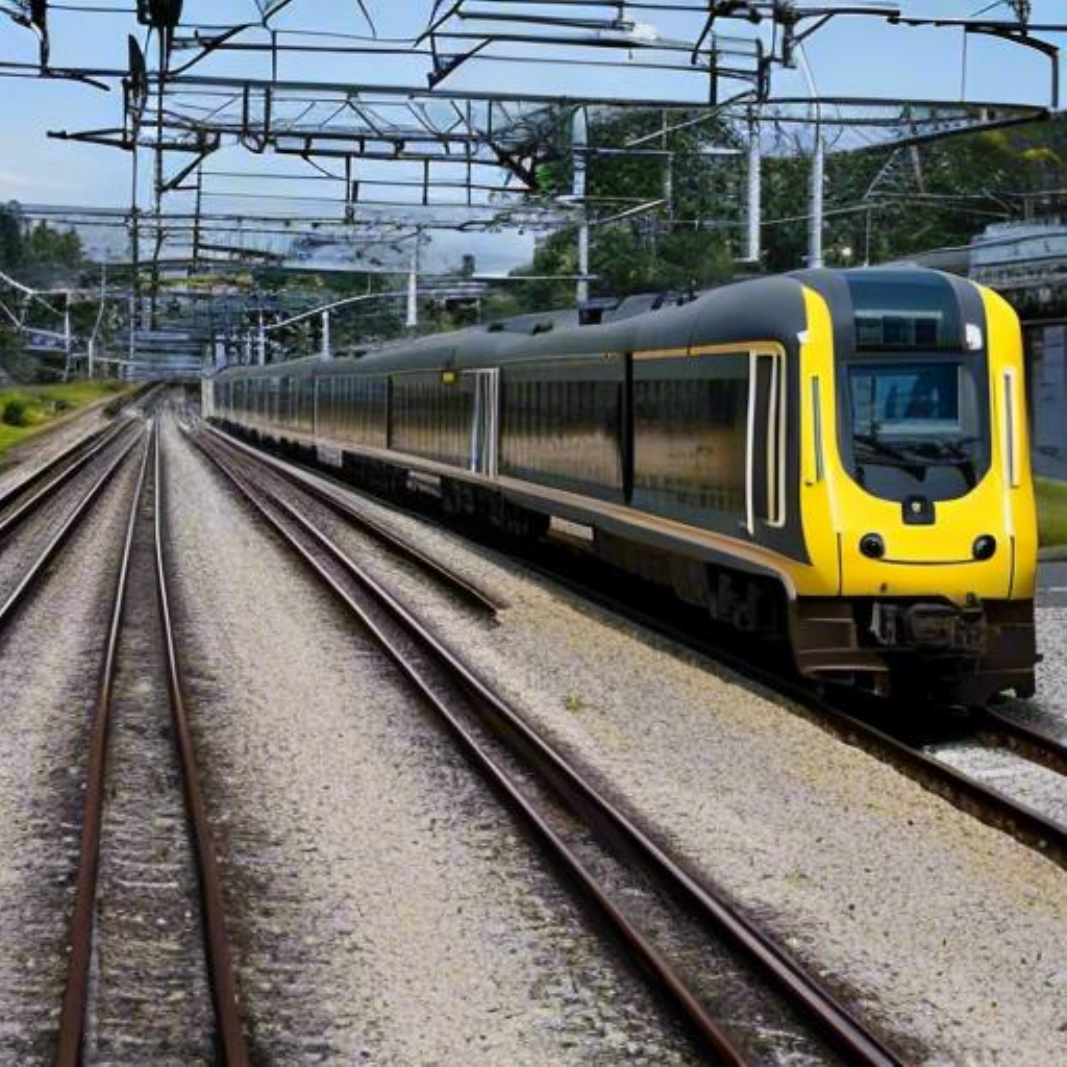} &
        \includegraphics[width=0.16\textwidth]{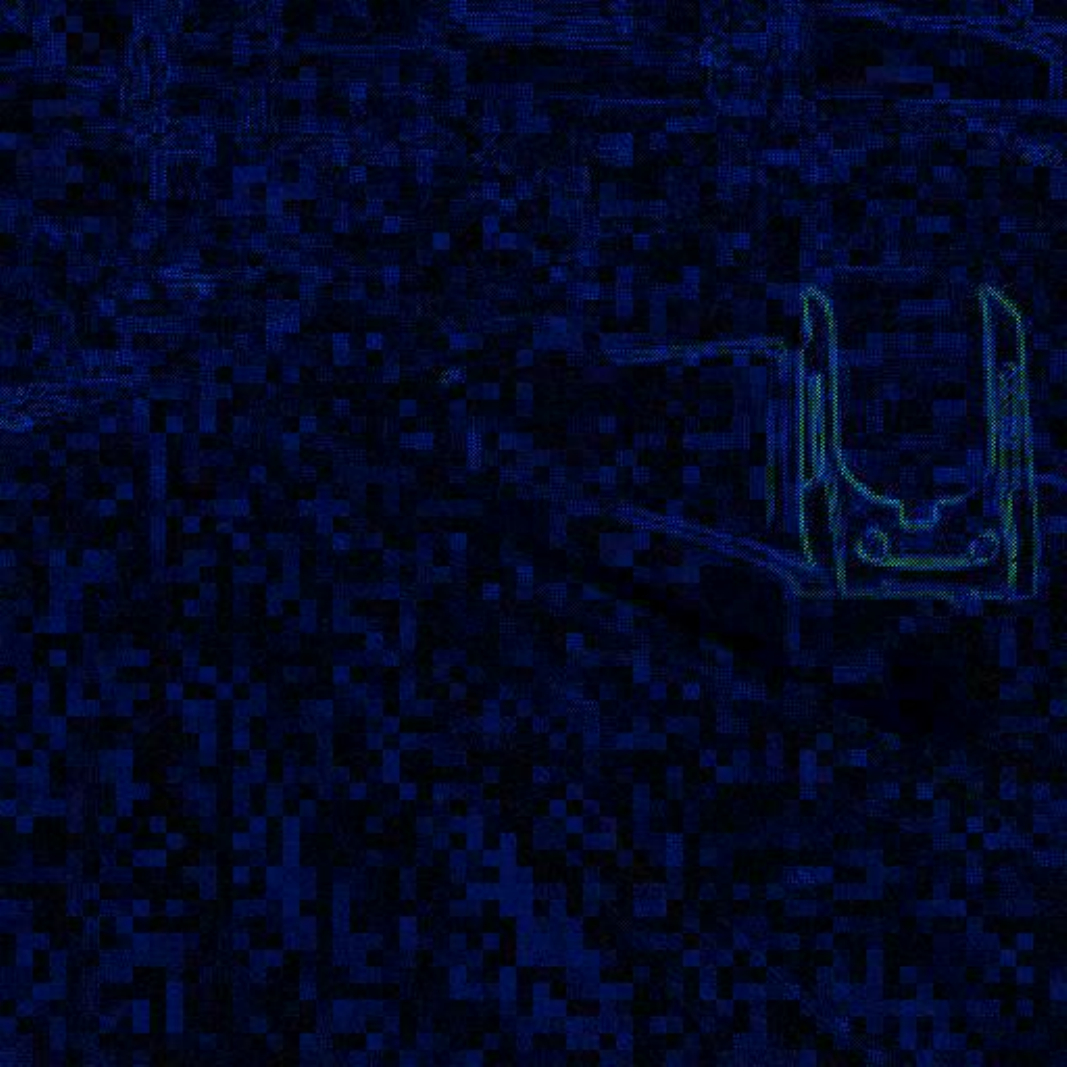}\\
    \midrule
    \makecell[c]{RivaGAN \\ \cite{zhang2019robust}}&
        \includegraphics[width=0.16\textwidth]{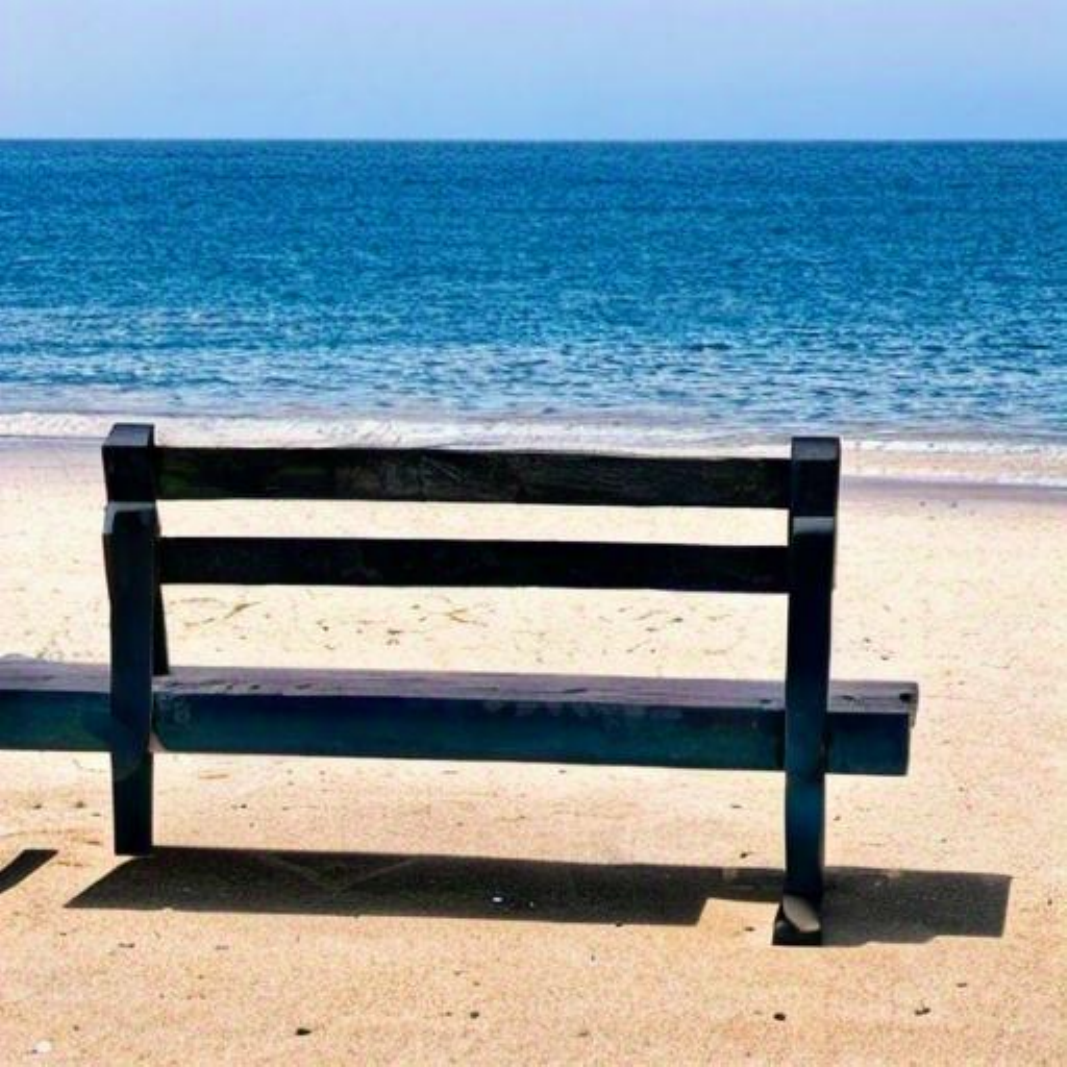} &
        \includegraphics[width=0.16\textwidth]{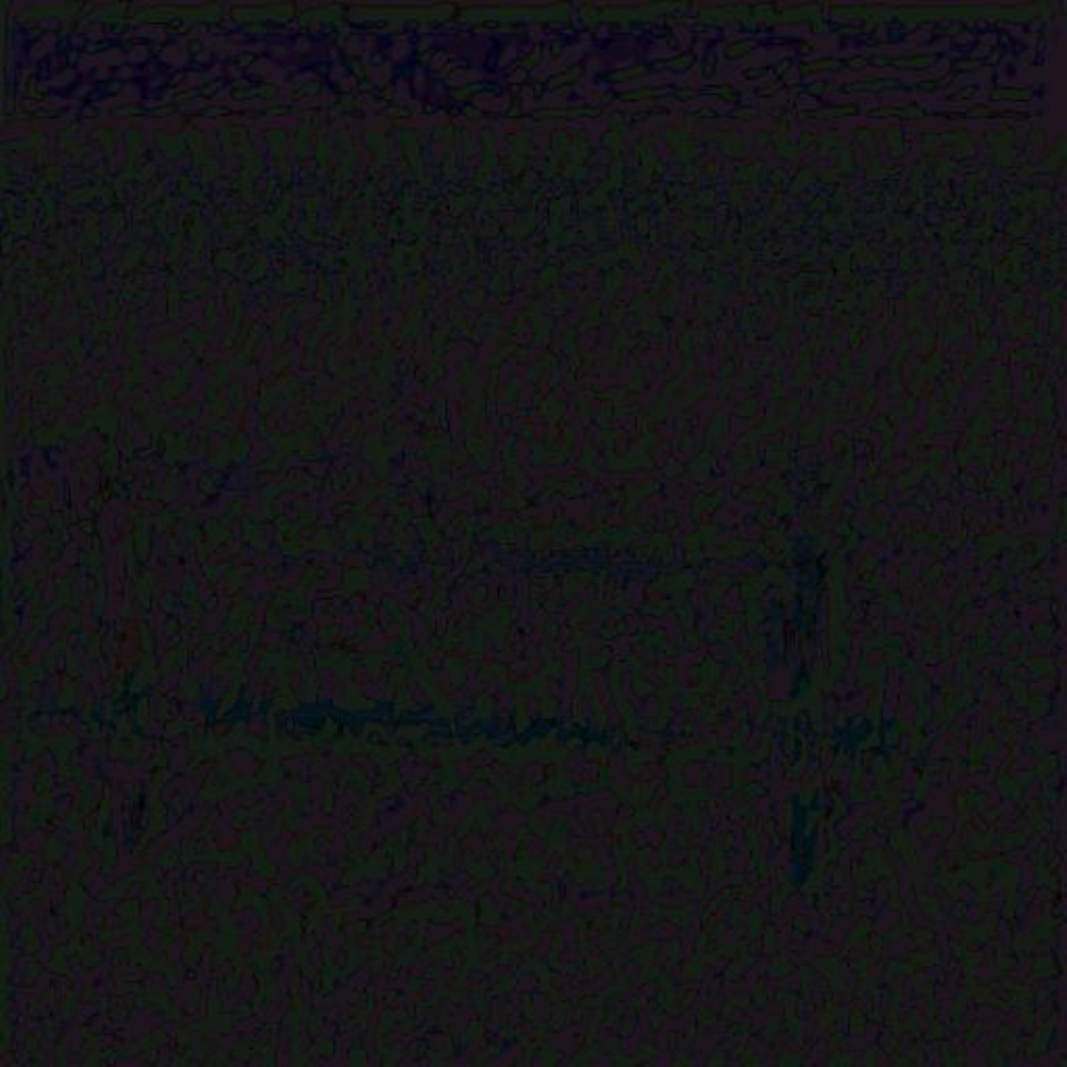} &
        \includegraphics[width=0.16\textwidth]{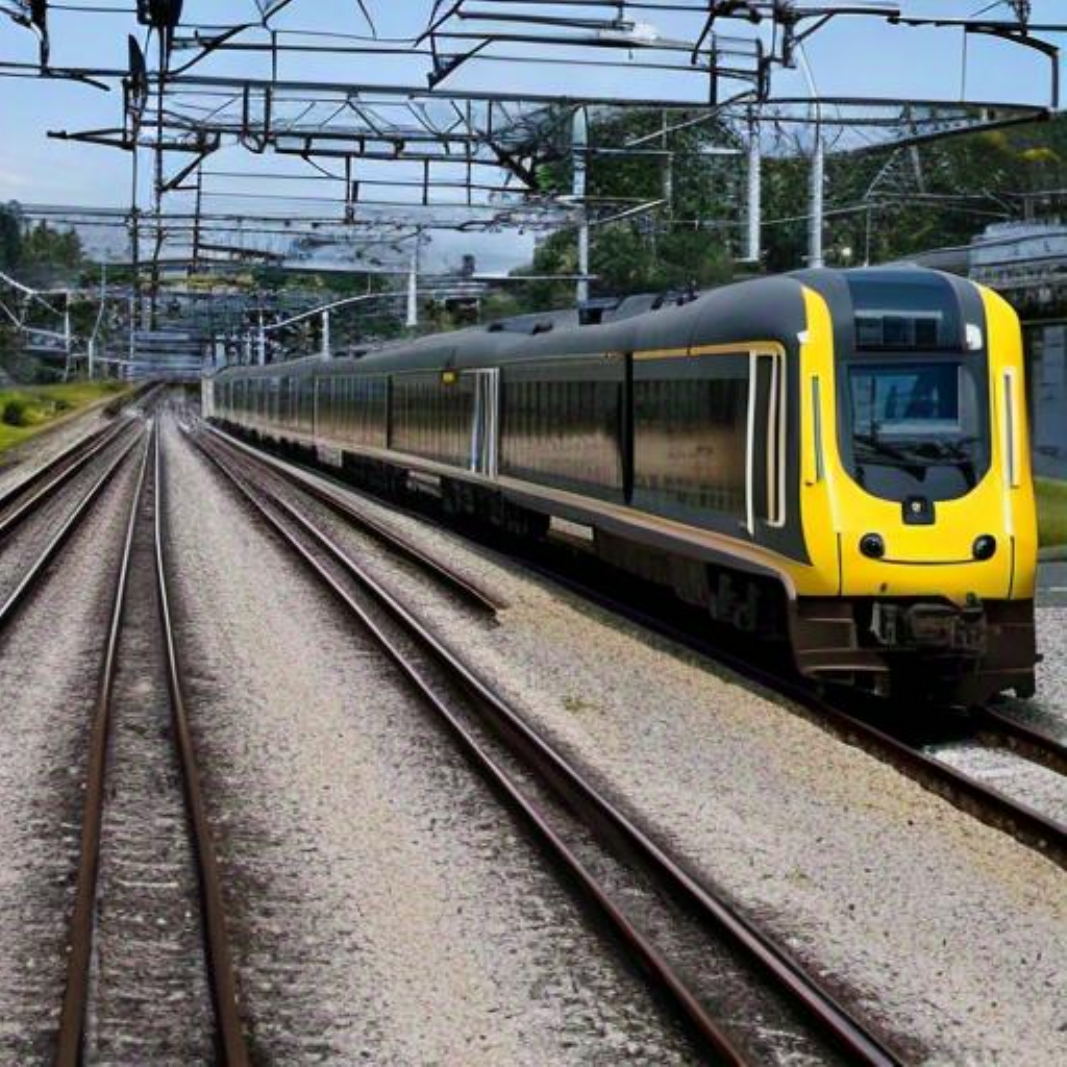} &
        \includegraphics[width=0.16\textwidth]{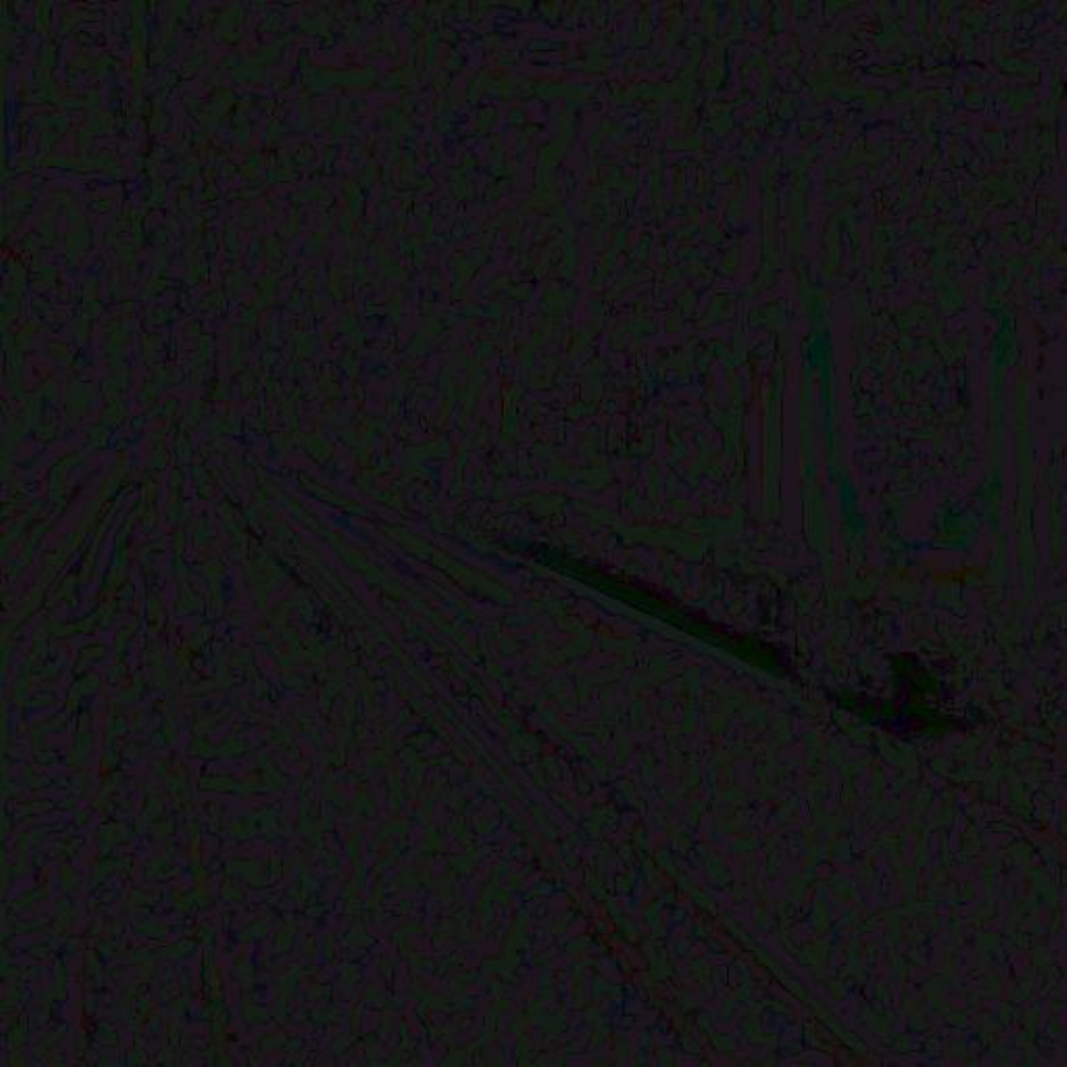}\\
    \midrule
    \makecell[c]{ Stable \\ Signature \\ \cite{fernandez2023stable}}
       &
        \includegraphics[width=0.16\textwidth]{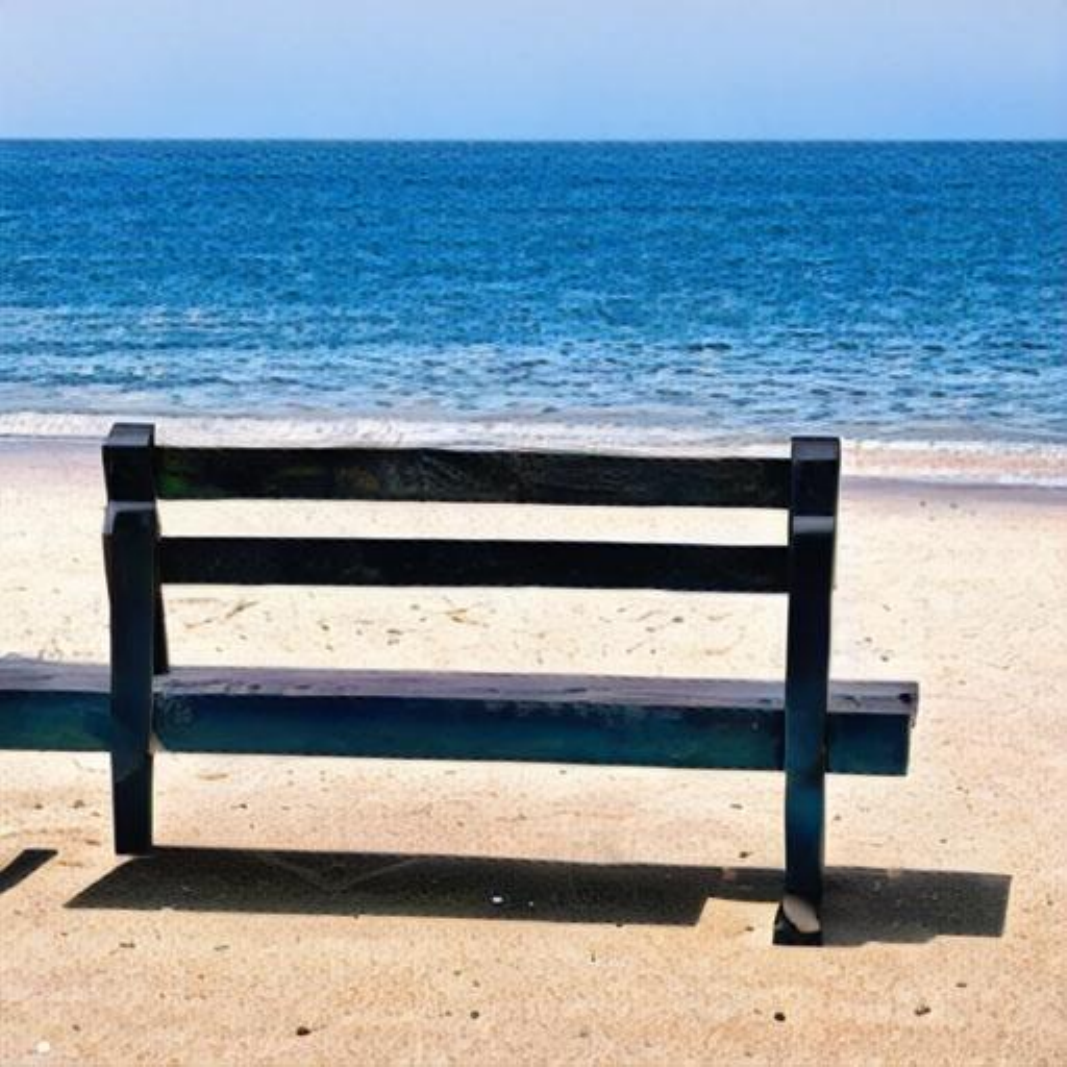} &
        \includegraphics[width=0.16\textwidth]{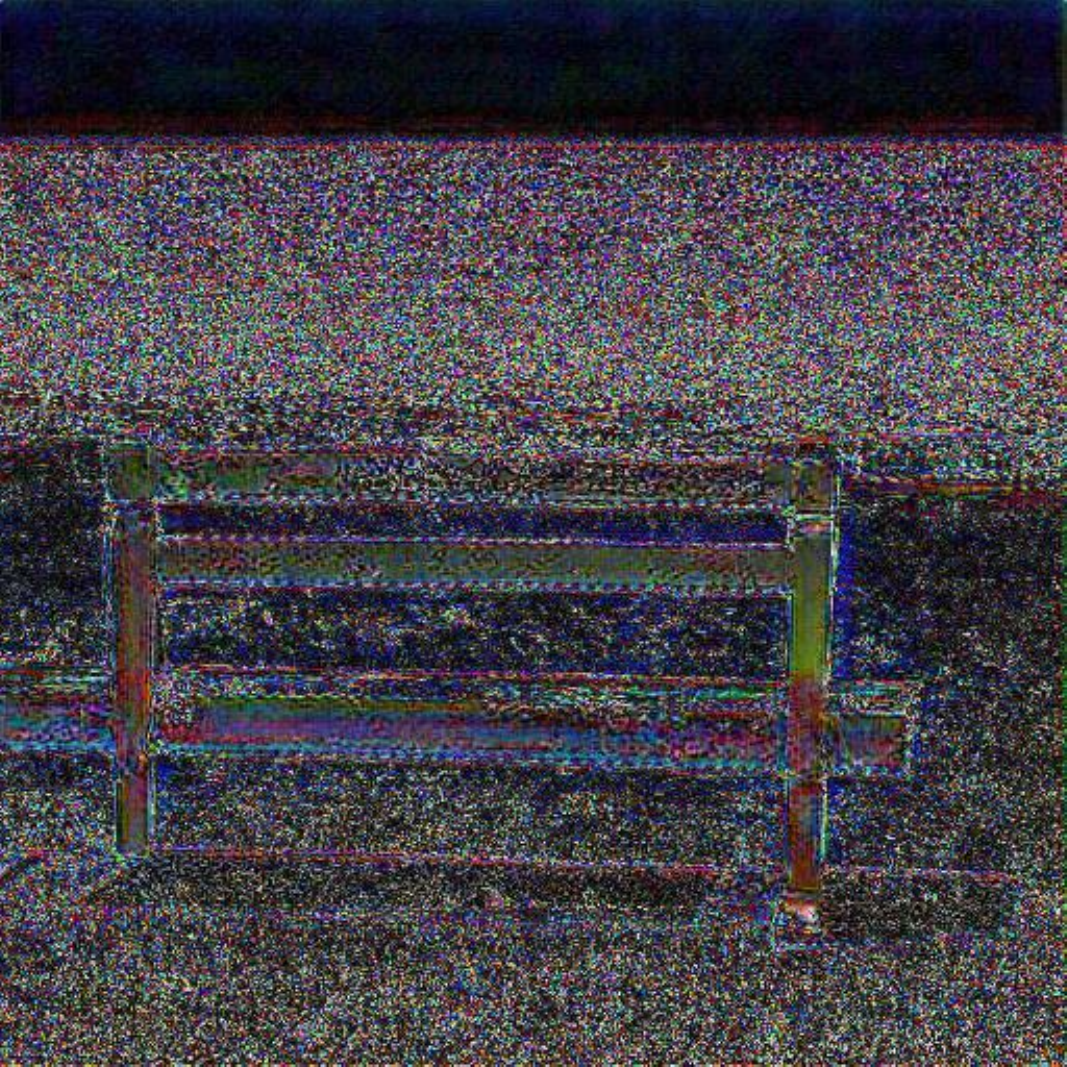} &
        \includegraphics[width=0.16\textwidth]{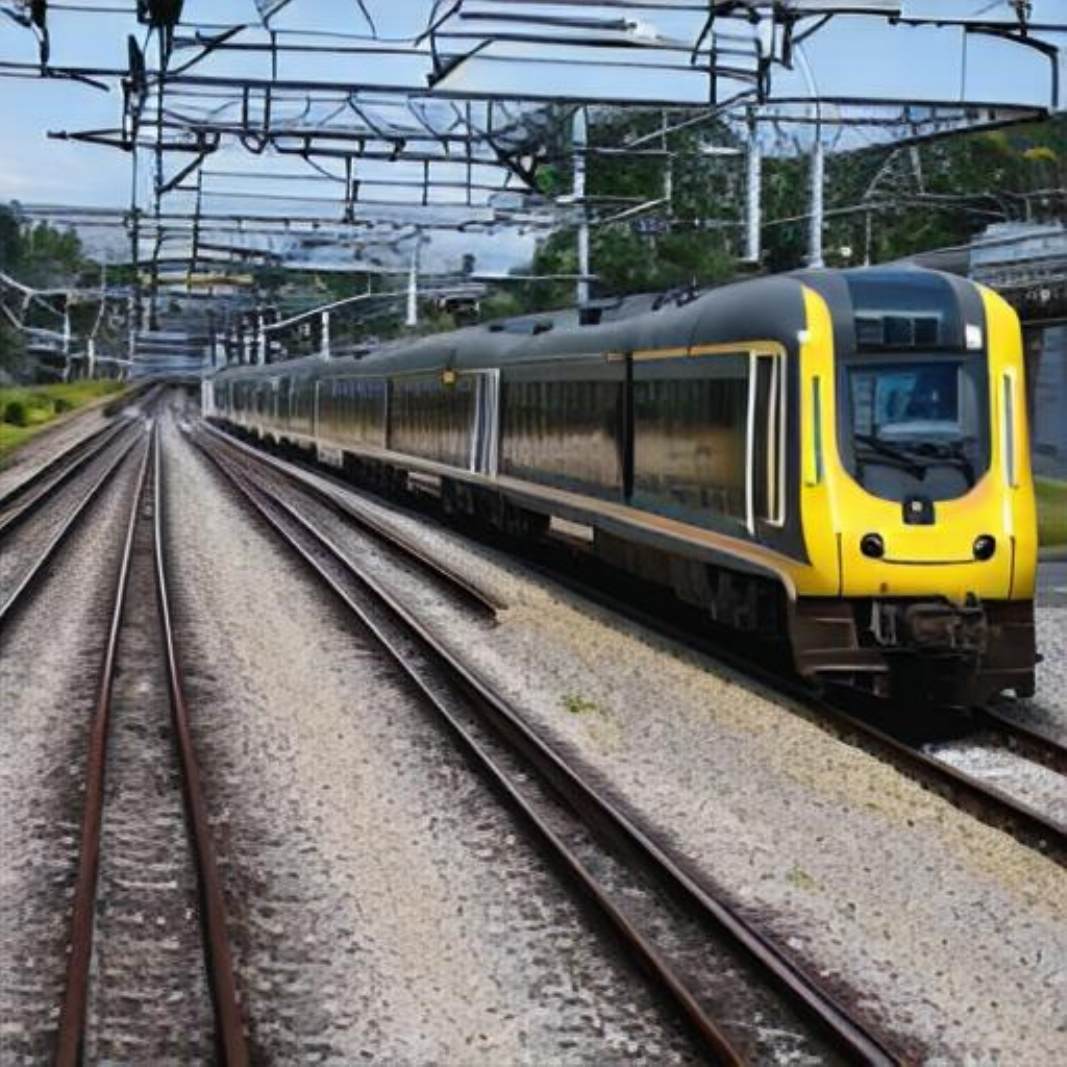} &
        \includegraphics[width=0.16\textwidth]{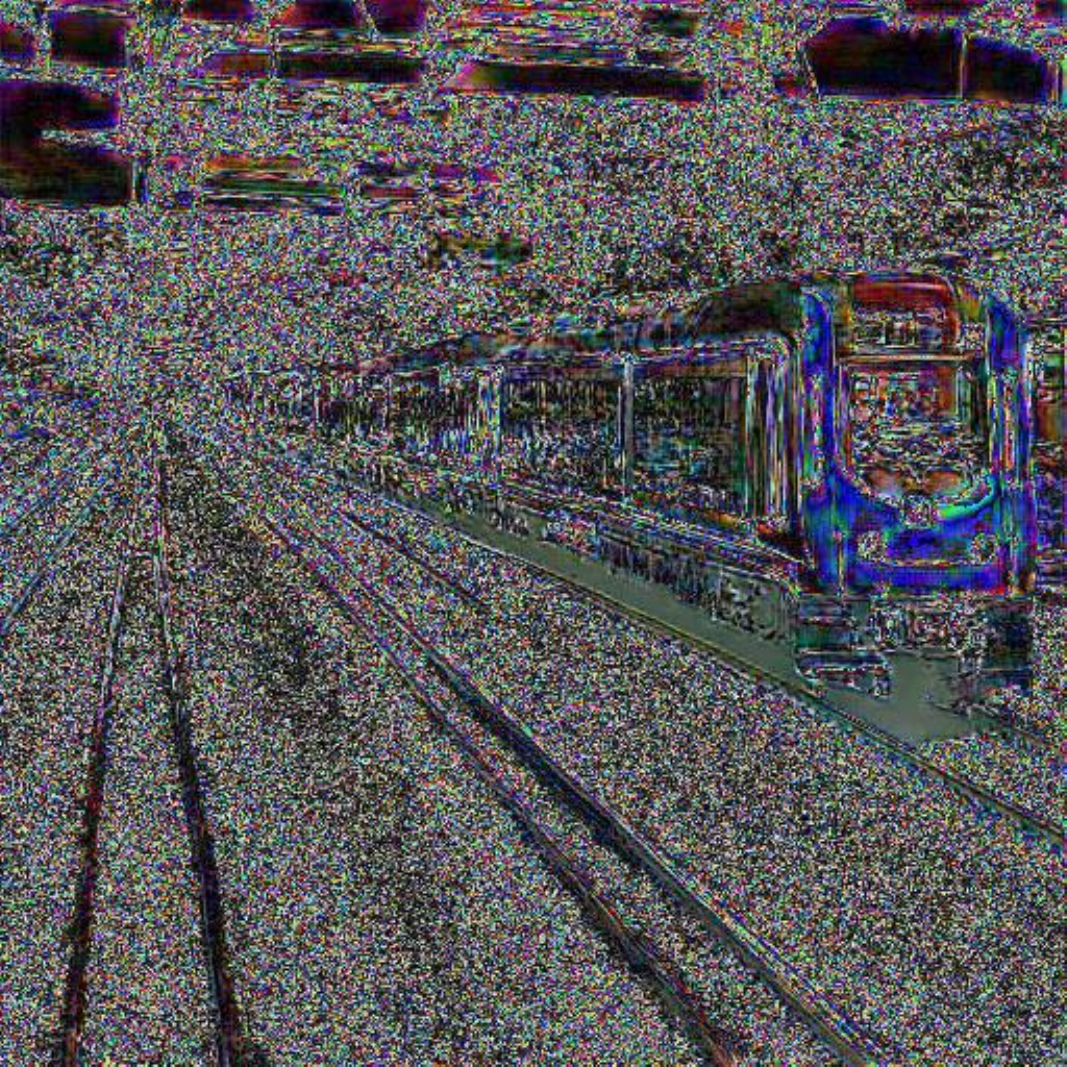}\\
         \midrule

        \makecell[c]{\textbf{Ours}} &
         \includegraphics[width=0.16\textwidth]{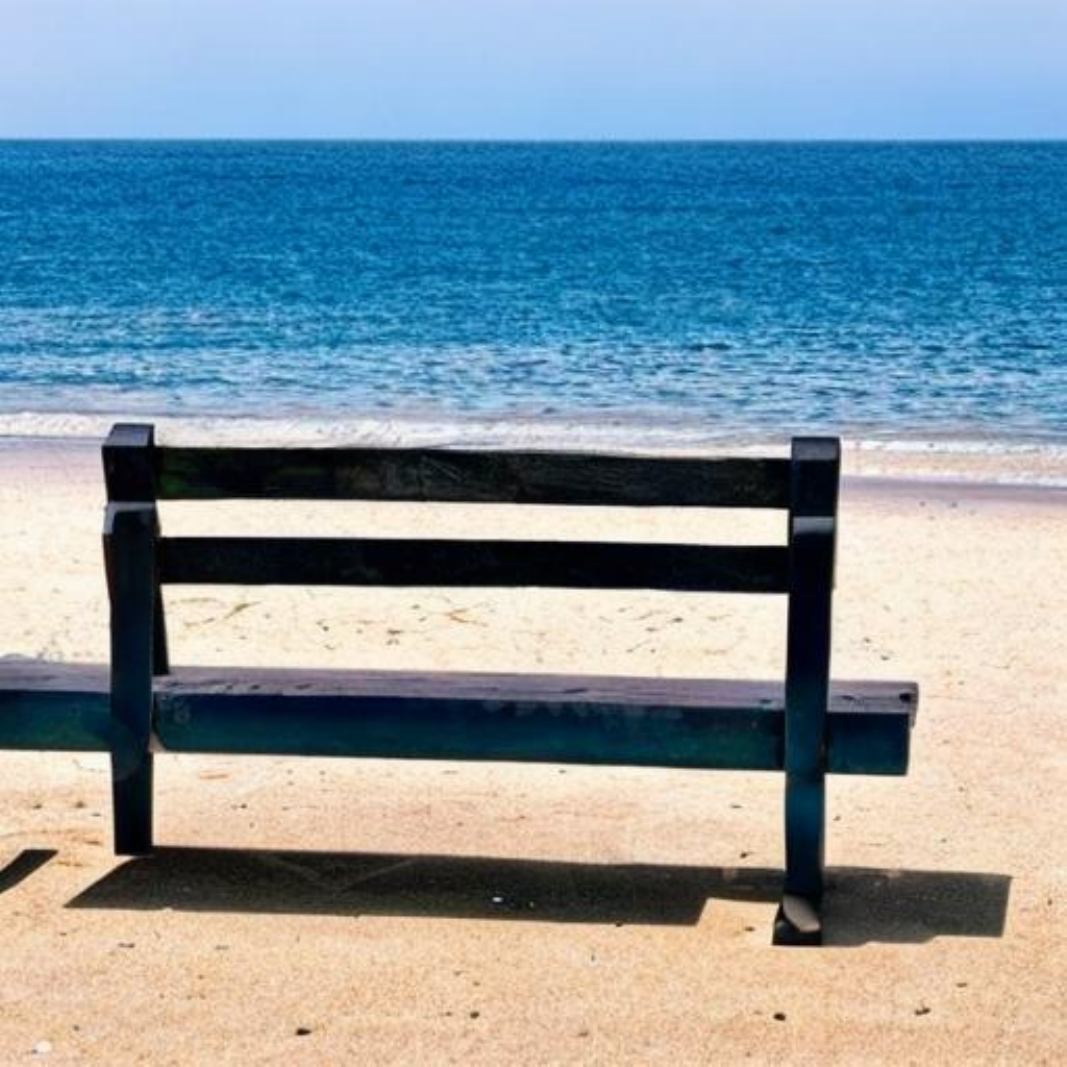} &
         
        &
        \includegraphics[width=0.16\textwidth]{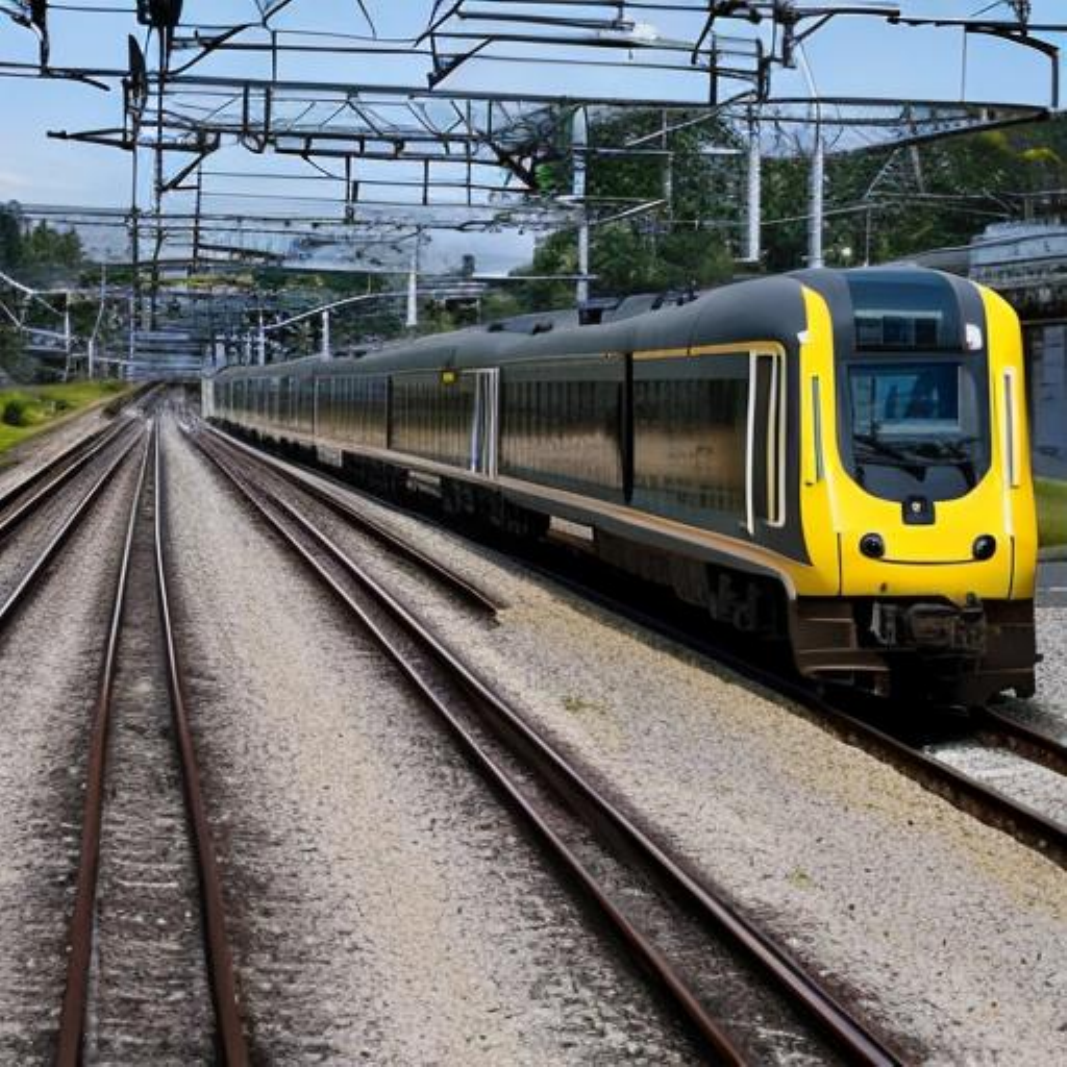} &
        
       \\
    \bottomrule
    \end{tabular}
    \caption{Additional visual results of different watermarking methods, excluding Tree-Ring, on prompts of the validation set of MS-COCO-2017, at resolution 512. All methods are applied with the same input latent representations. Comparison with Tree-Ring is on the next page.}
    \label{fig:quality_1}
\end{figure*}

\begin{figure*}[]
    \centering
    \begin{tabular}{m{3cm}<{\centering} m{4cm}<{\centering} m{4cm}<{\centering} m{4cm}<{\centering}}
    \toprule
    Prompt& Original& Tree-Ring~\cite{wen2023tree}& \textbf{Ours}\\
    \midrule
       
        \makecell[c]{A bird is  sitting \\on a bowl\\ of birdseed.}&\includegraphics[width=0.22\textwidth]{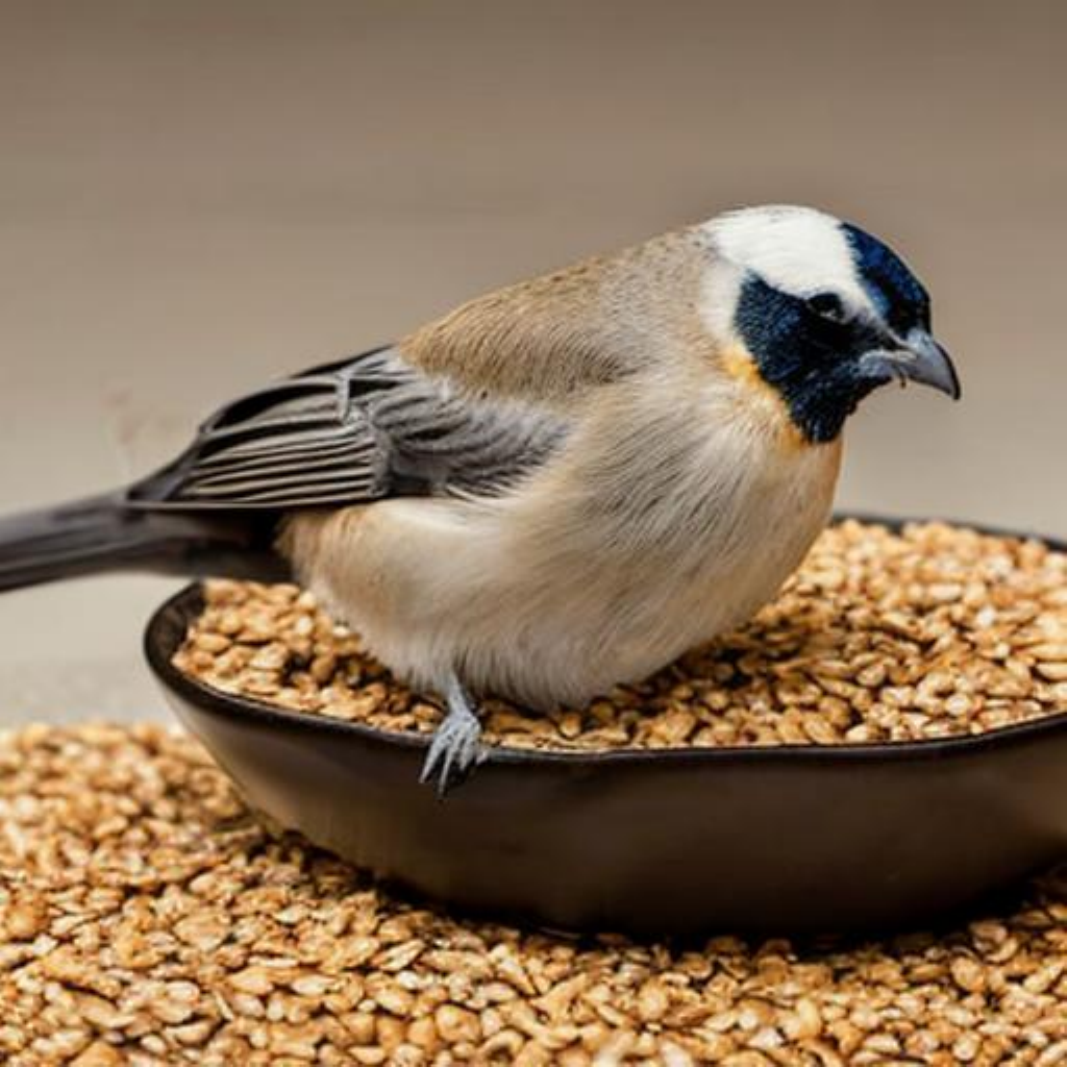} &
         \includegraphics[width=0.22\textwidth]{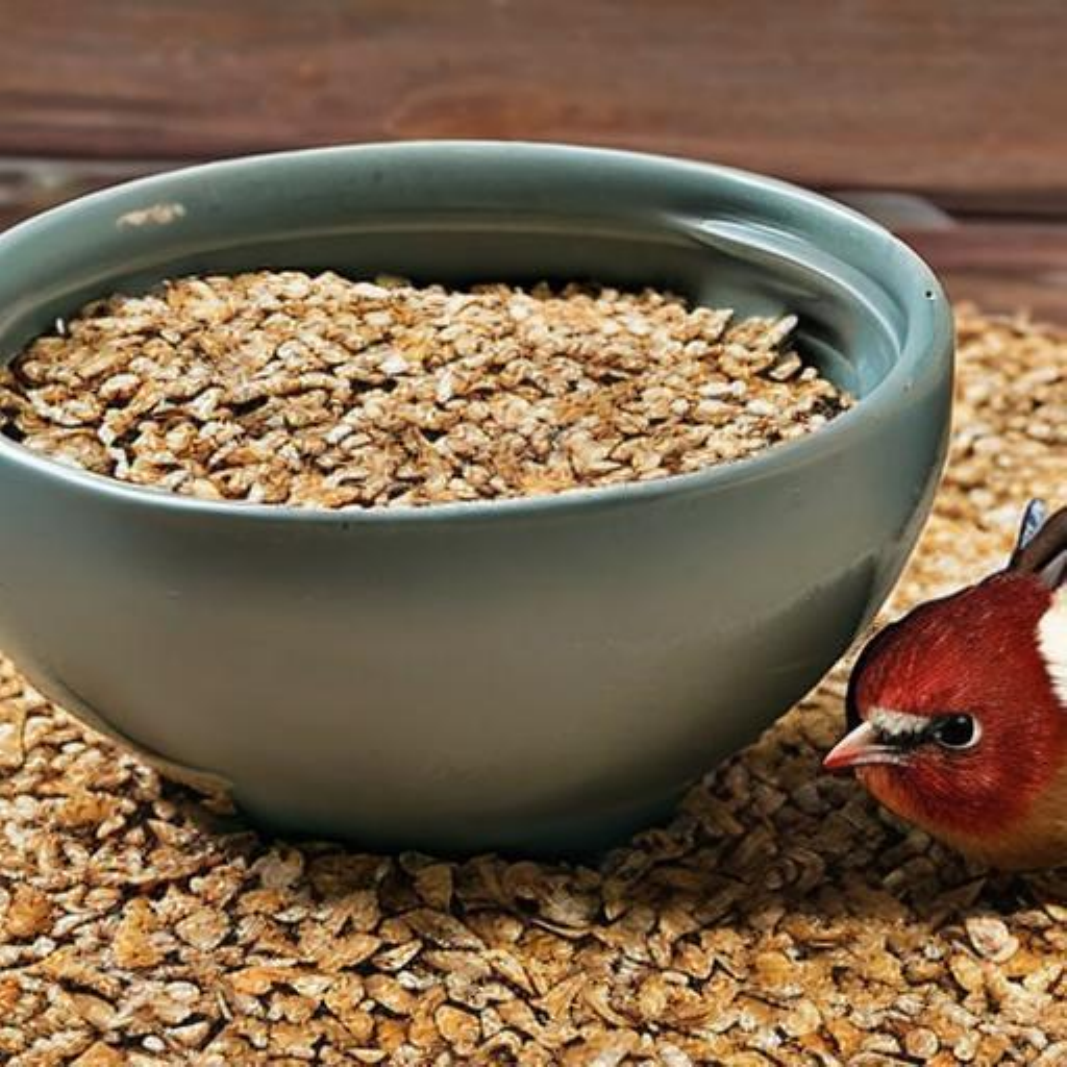}&
        \includegraphics[width=0.22\textwidth]{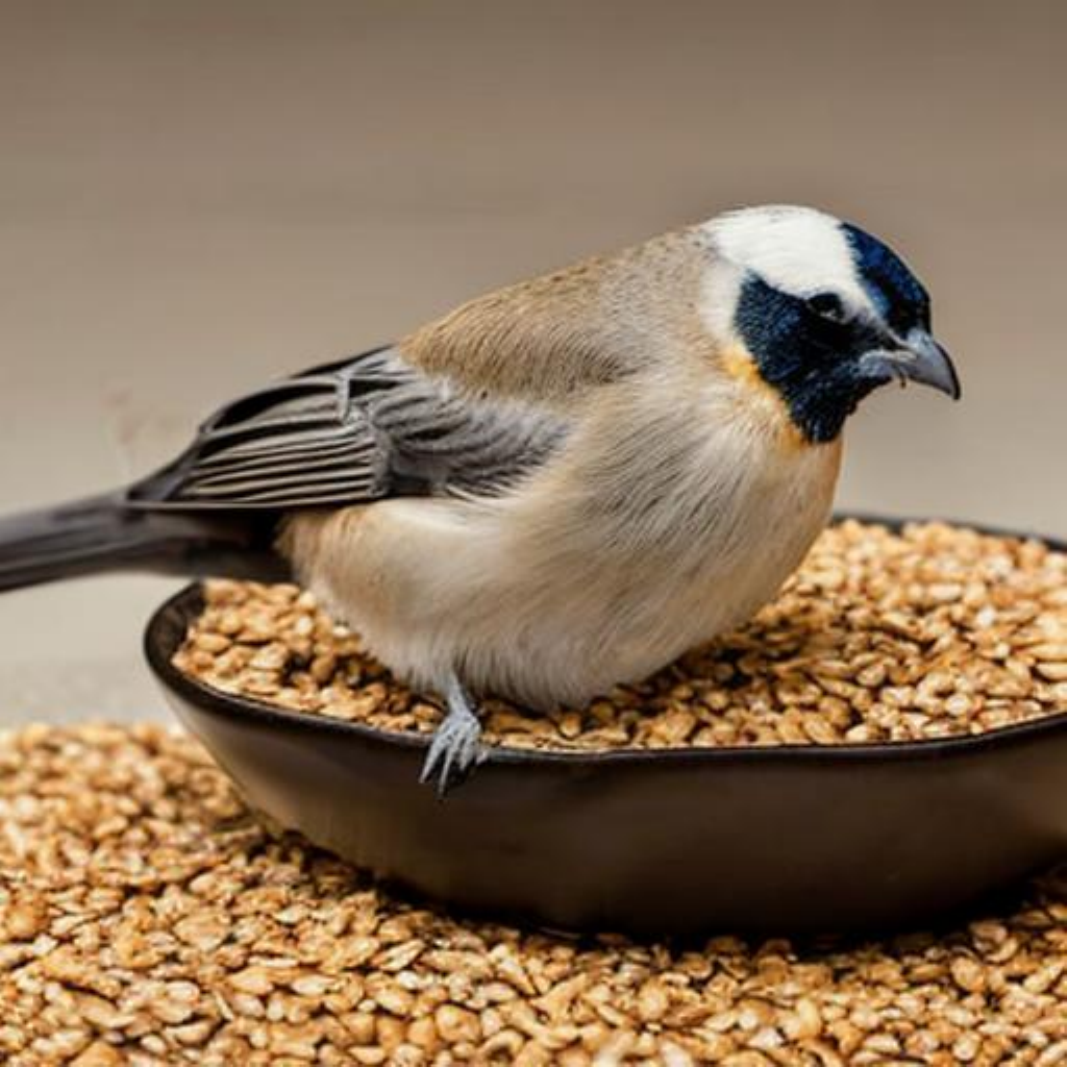} 
       \\
    \midrule
        \makecell[c]{ A man holding \\open an oven door\\ in a kitchen.}&
        \includegraphics[width=0.22\textwidth]{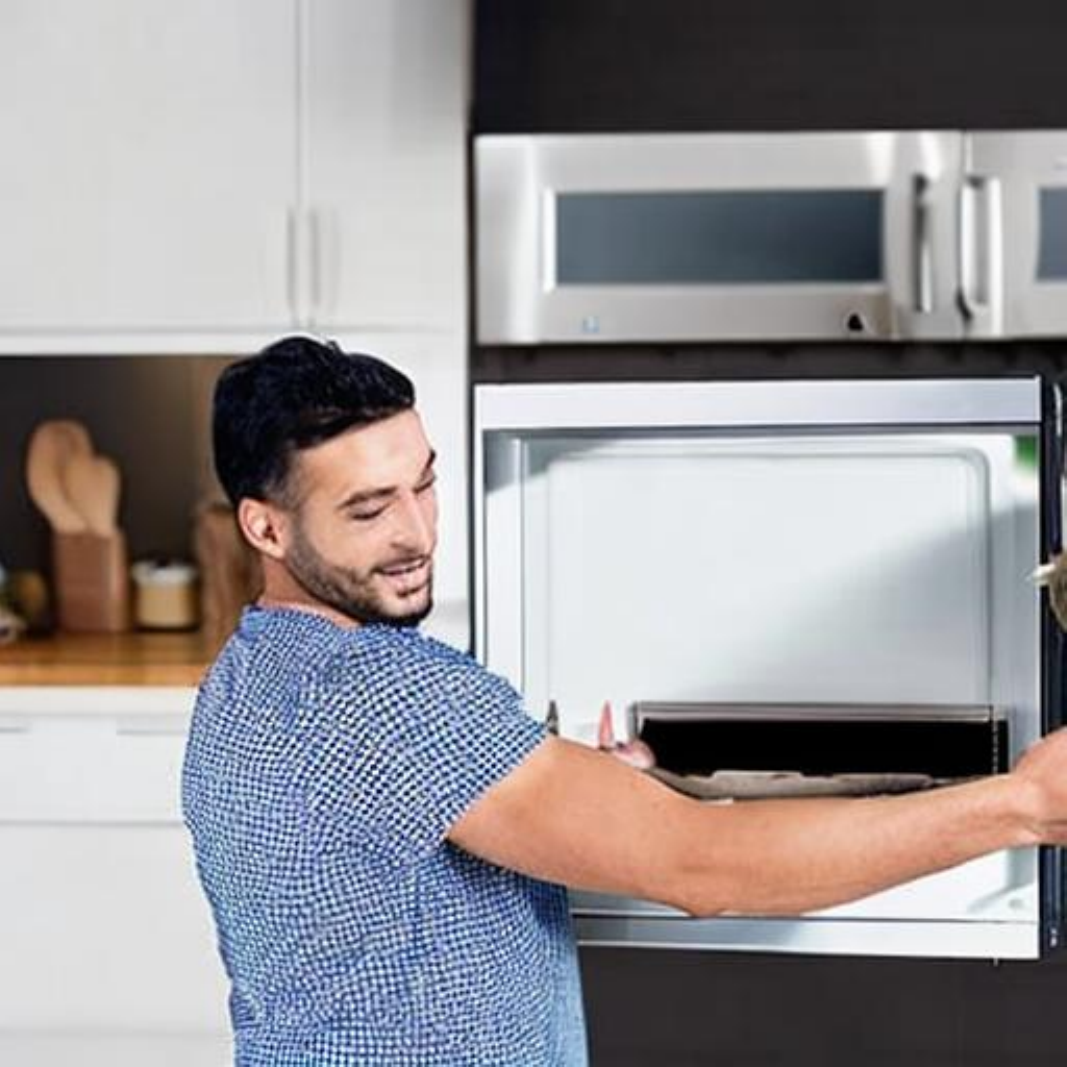}&
        \includegraphics[width=0.22\textwidth]{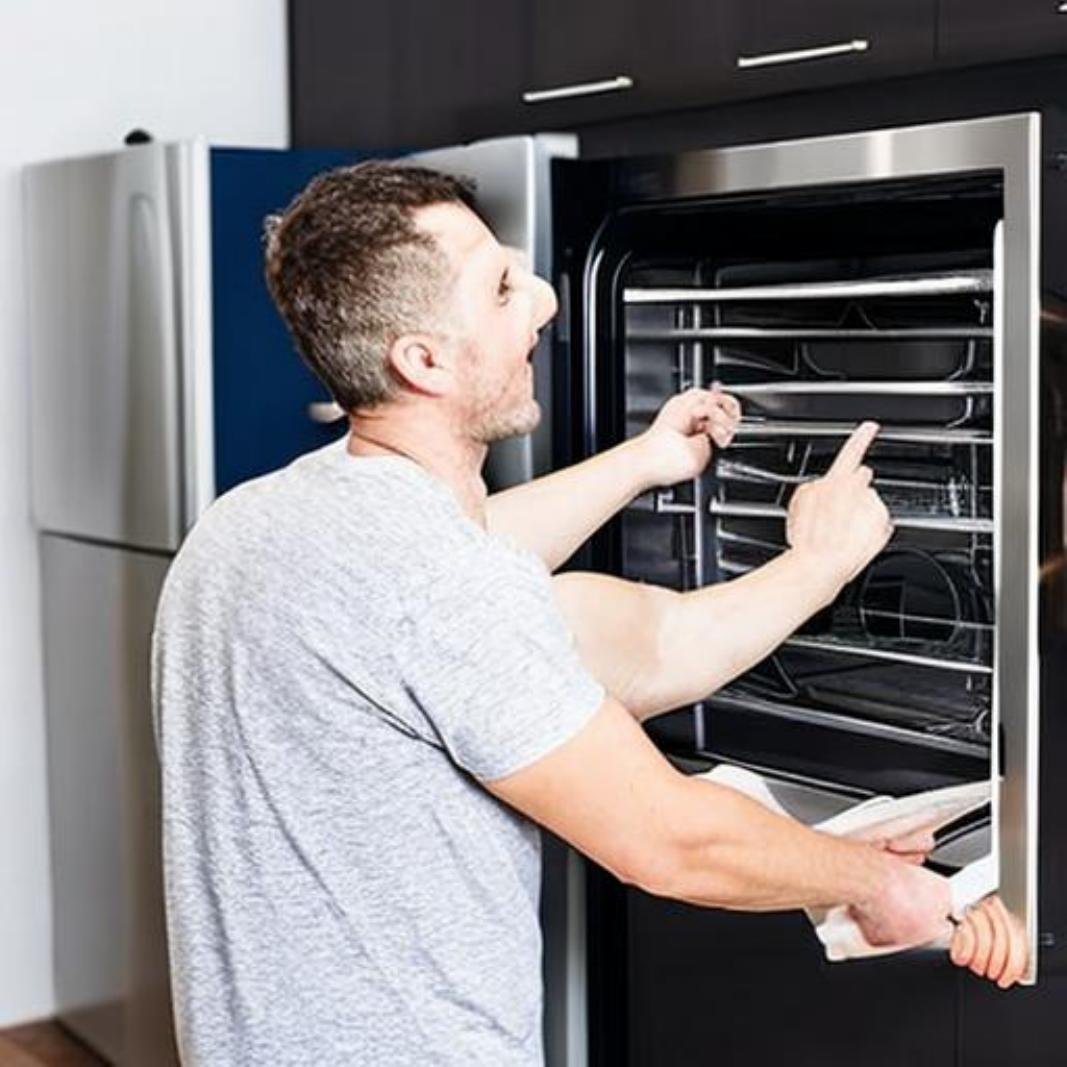}&
        \includegraphics[width=0.22\textwidth]{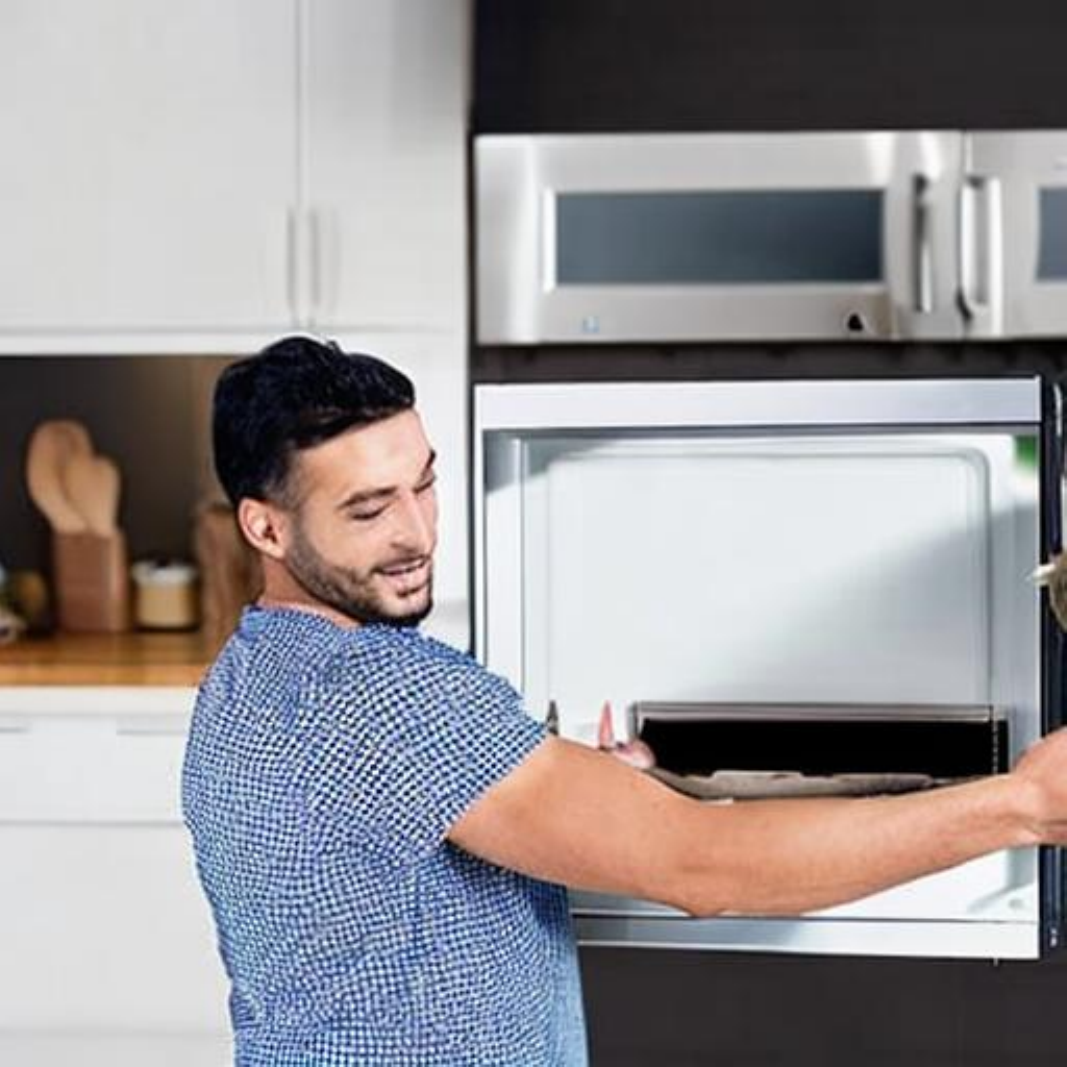}\\
    \midrule
        \makecell[c]{ A skillet on \\a stove with \\vegetables in it.}&
        \includegraphics[width=0.22\textwidth]{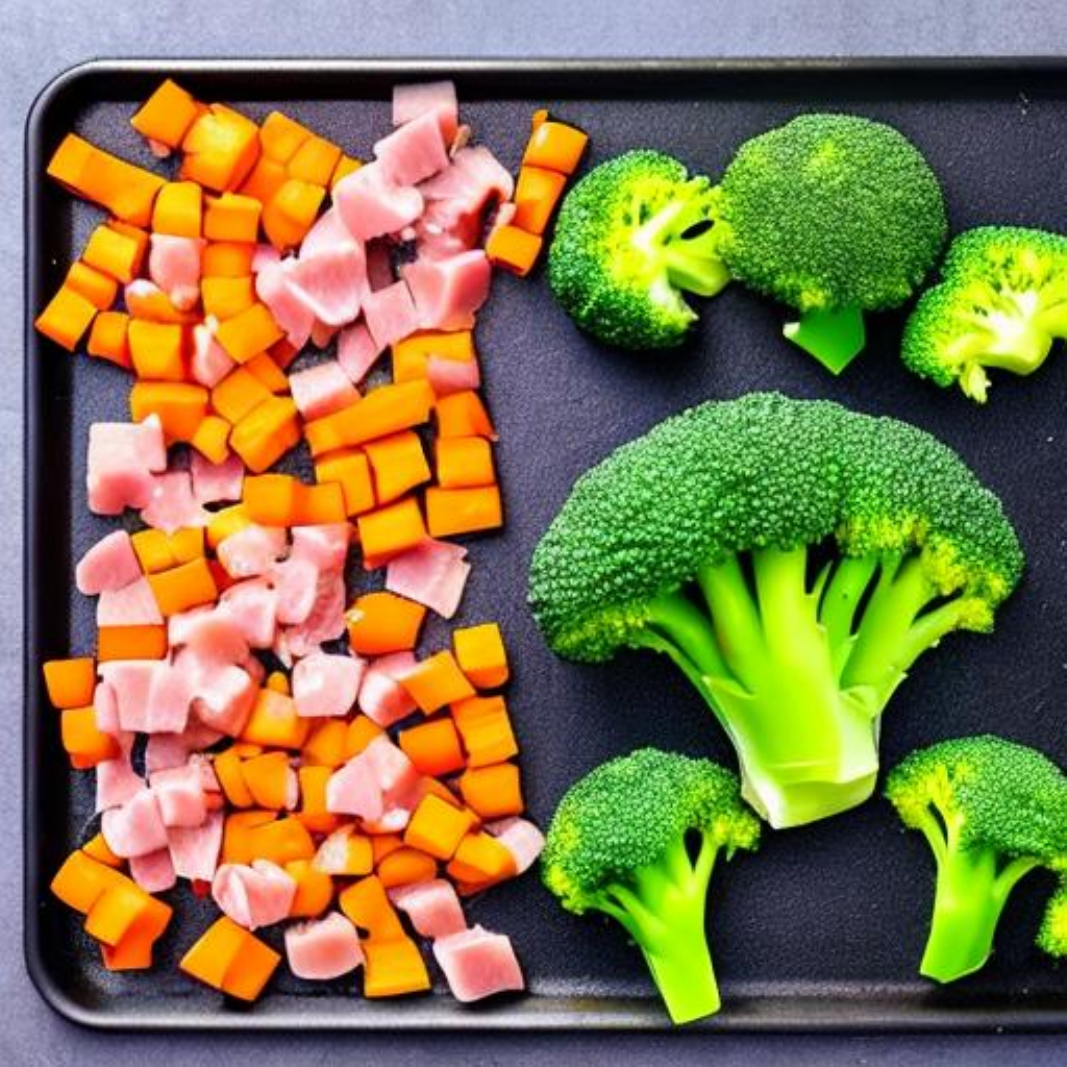}&
        \includegraphics[width=0.22\textwidth]{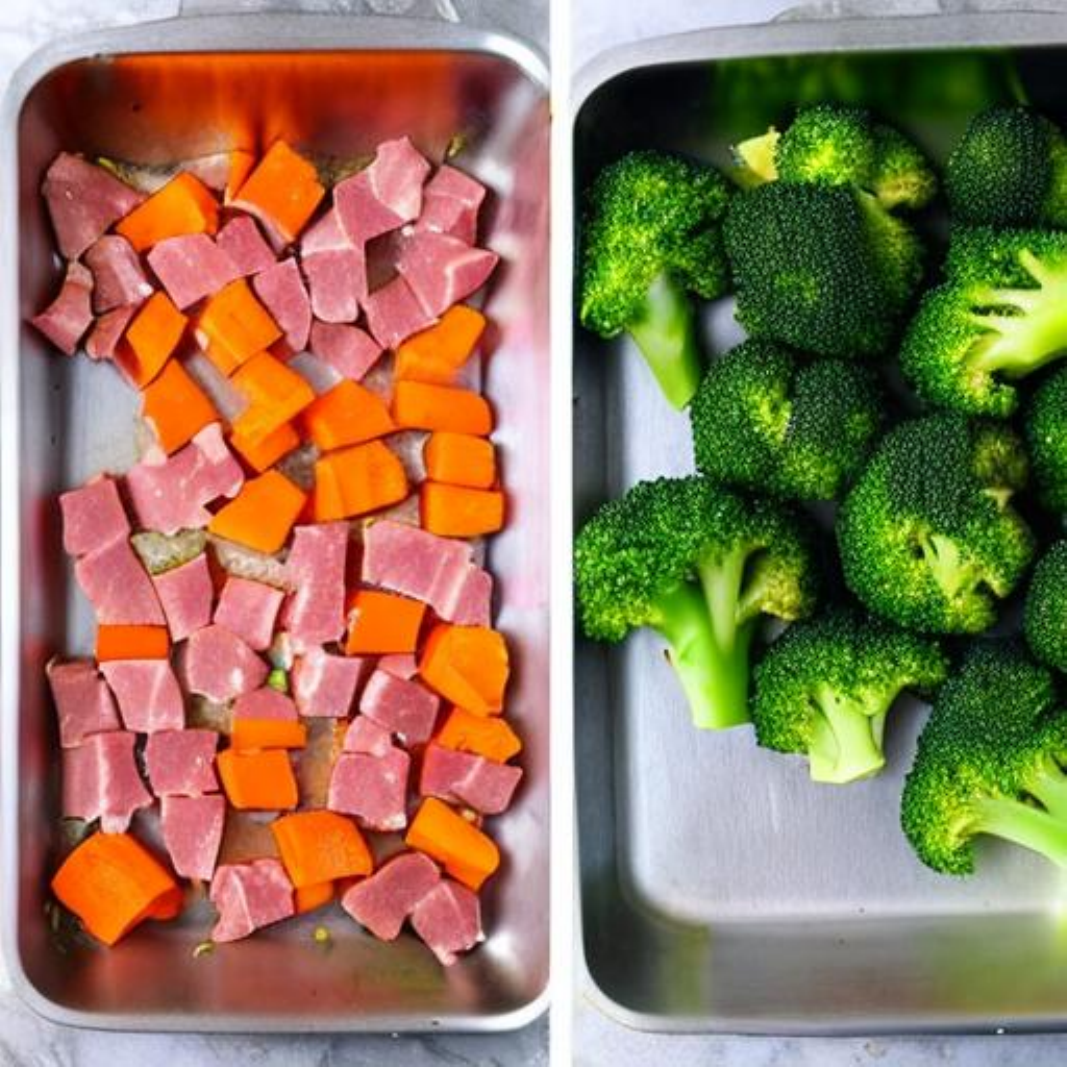}&
        \includegraphics[width=0.22\textwidth]{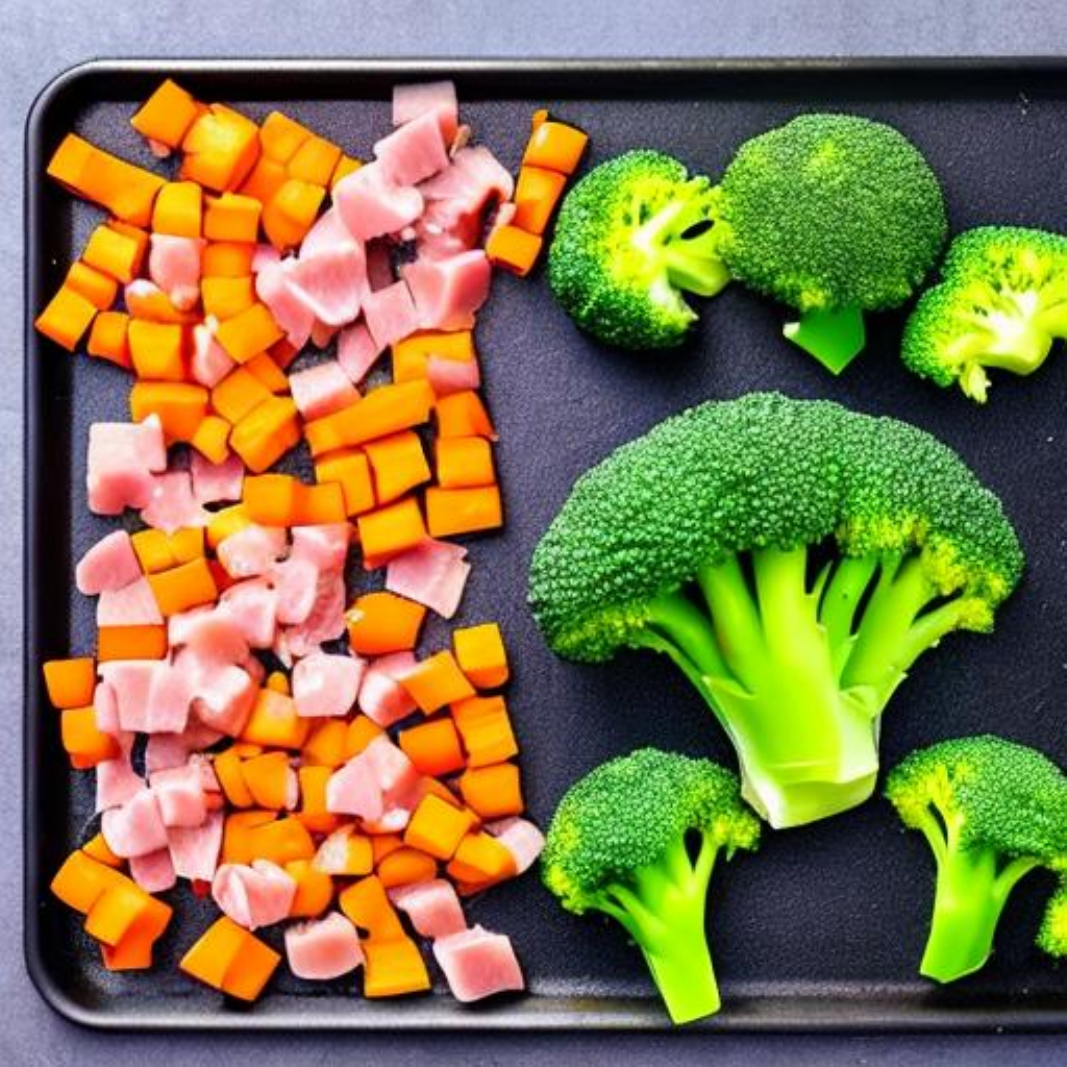}\\
    \midrule
        \makecell[c]{ This is two birds \\pecking at the\\ remnants of a \\burger at an \\outdoor restaurant.}&
        \includegraphics[width=0.22\textwidth]{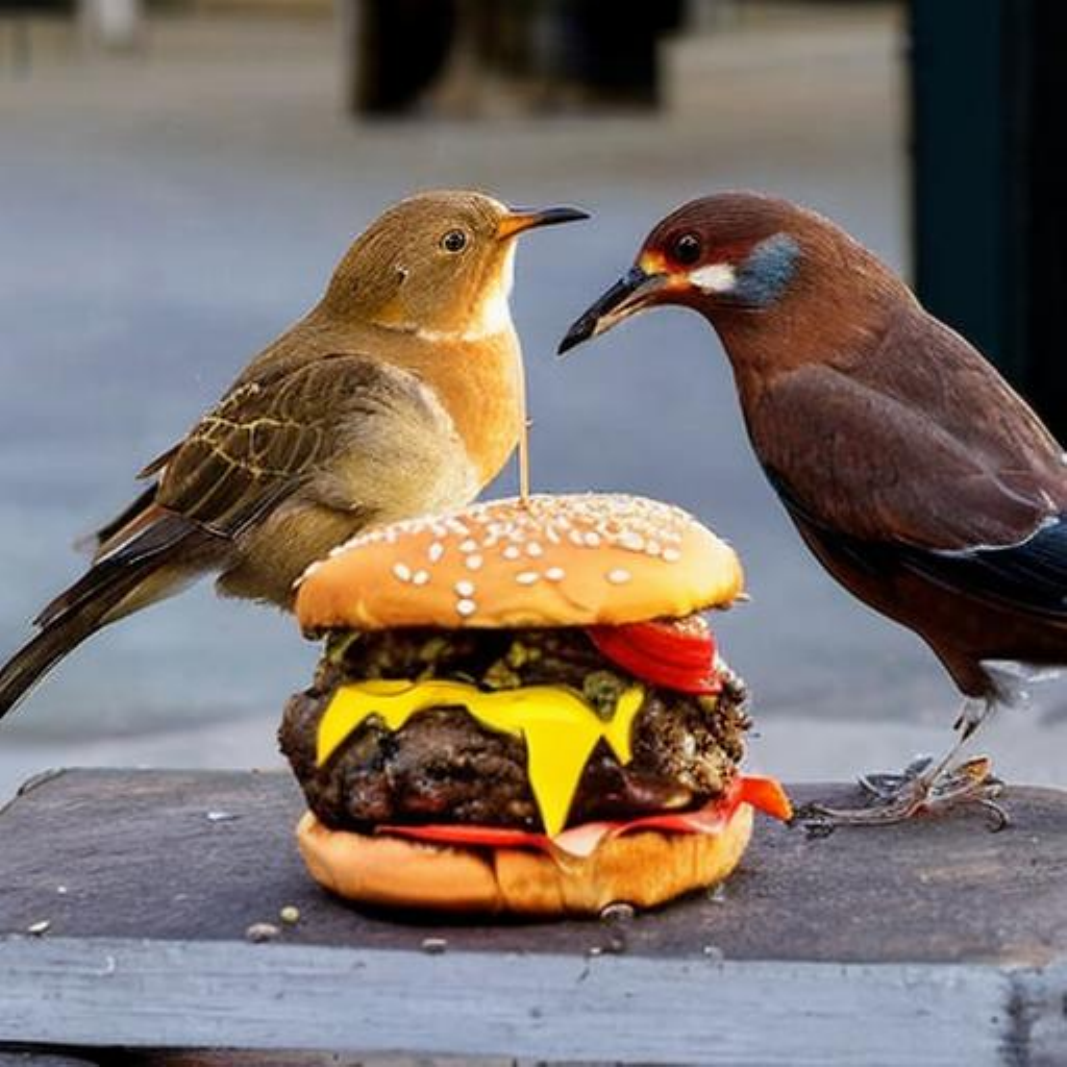}&
        \includegraphics[width=0.22\textwidth]{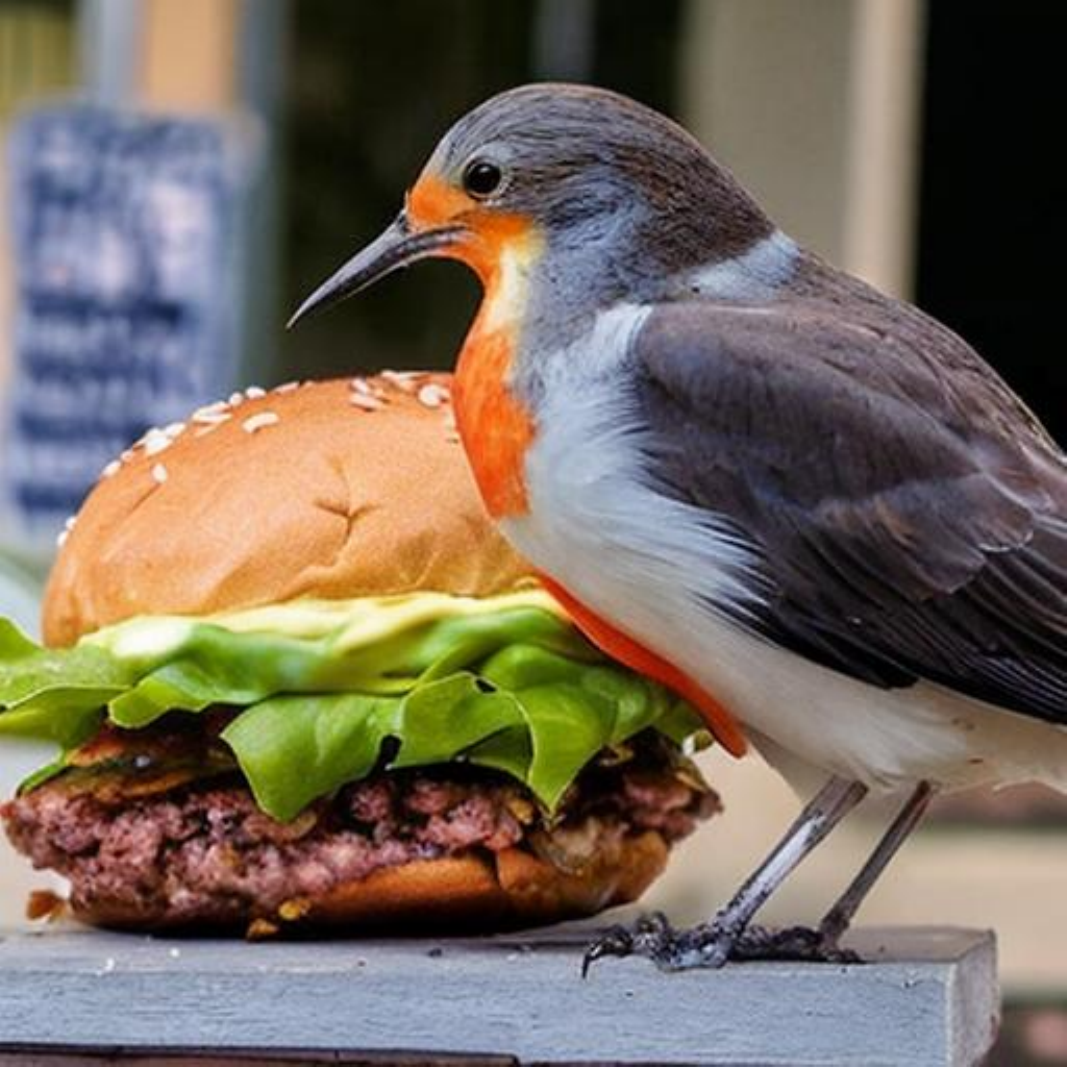}&
        \includegraphics[width=0.22\textwidth]{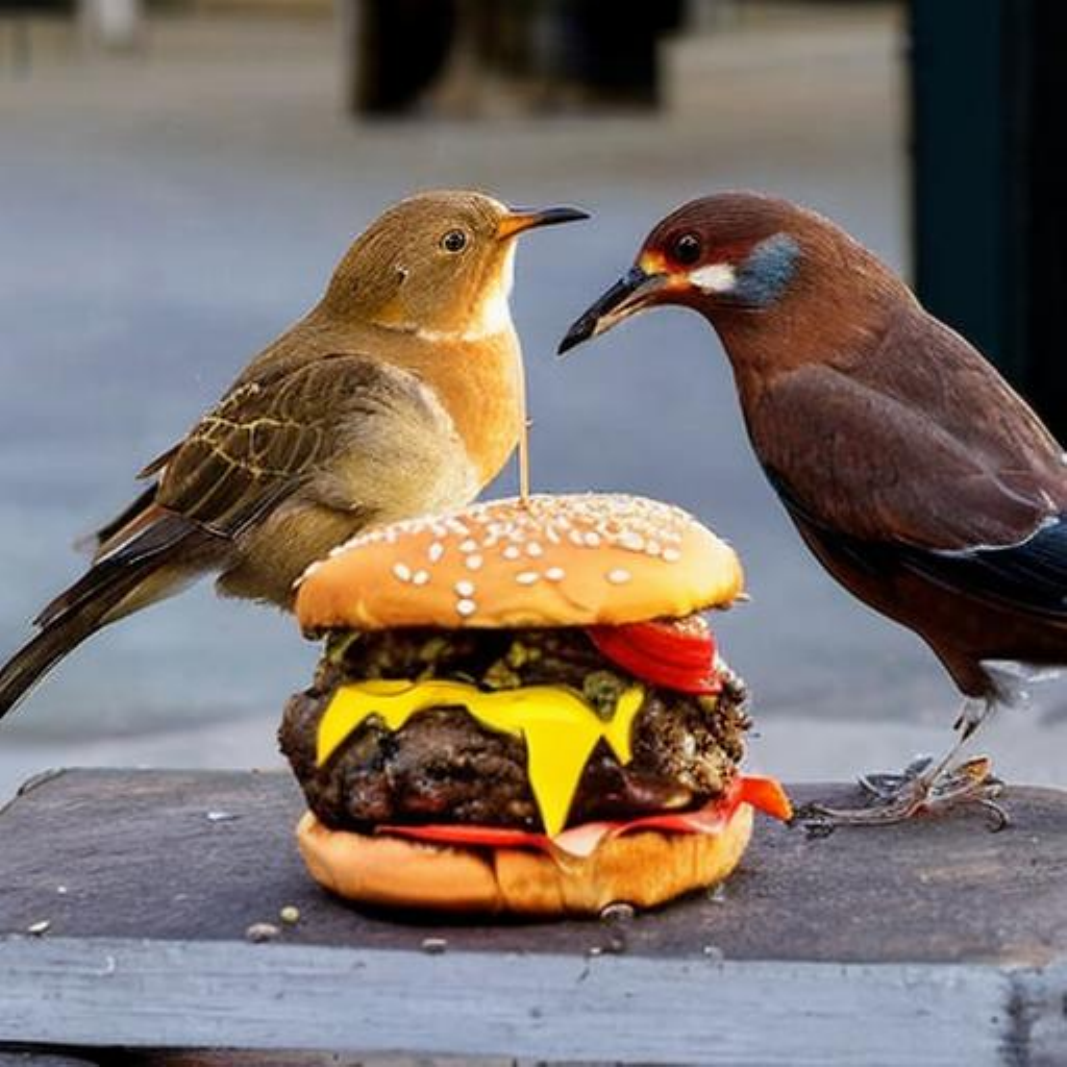}\\
    \midrule
        \makecell[c]{Many surfboards \\are propped \\against a rail \\on the beach.}&
        \includegraphics[width=0.22\textwidth]{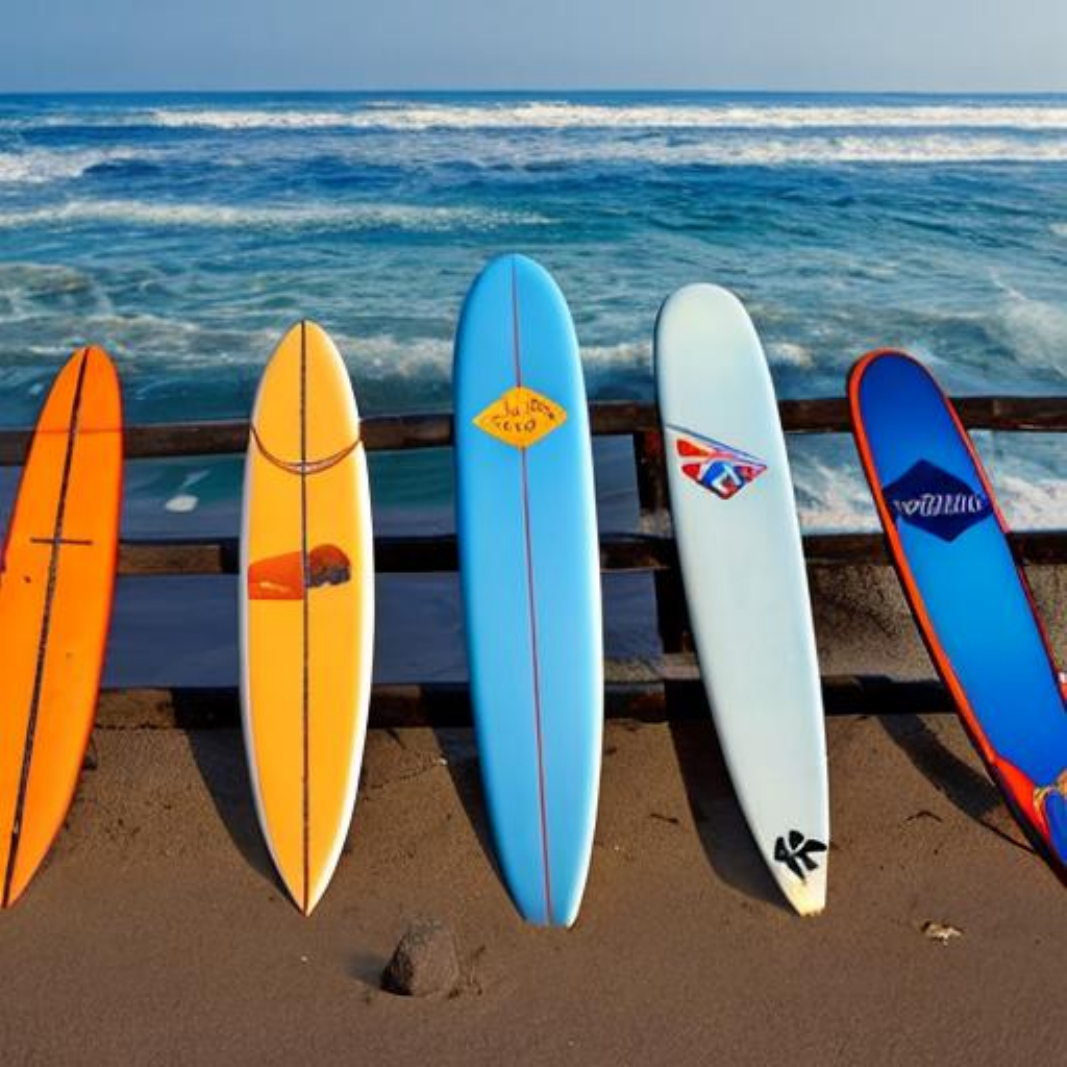}&
        \includegraphics[width=0.22\textwidth]{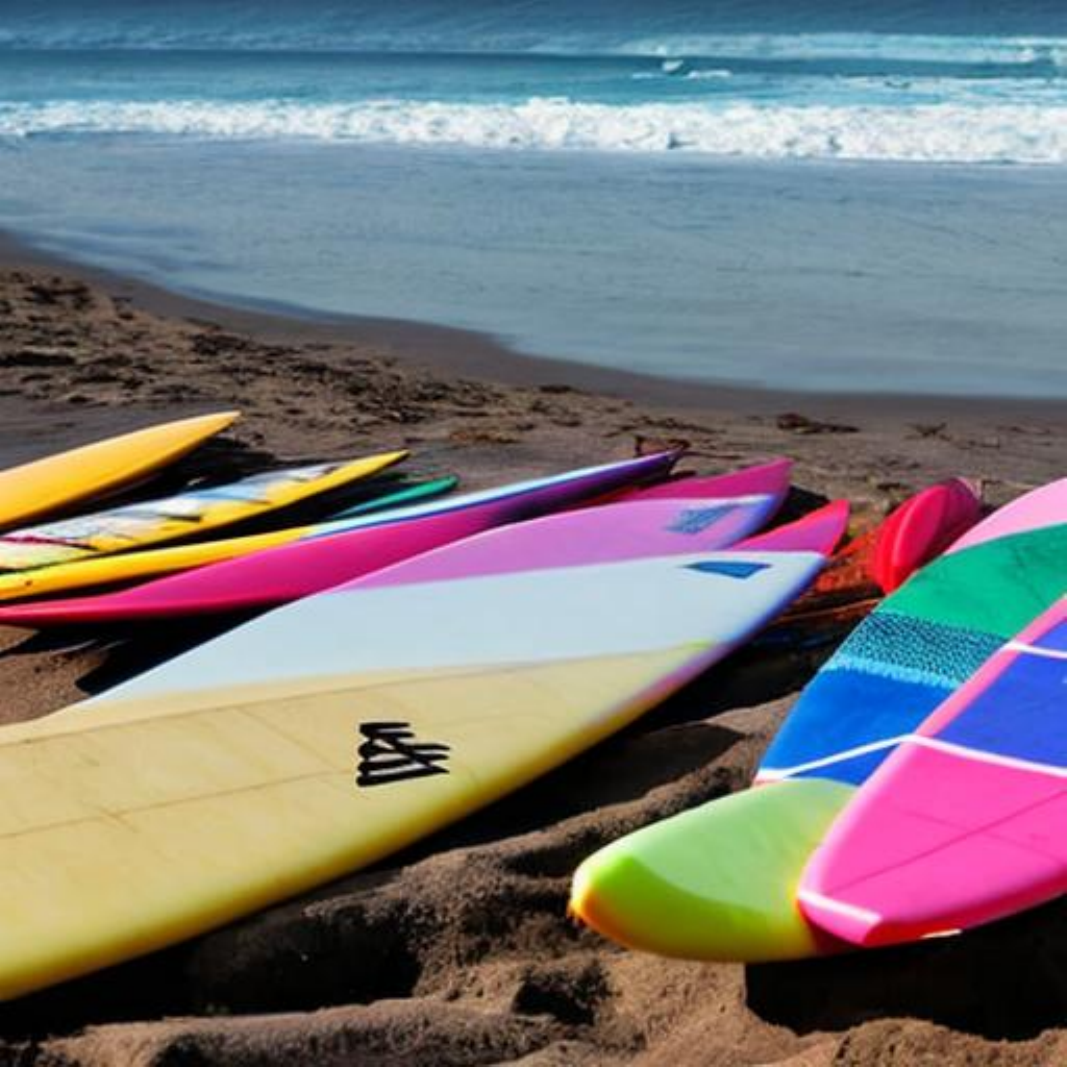}&
        \includegraphics[width=0.22\textwidth]{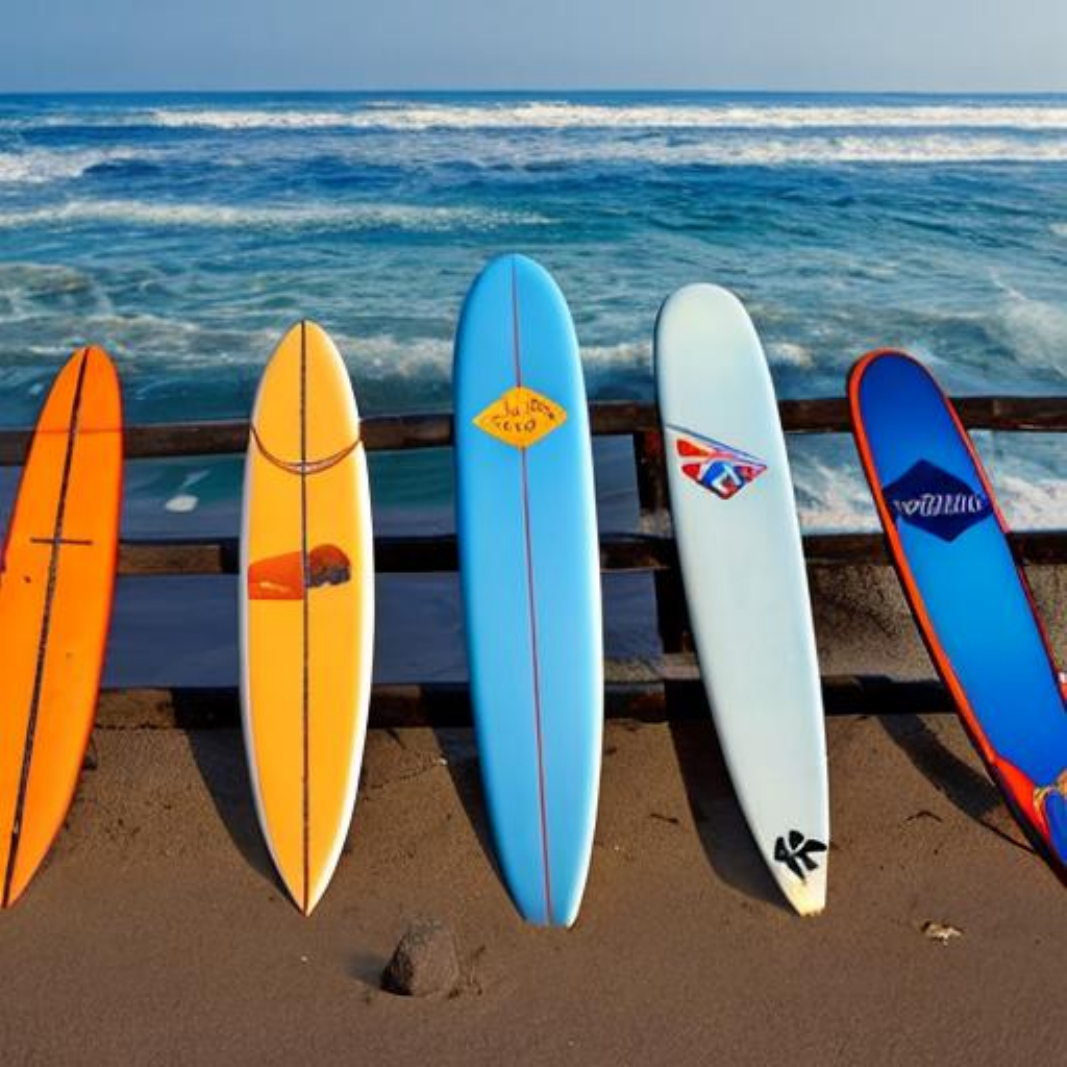}\\
        
    \bottomrule
    \end{tabular}
    \caption{Visual comparison between Tree-Ring and Gaussian Shading on prompts of the validation set of MS-COCO-2017, at resolution 512.  Two methods are applied with the same input latent representations. In contrast to the original model and our Gaussian Shading, Tree-Ring alters the distribution of the latent representations, potentially resulting in the generation of images characterized by semantic inconsistencies or diminished quality. This figure illustrates an instance of such a case, where the Gaussian Shading preserves the distribution, thereby avoiding this issue.}
    \label{fig:quality_tree}
\end{figure*}

See \cref{fig:quality_1} and \cref{fig:quality_tree}, we present the visual results of different watermarking methods on prompts from the MS-COCO-2017~\cite{lin2014microsoft} validation set. From the residual images in  \cref{fig:quality_1}, it can be observed that DwtDct~\cite{cox2007digital}, DwtDctSvd~\cite{cox2007digital}, RivaGAN~\cite{zhang2019robust}, and Stable Signature~\cite{fernandez2023stable} introduce noticeable watermark artifacts, leading to a degradation in model performance. As shown in \cref{fig:quality_tree}, although Tree-Ring~\cite{wen2023tree} watermark is imperceptible, its embedding may directly impair the image quality. Additionally, it may also introduce changes in the object count and spatial relationships, causing inconsistency with the prompt. In the case of Gaussian Shading, as long as the latent representations where the watermark is mapped remain consistent with that of the original image, no changes occur in the generated image.

\begin{figure*}[]
    \centering
    \begin{tabular}{ccccc}
    \toprule
    \multicolumn{5}{c}{\makecell[c]{Red dead redemption 2, cinematic view, epic sky, detailed, concept art, low angle, high detail, warm lighting, volumetric,\\ godrays, vivid, beautiful, trending on artstation, by jordan grimmer, huge scene, grass, art greg rutkowski.}}\\
    \midrule
        \includegraphics[width=3cm]{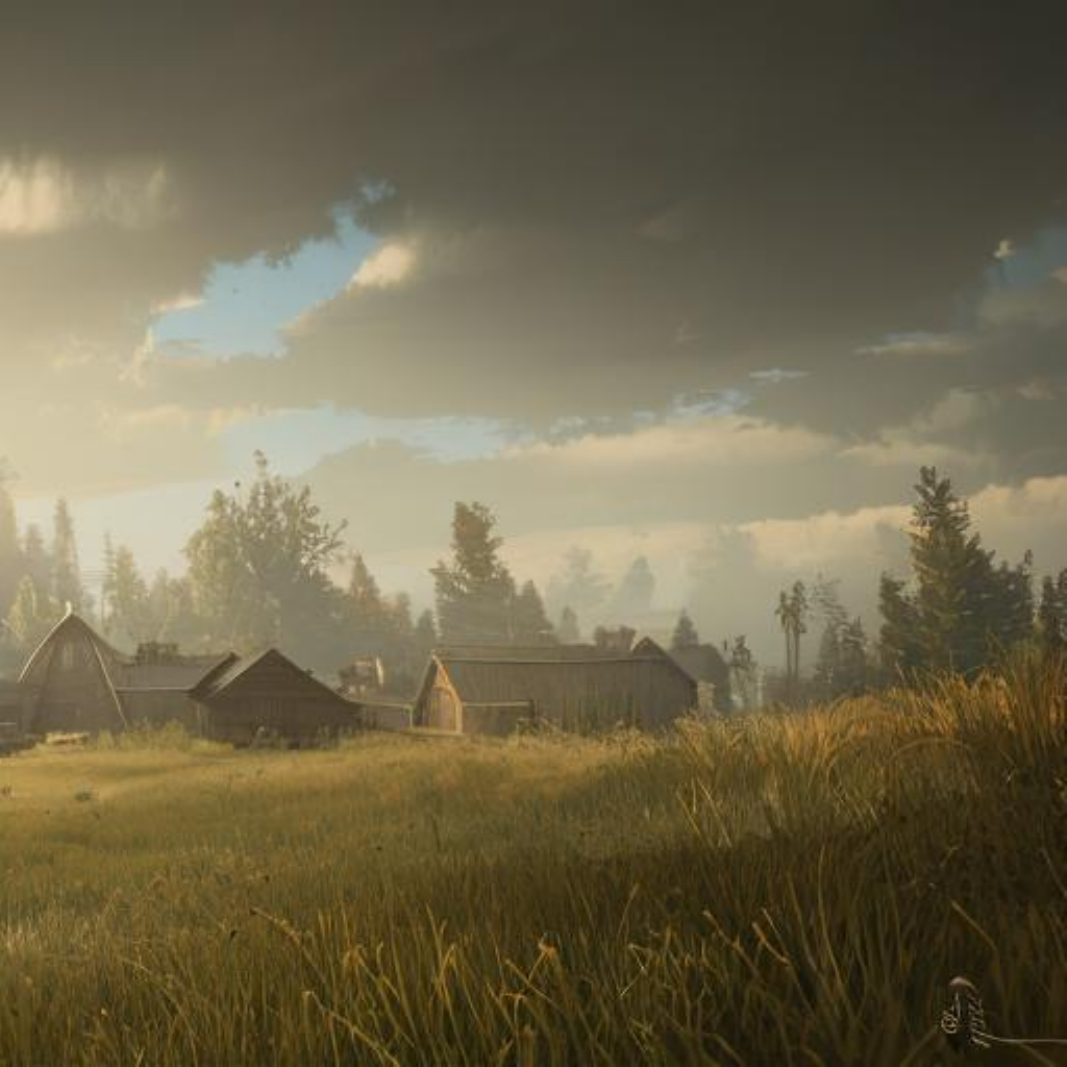} &
        \includegraphics[width=3cm]{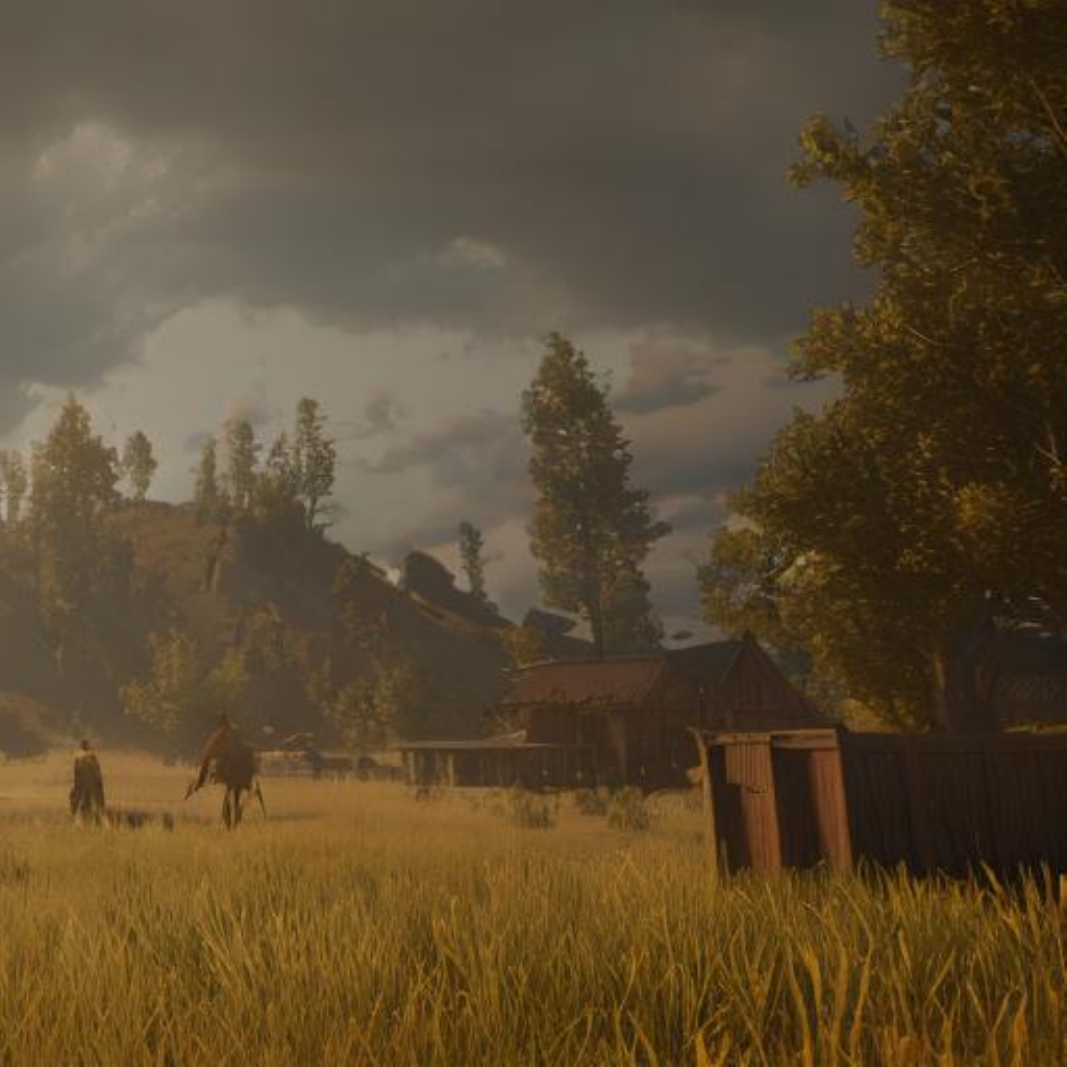} &
        \includegraphics[width=3cm]{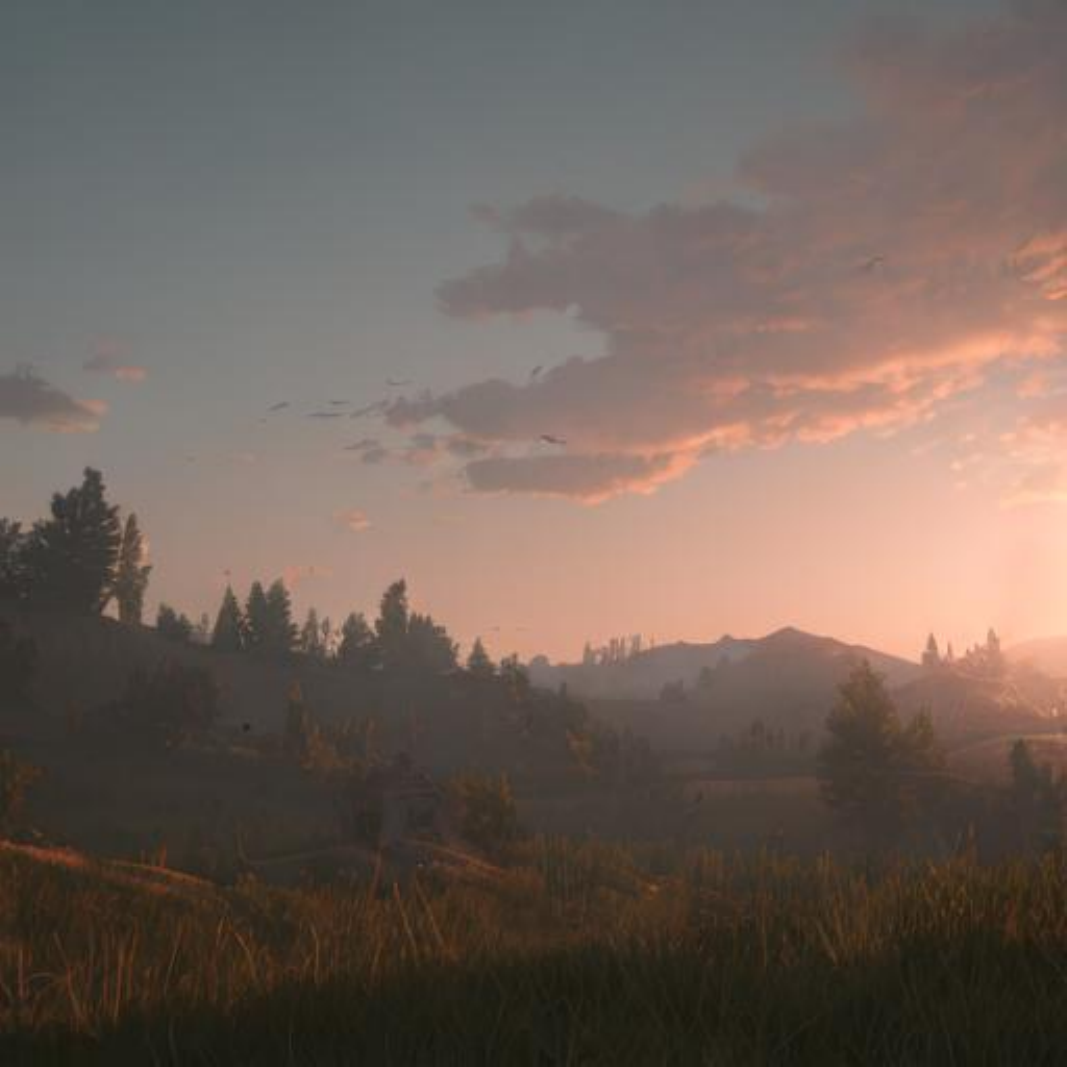} &
        \includegraphics[width=3cm]{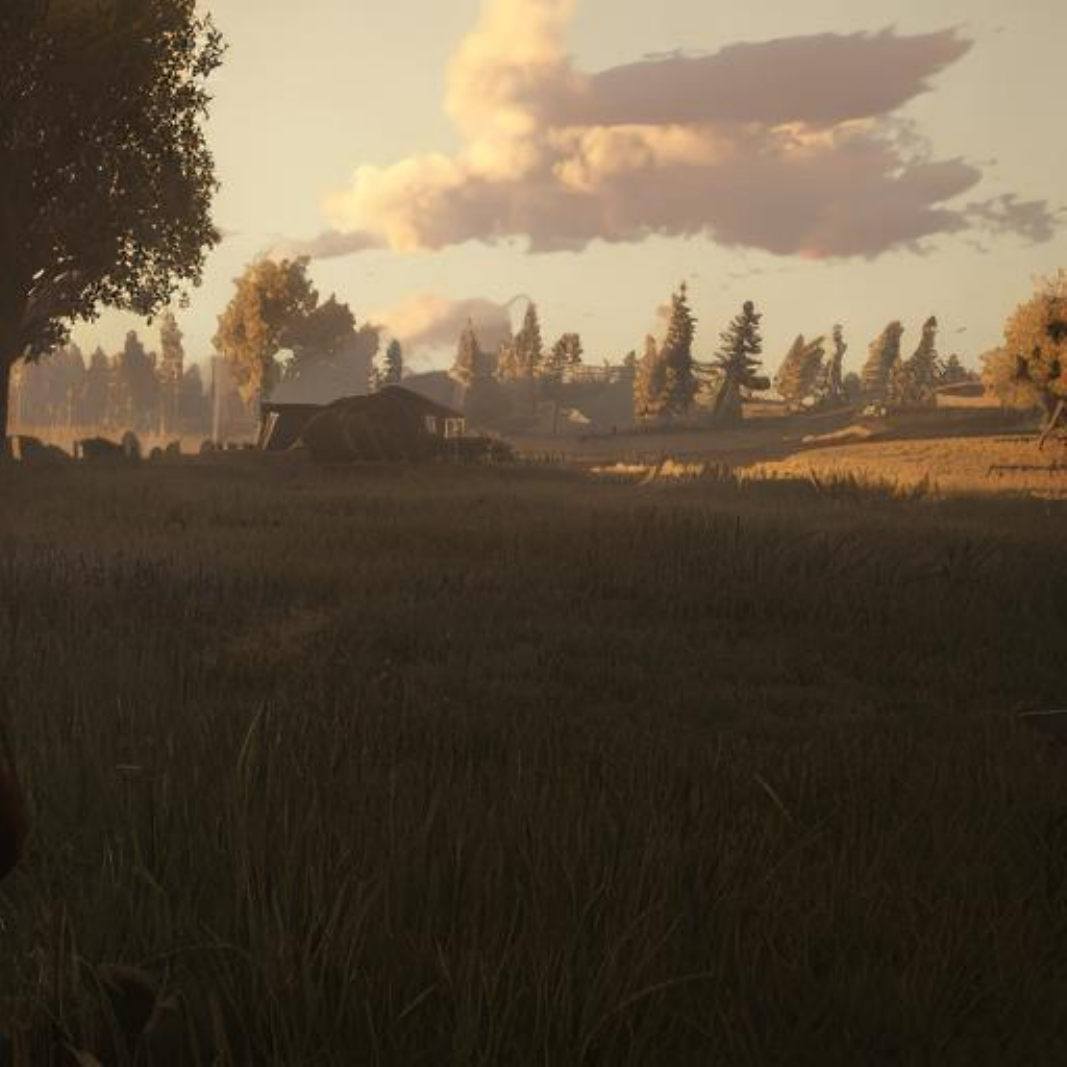} &
        \includegraphics[width=3cm]{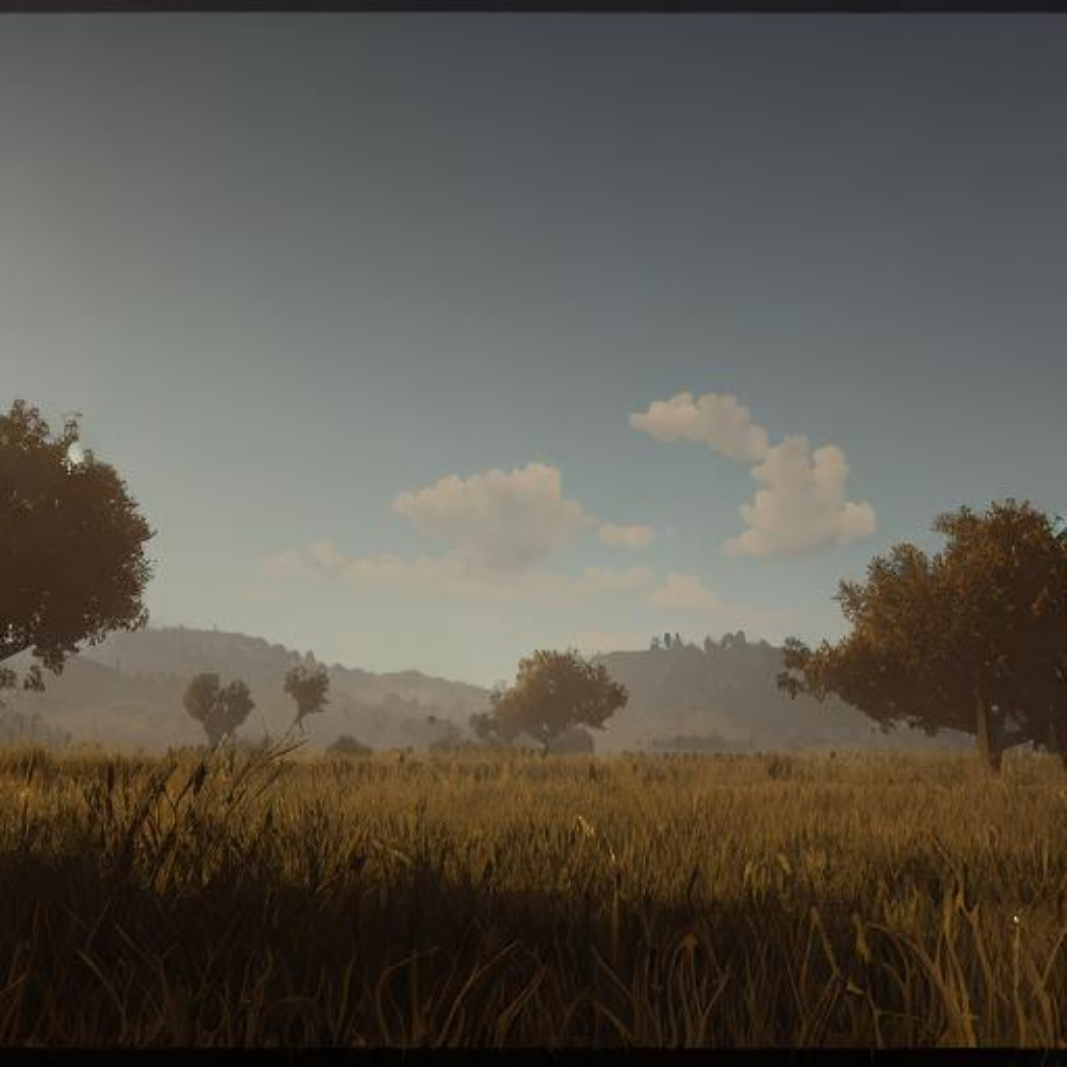} \\
    \midrule
     \multicolumn{5}{c}{\makecell[c]{Official Portrait of a smiling WWI admiral, male, cheerful, happy, detailed face, 20th century, \\ highly detailed, cinematic lighting, digital art painting by greg rutkowski.}}\\
    \midrule
        \includegraphics[width=3cm]{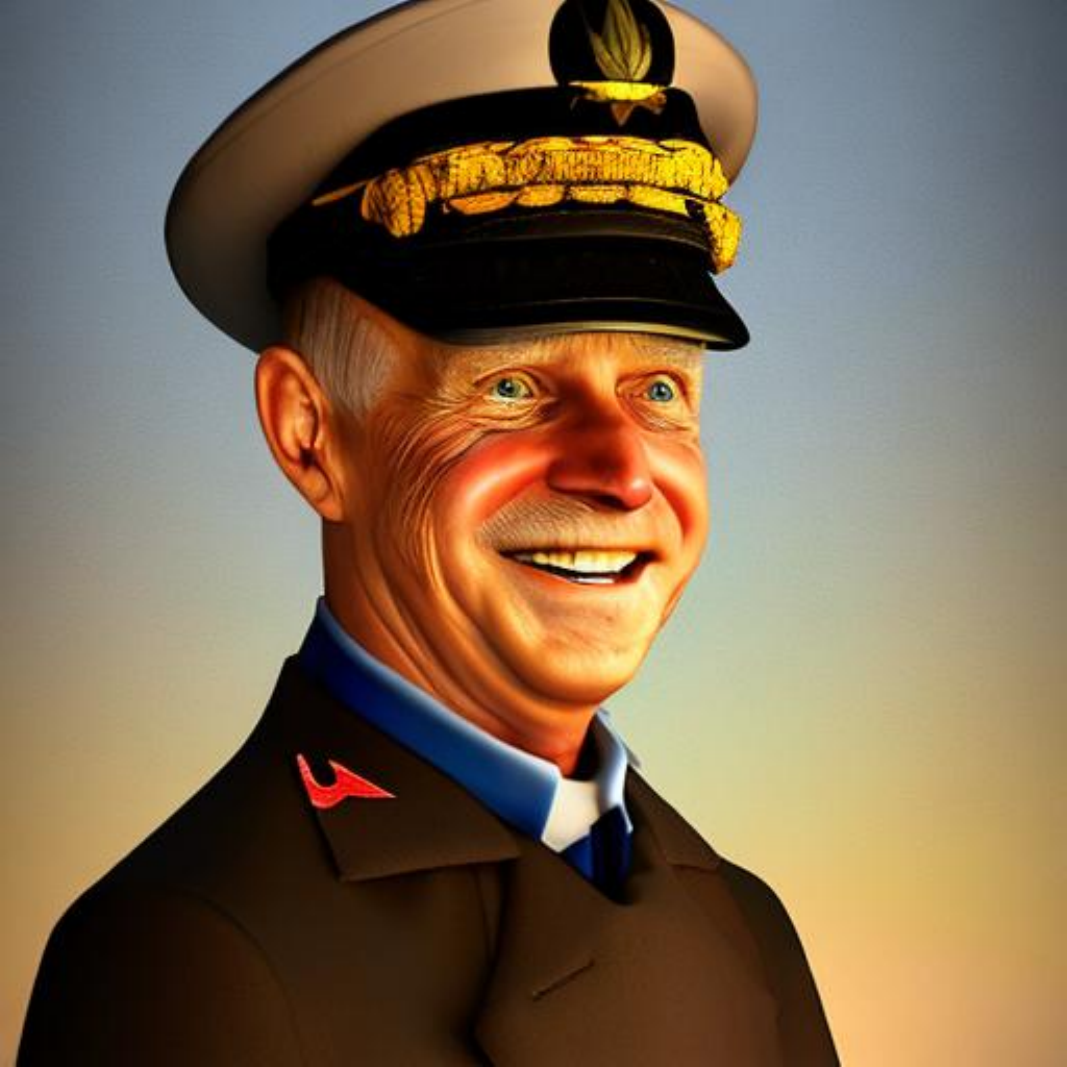} &
        \includegraphics[width=3cm]{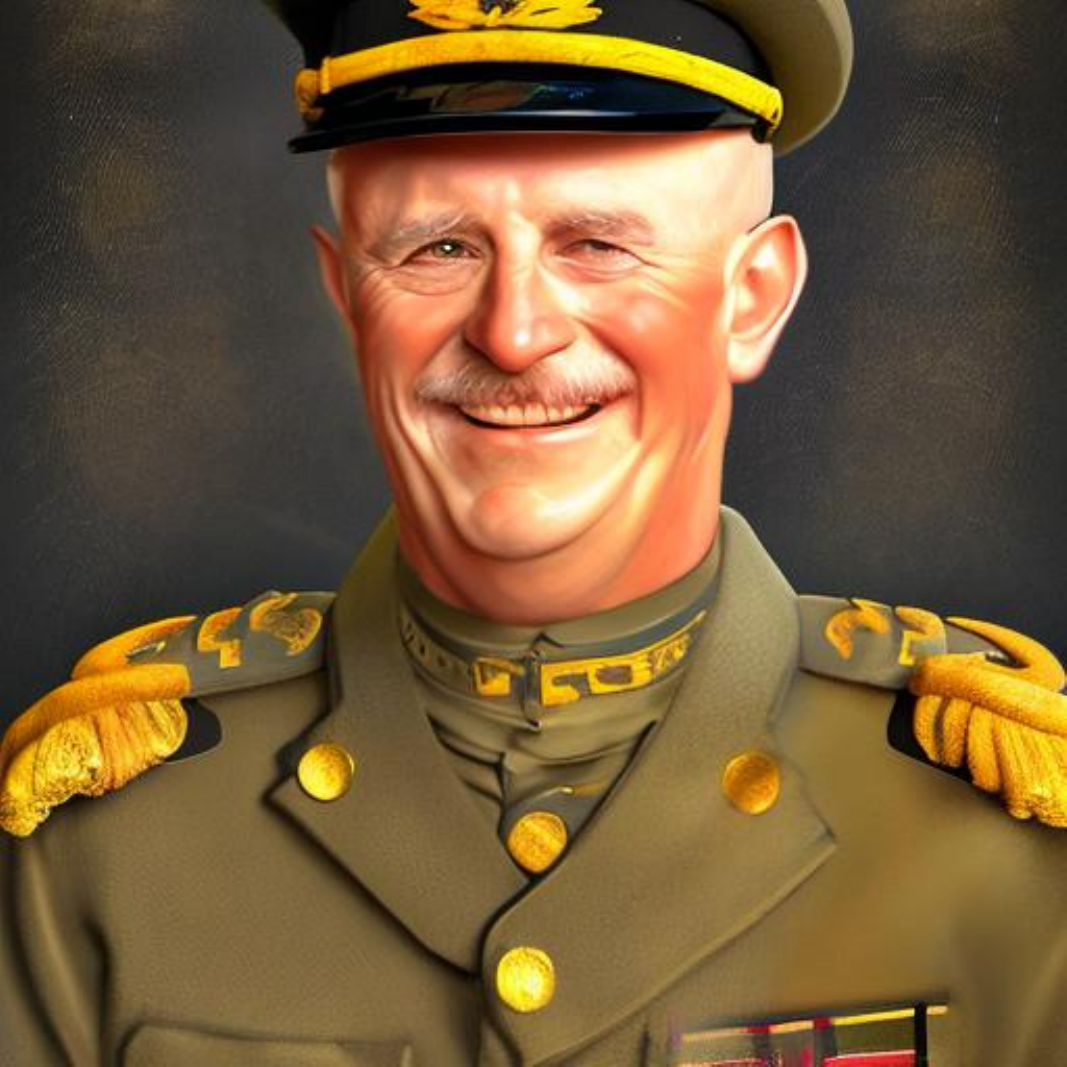} &
        \includegraphics[width=3cm]{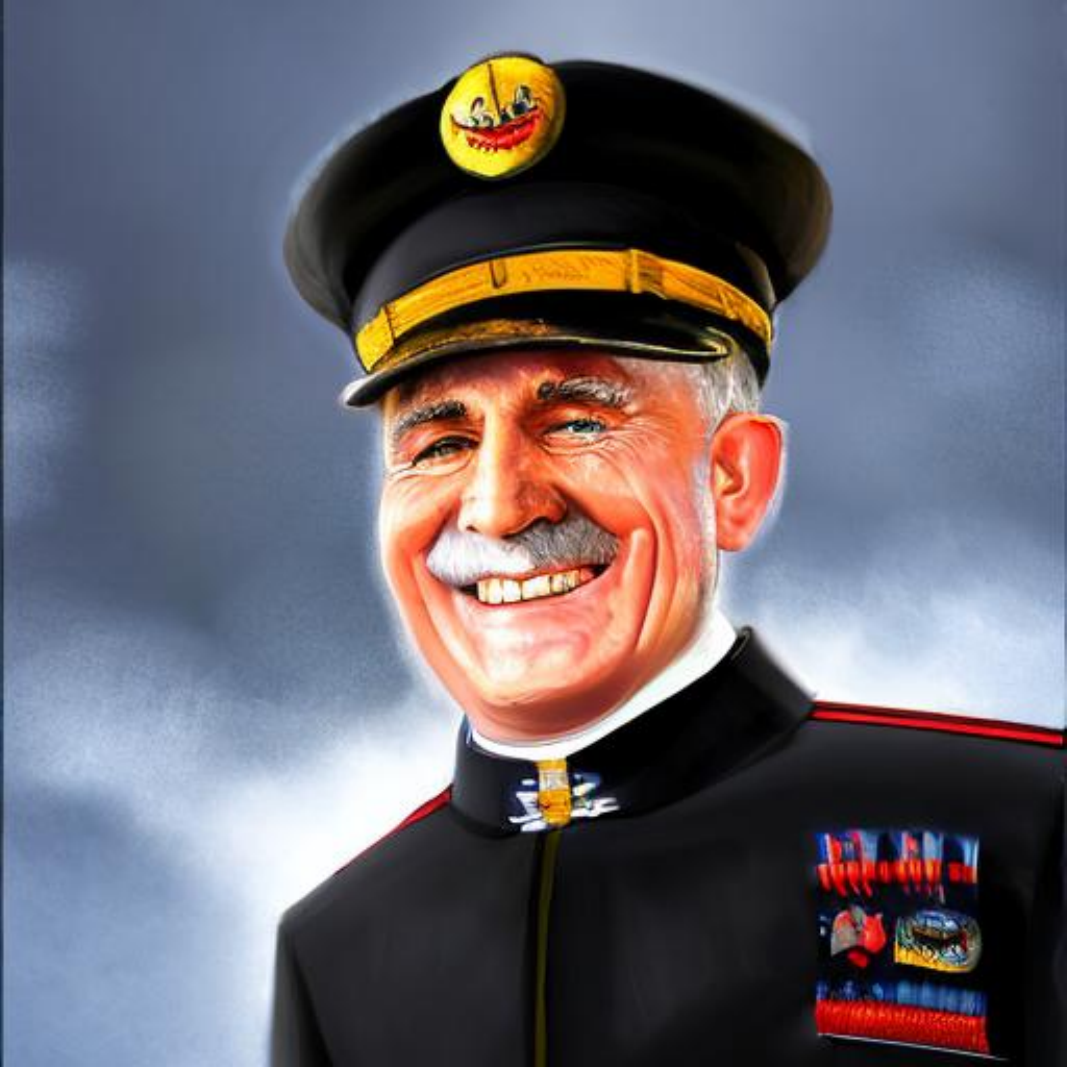} &
        \includegraphics[width=3cm]{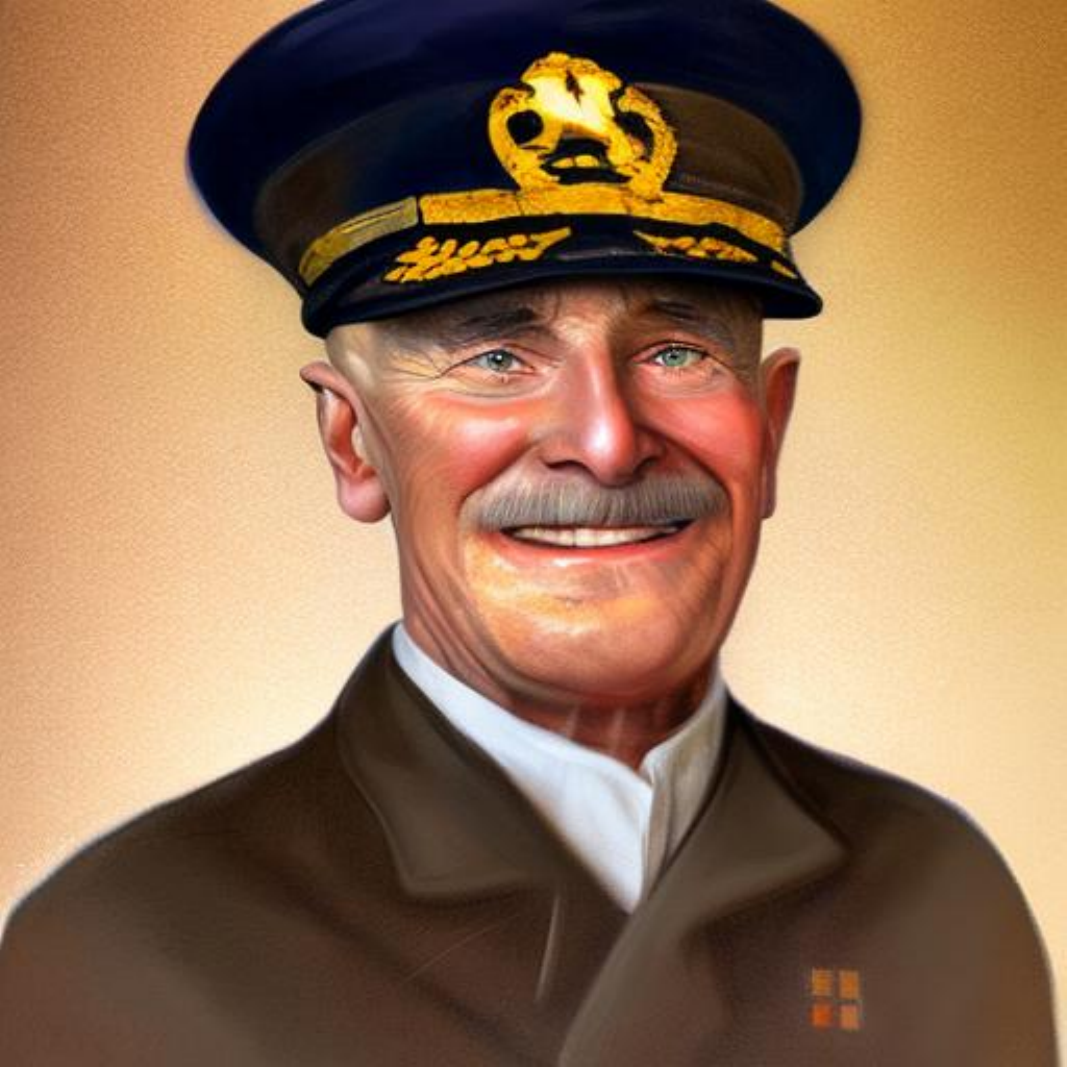} &
        \includegraphics[width=3cm]{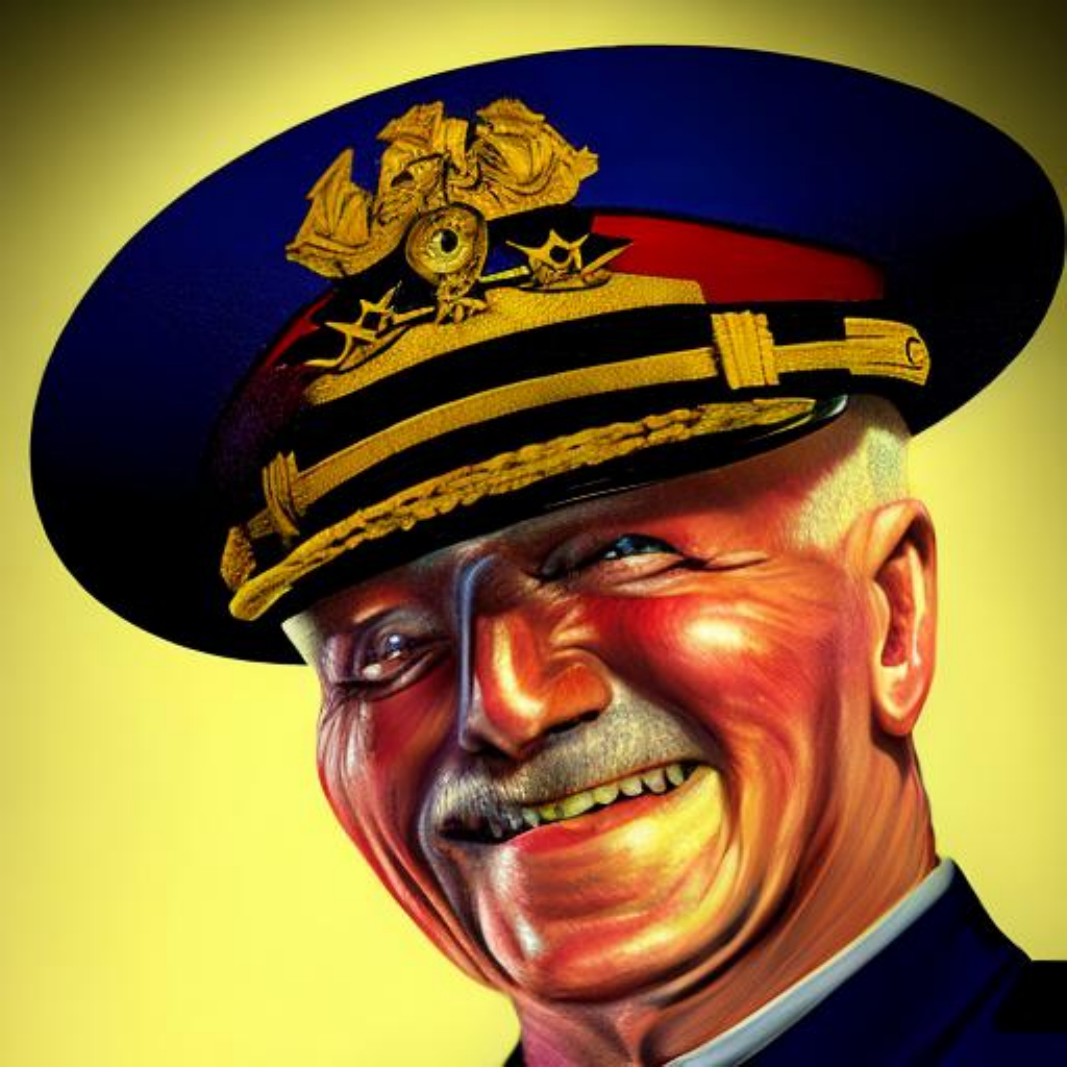} \\
    \midrule
     \multicolumn{5}{c}{\makecell[c]{Post apocalyptic city overgrown abandoned city, highly detailed, art by Range Murata, highly detailed, \\ 3d, octane render, bright colors, digital painting, trending on artstation, sharp focus.}}\\
    \midrule
        \includegraphics[width=3cm]{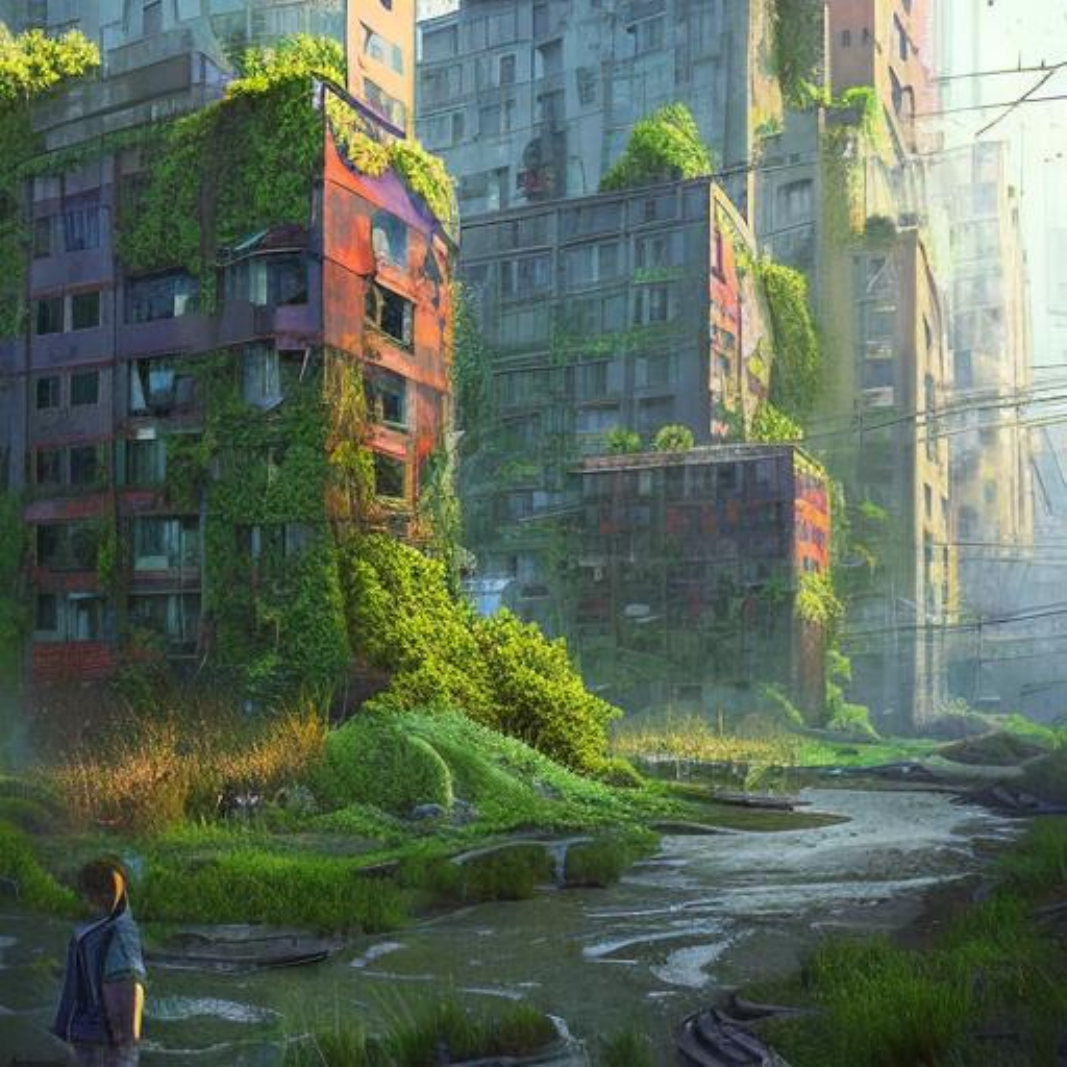} &
        \includegraphics[width=3cm]{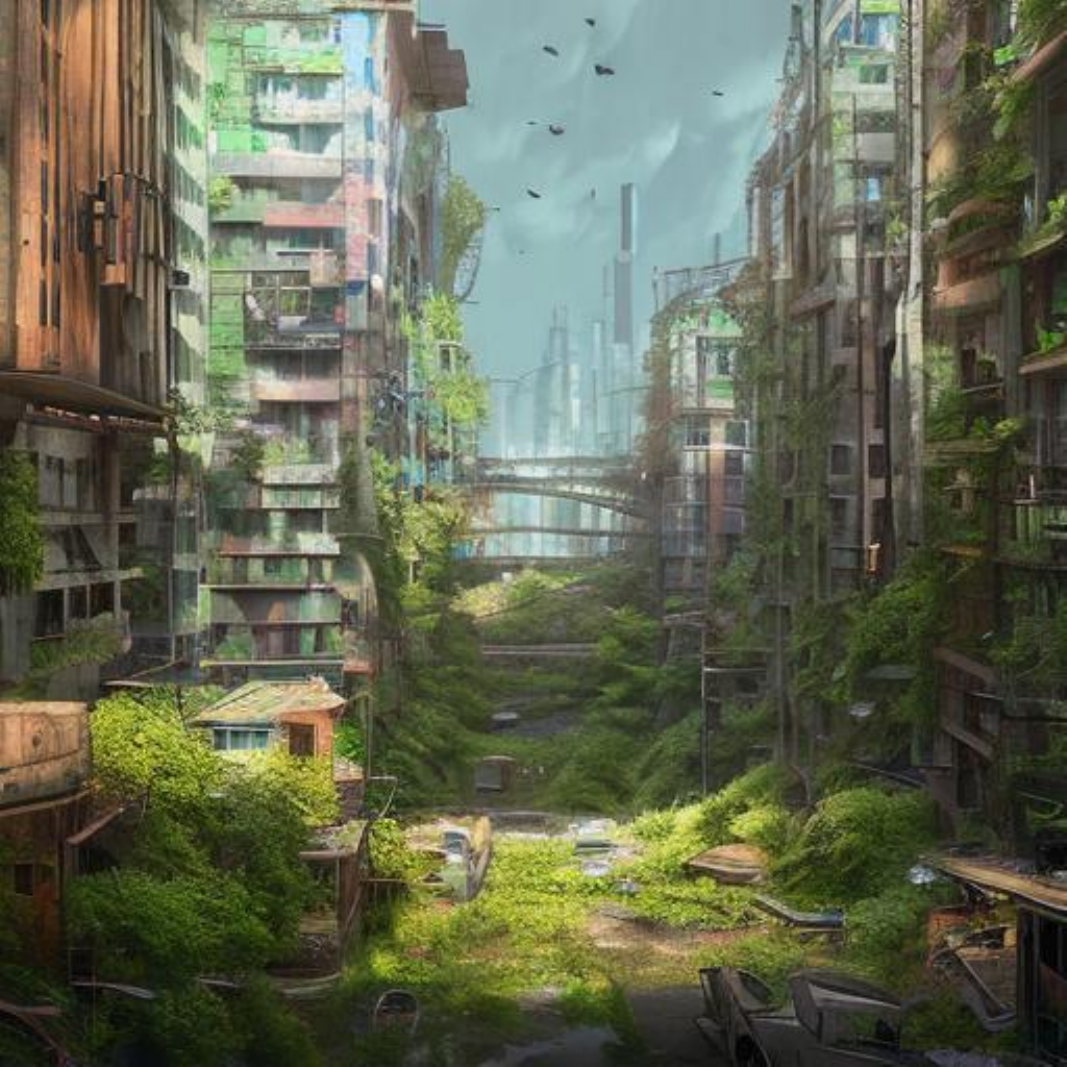} &
        \includegraphics[width=3cm]{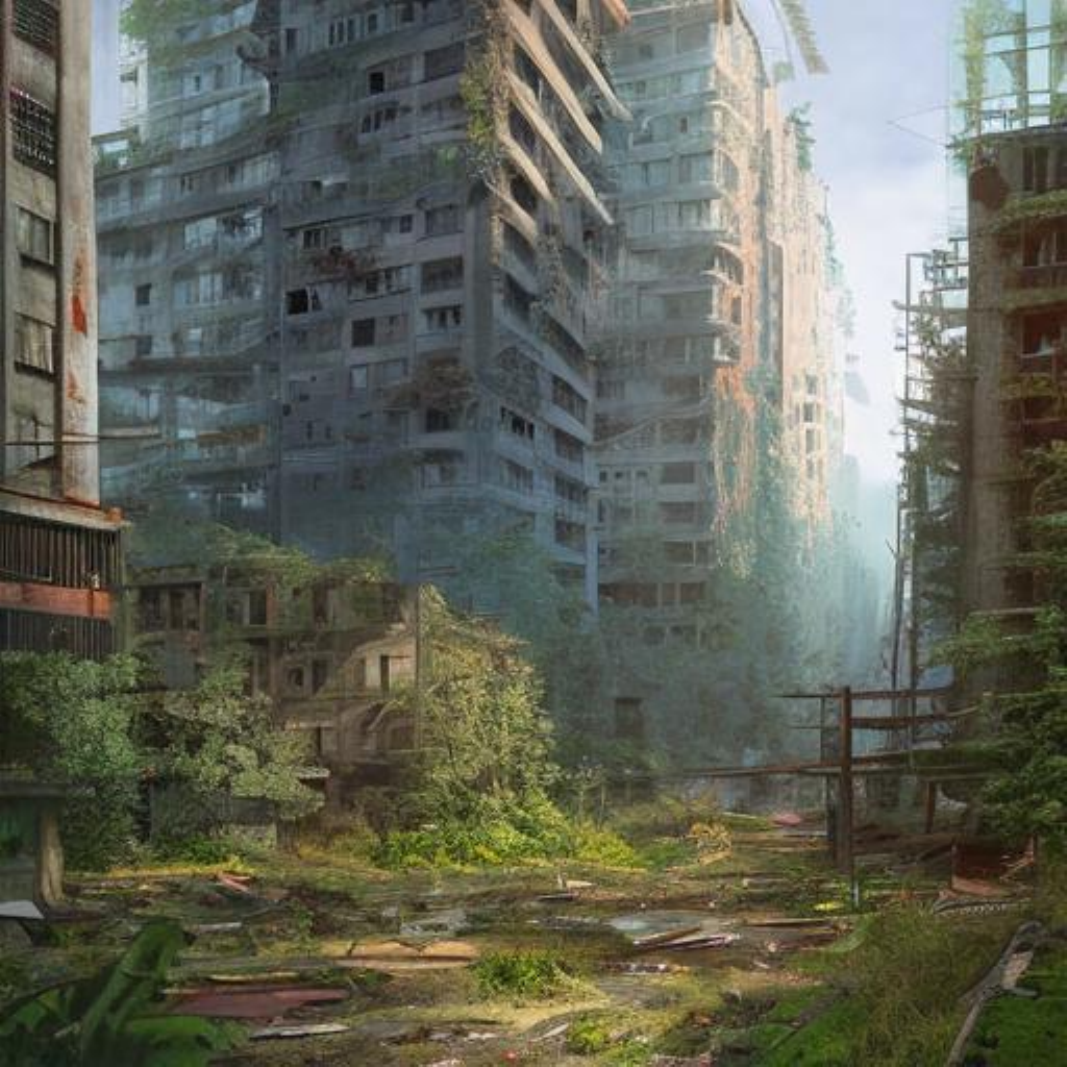} &
        \includegraphics[width=3cm]{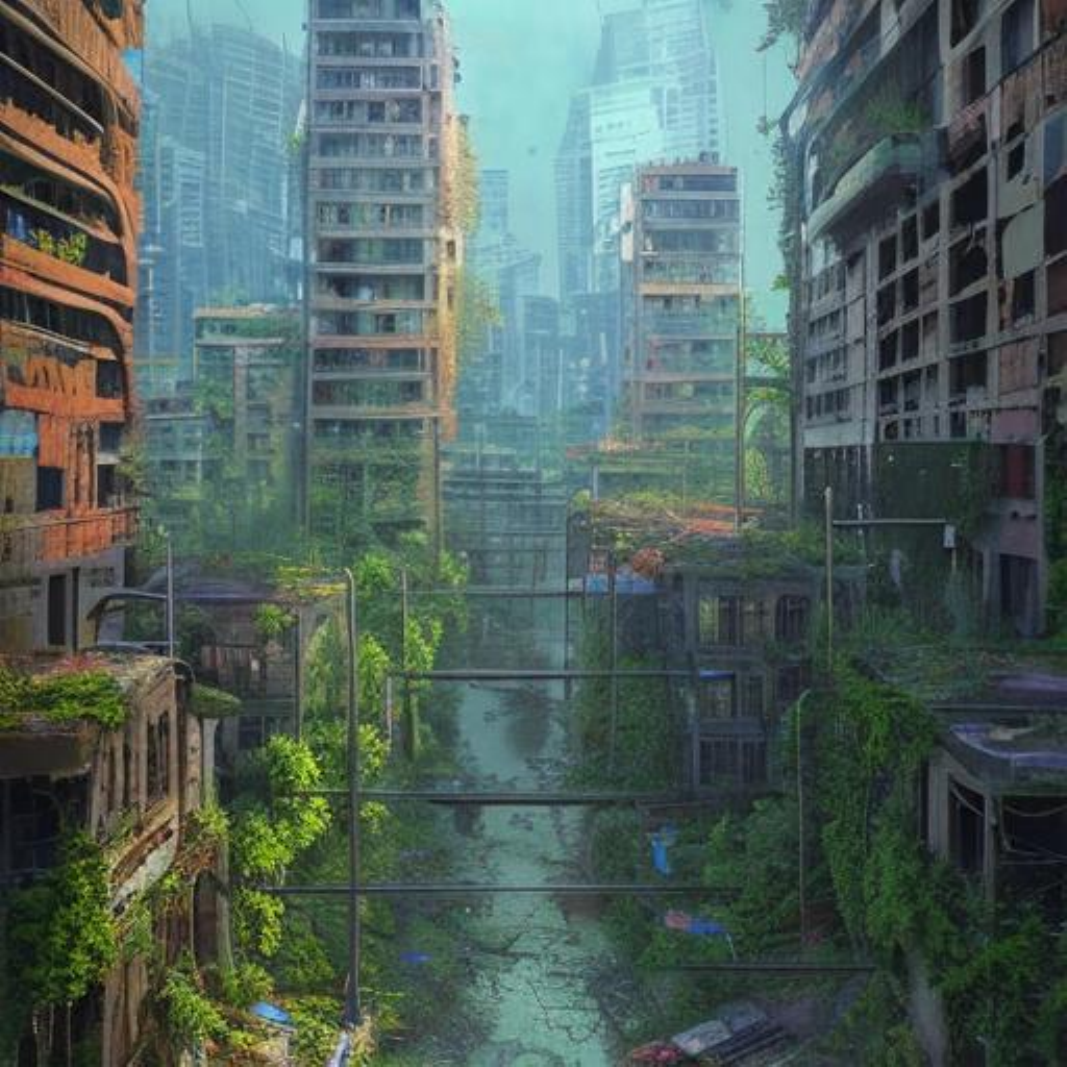} &
        \includegraphics[width=3cm]{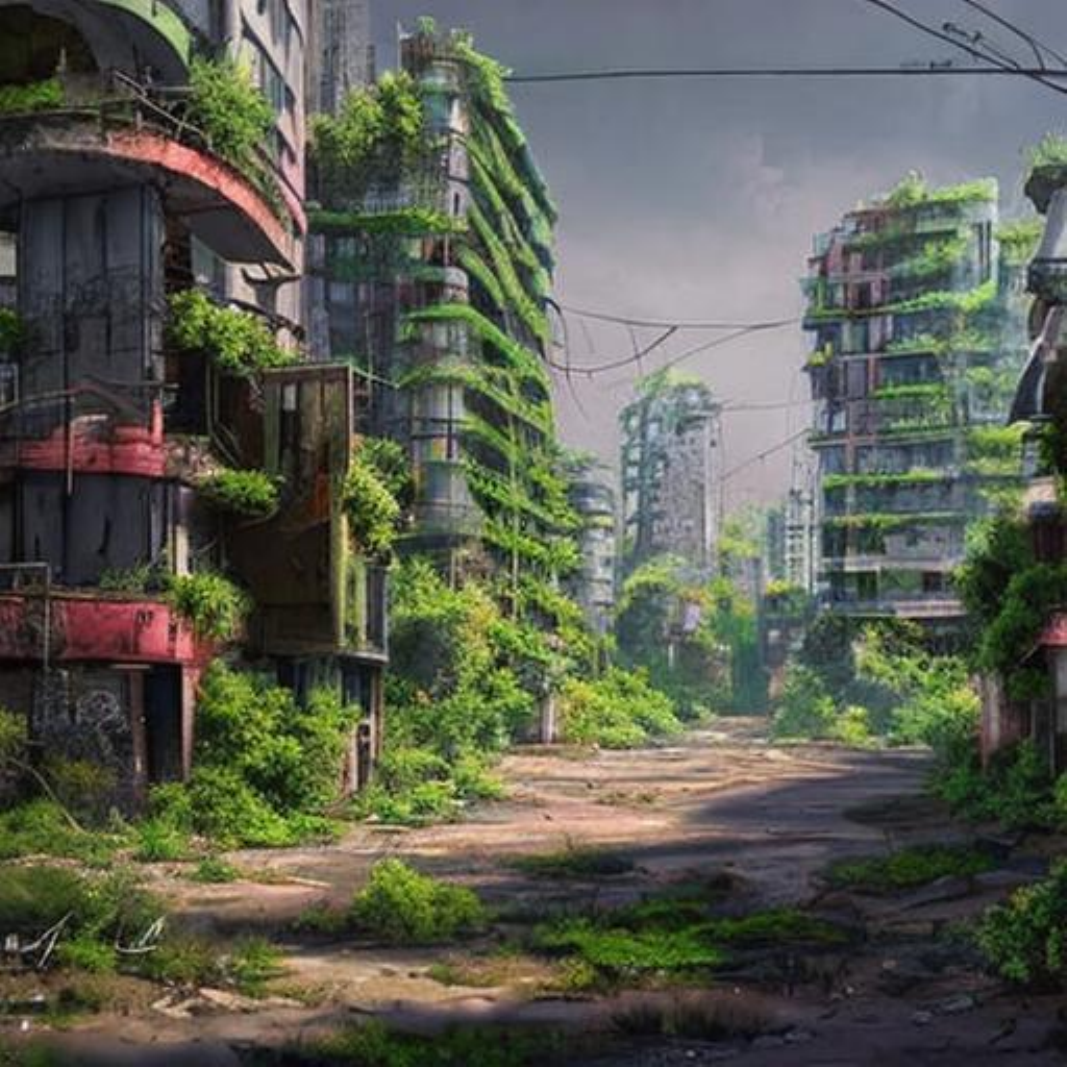} \\
    \midrule
     \multicolumn{5}{c}{\makecell[c]{A female master, character art portrait, anime key visual, official media, illustrated by wlop, \\ extremely detailed, 8 k, trending on artstation, cinematic lighting, beautiful.}}\\
    \midrule
        \includegraphics[width=3cm]{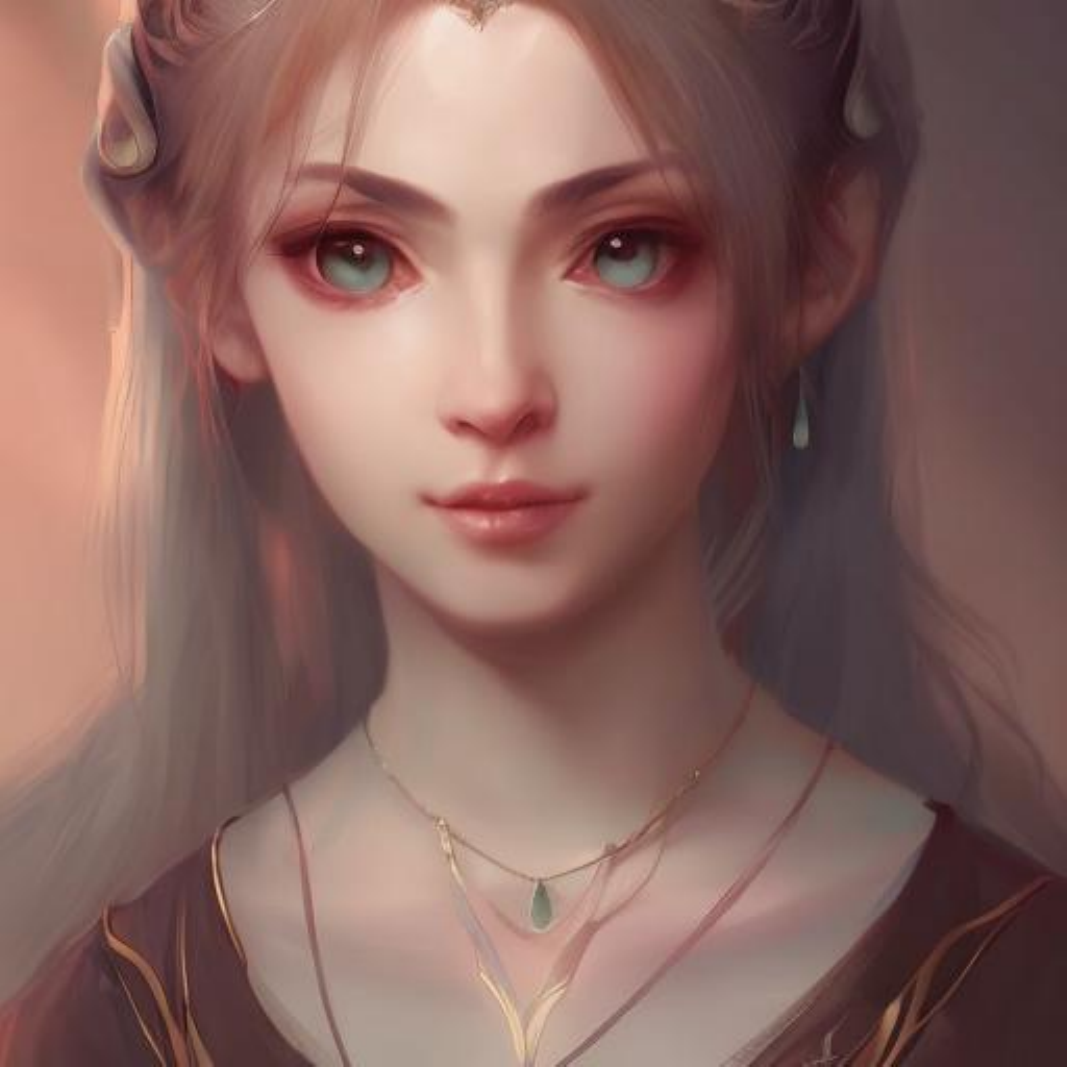} &
        \includegraphics[width=3cm]{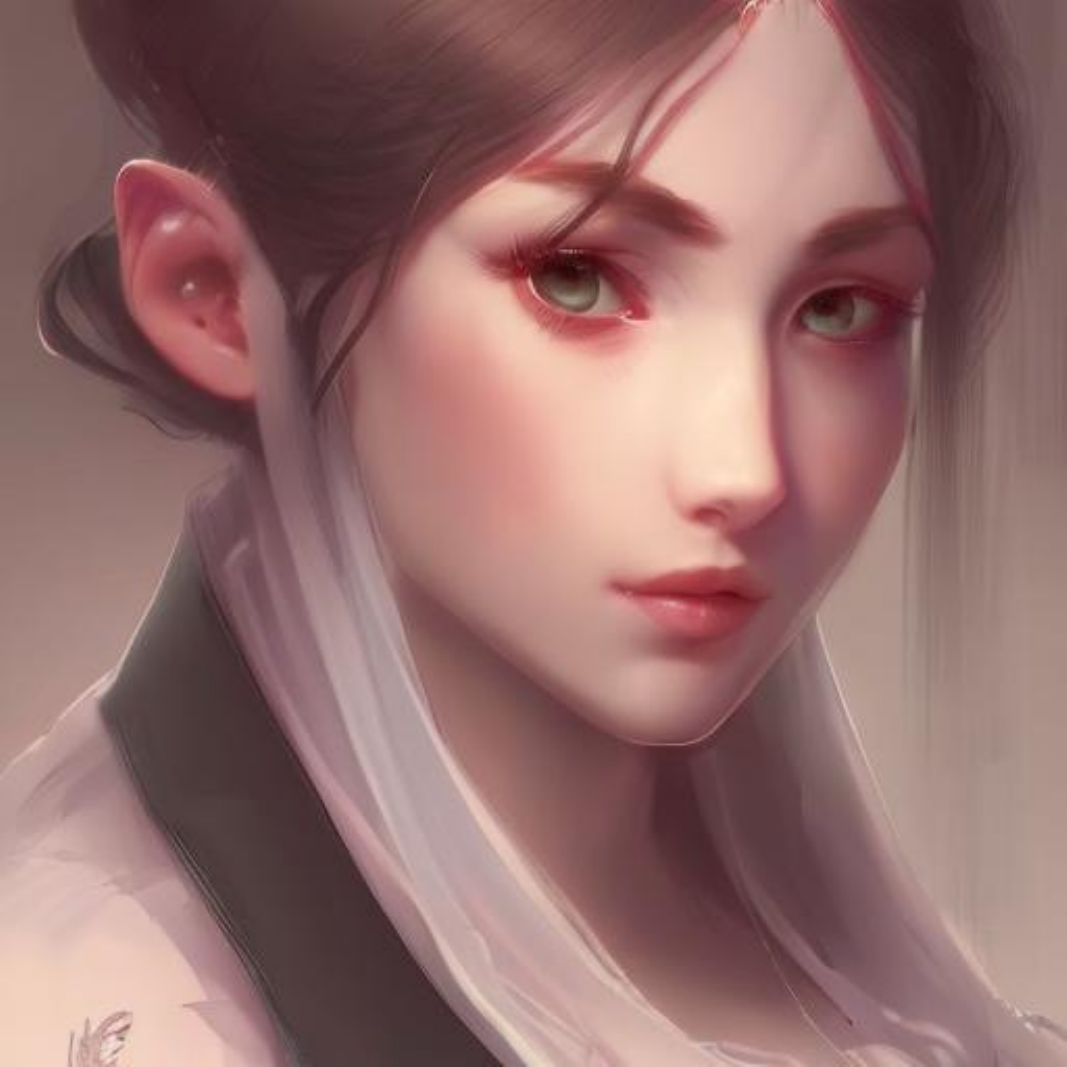} &
        \includegraphics[width=3cm]{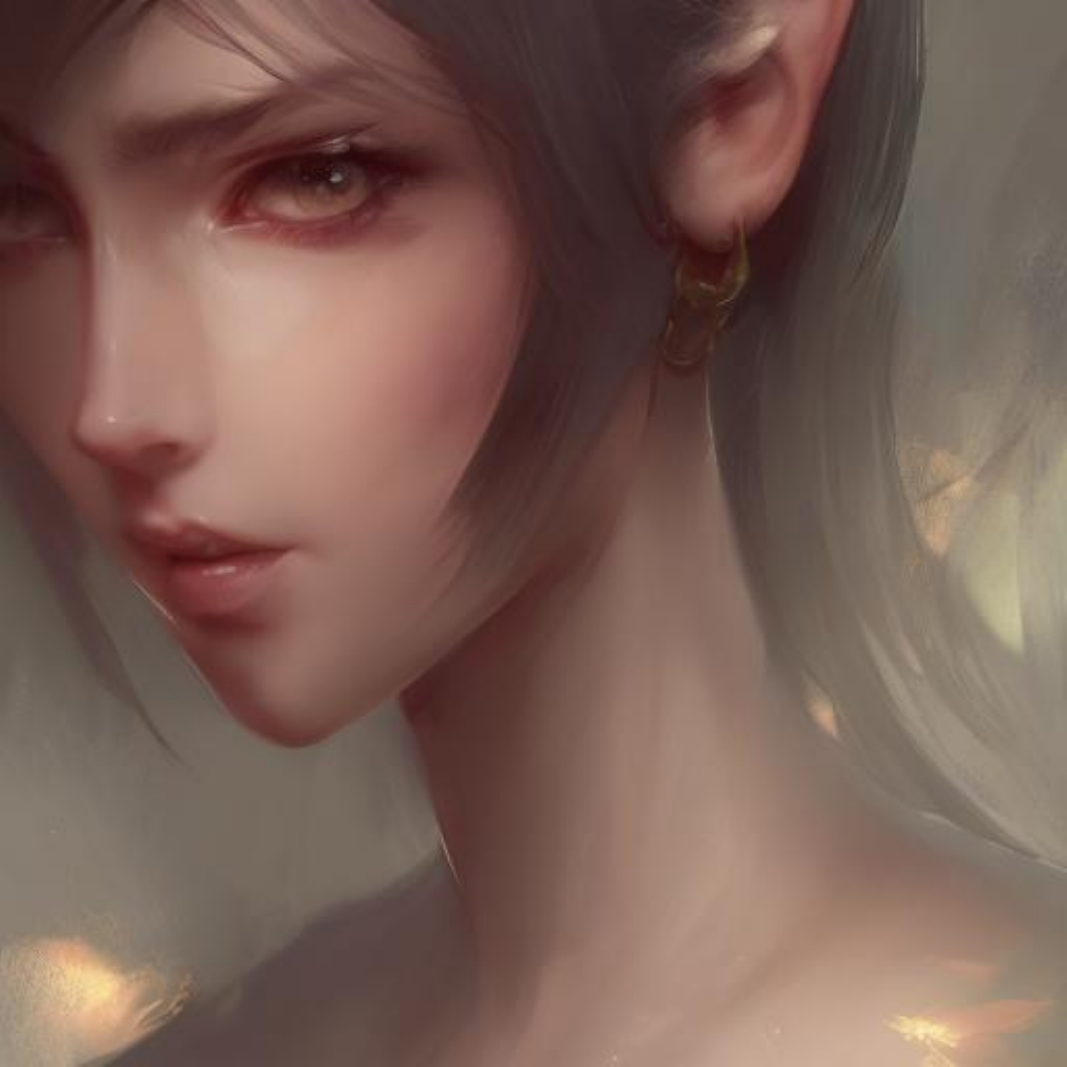} &
        \includegraphics[width=3cm]{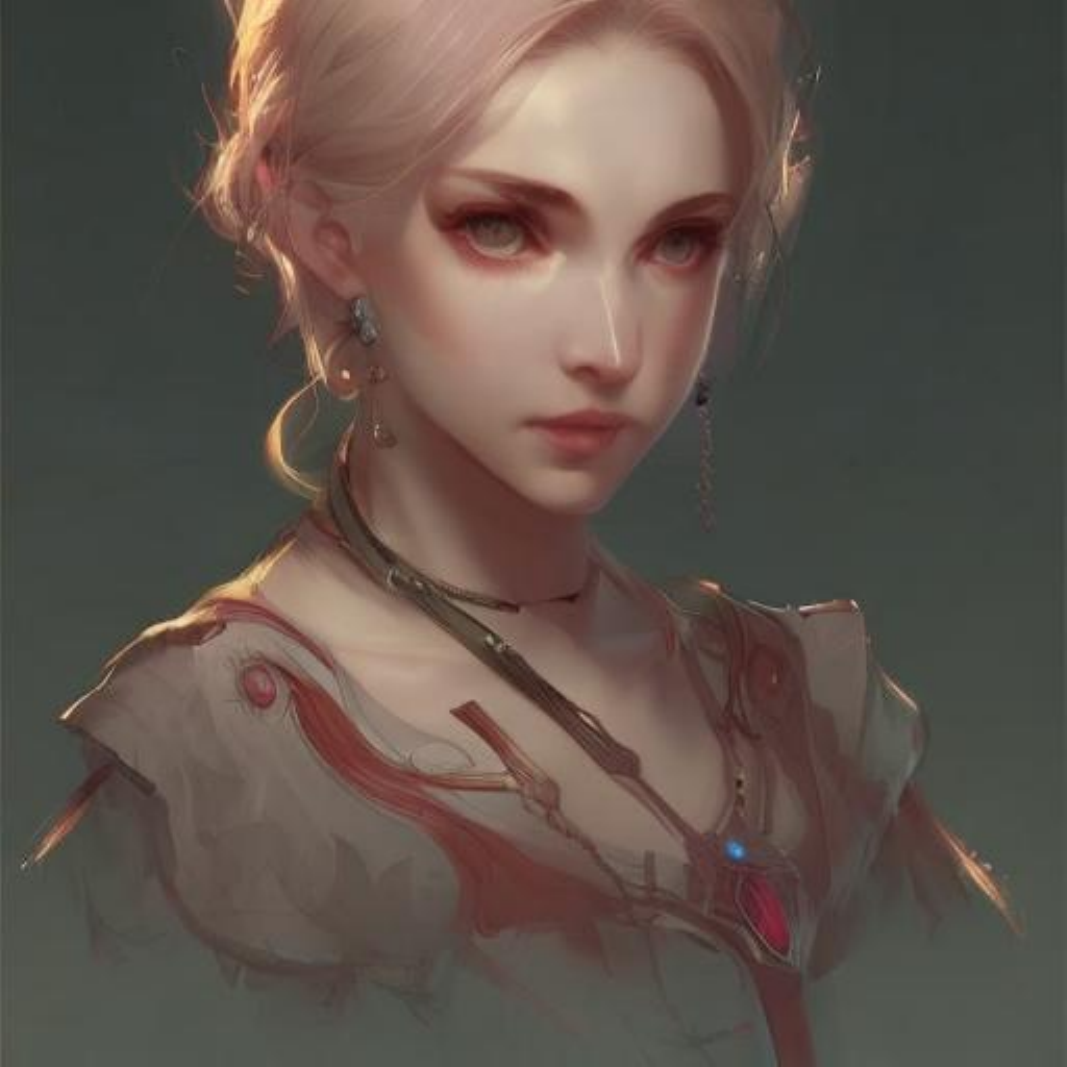} &
        \includegraphics[width=3cm]{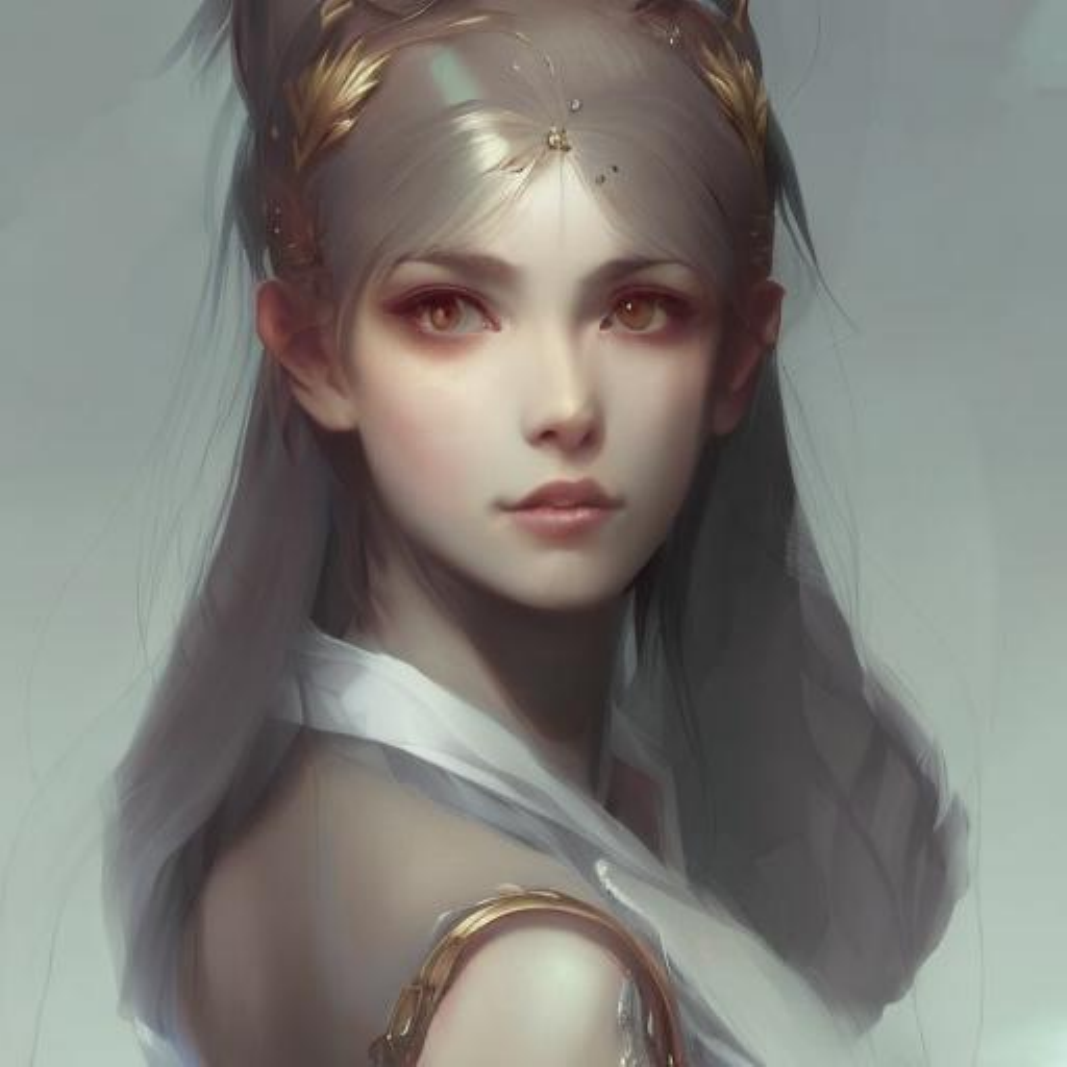} \\
    \midrule
     \multicolumn{5}{c}{Cat looking at beautiful colorful galaxy, high detail, digital art, beautiful , concept art,fantasy art, 4k.}\\
    \midrule
        \includegraphics[width=3cm]{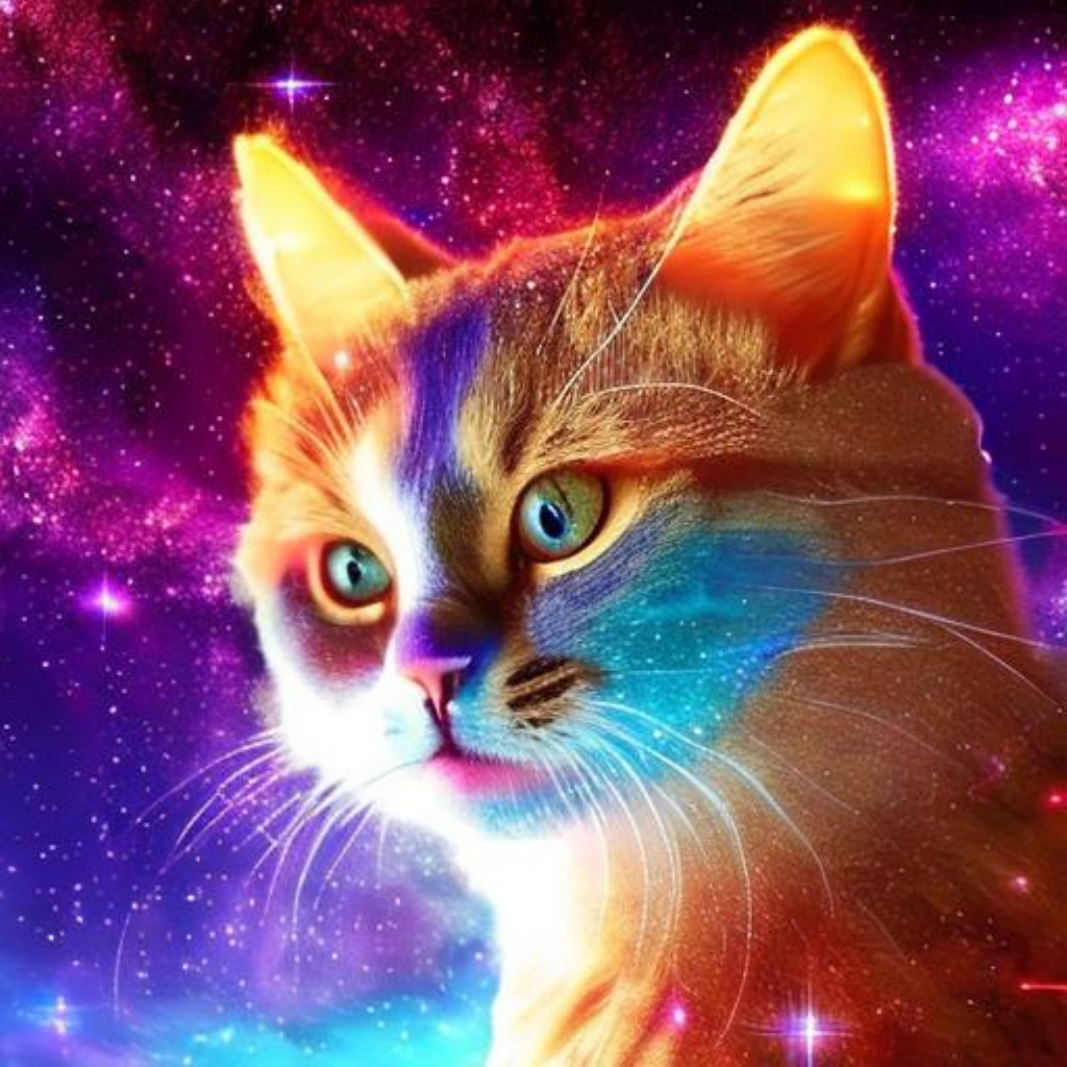} &
        \includegraphics[width=3cm]{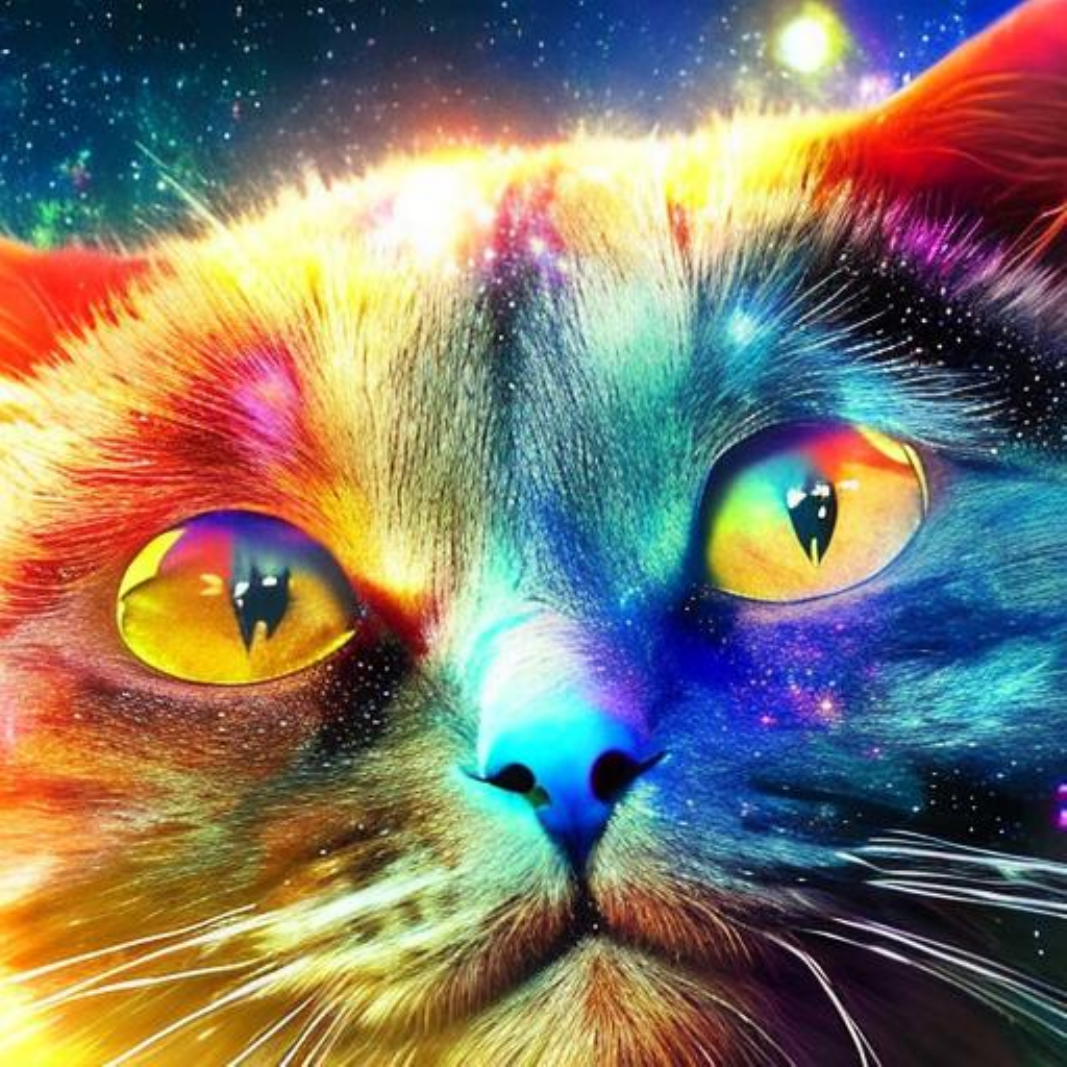} &
        \includegraphics[width=3cm]{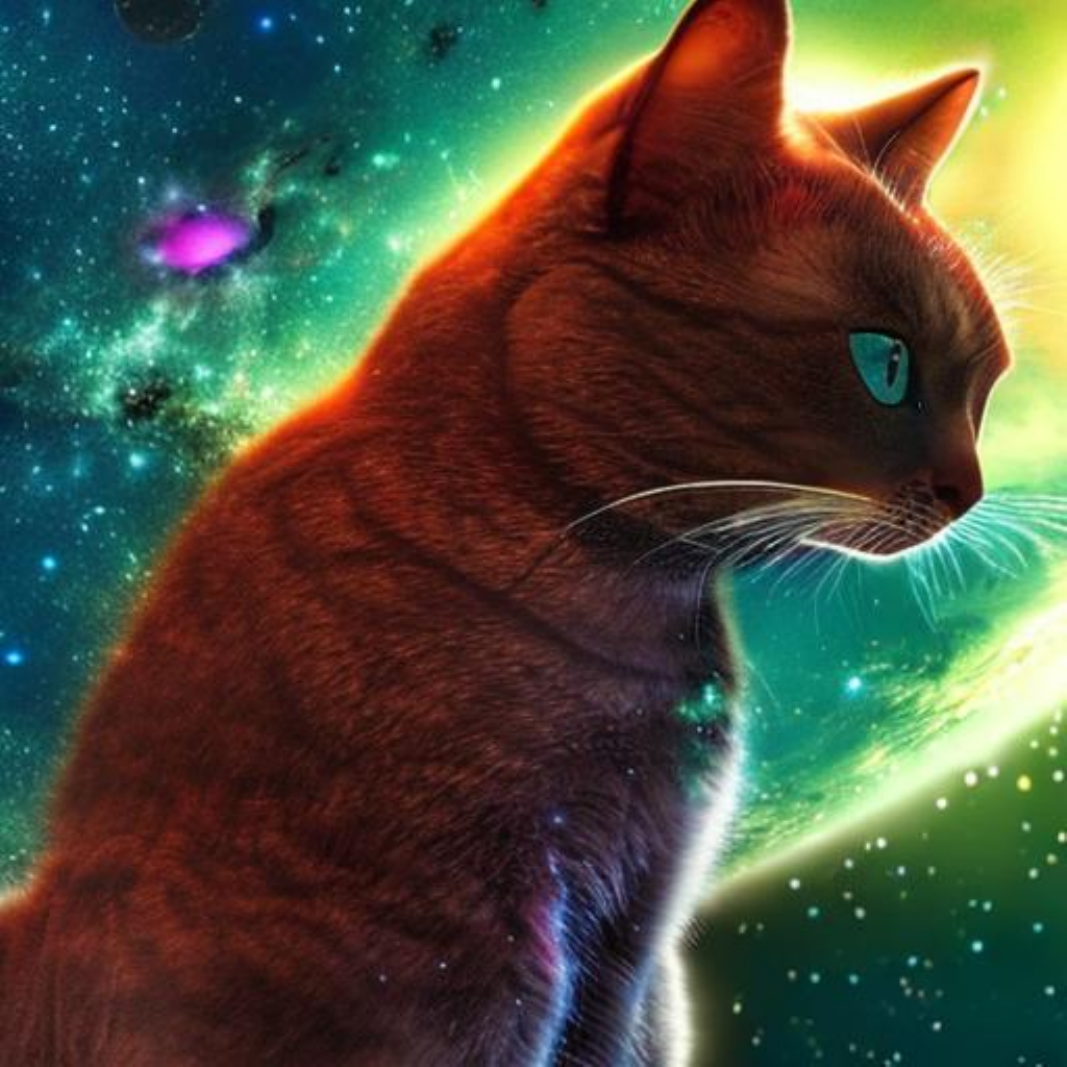} &
        \includegraphics[width=3cm]{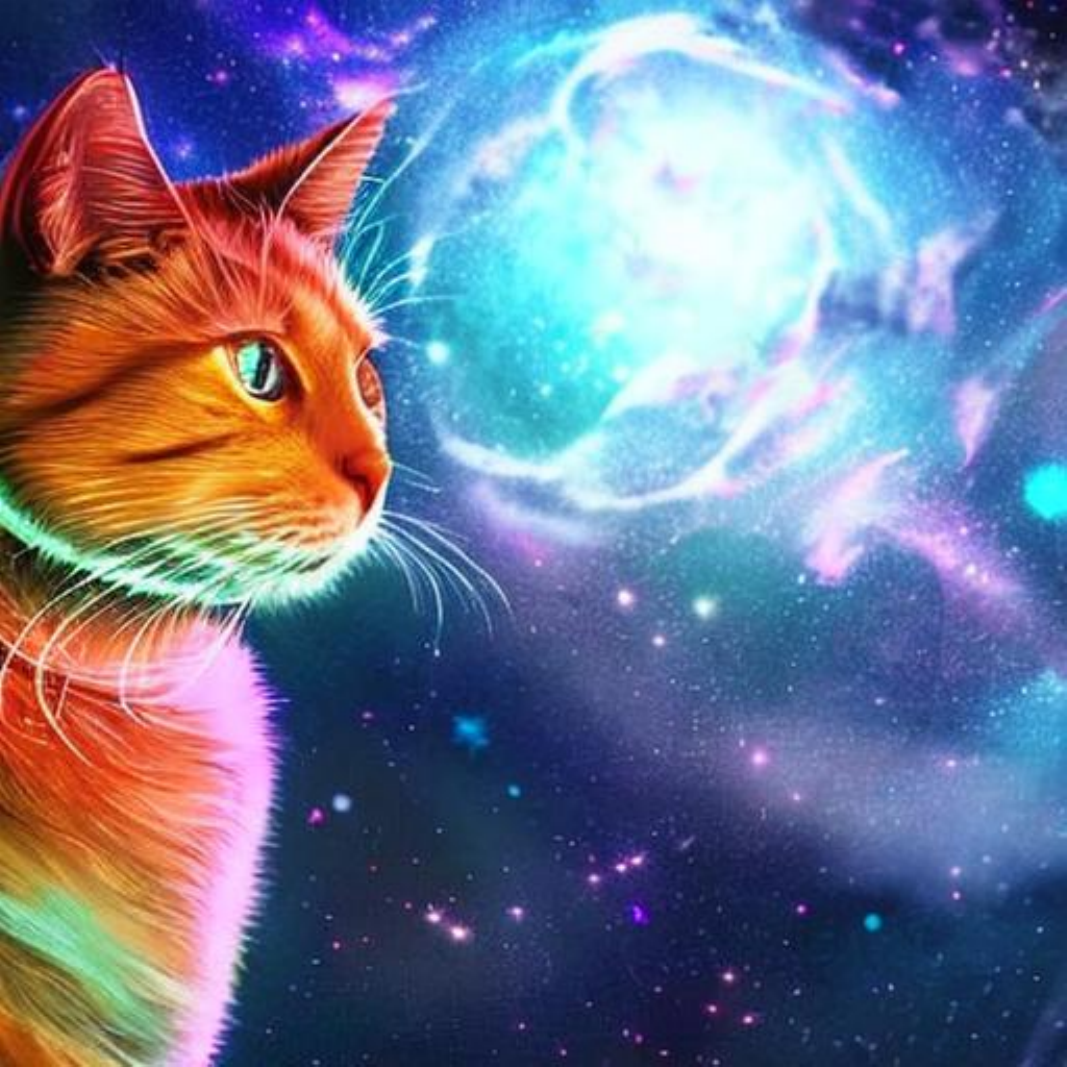} &
        \includegraphics[width=3cm]{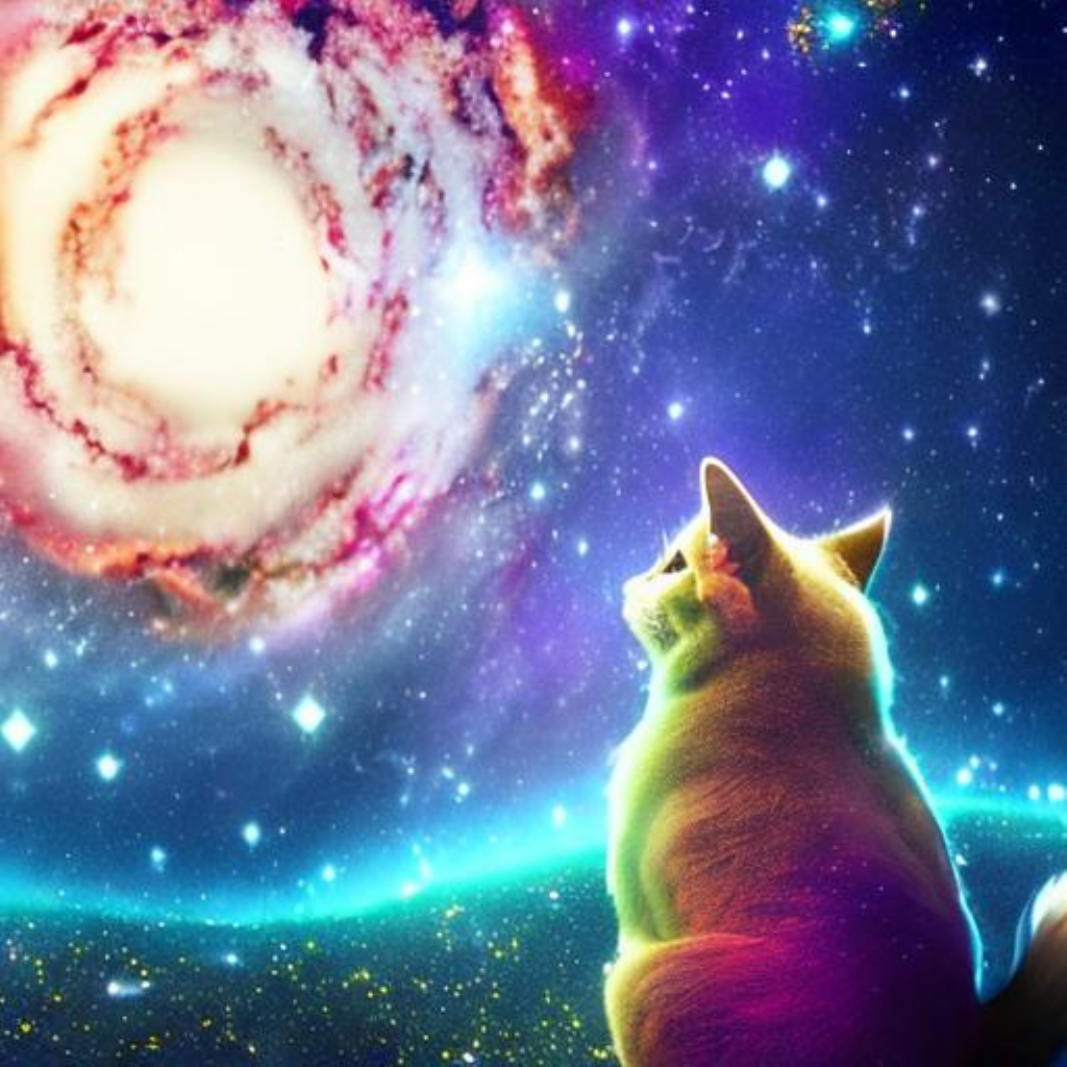} \\
    \bottomrule
    \end{tabular}
    \caption{Additional visual results of Gaussian Shading on generated images at resolution 512. We utilize five prompts in Stable-Diffusion-Prompt and generate images at five different embedding rates $l$, ranging from left to right as $l=1,2,3,4,5$.}
    \label{fig:quality_2}
\end{figure*}

To further showcase the visual performance of Gaussian Shading, we present the visual results at multiple embedding rates ranging from 1 to 5 on prompts from Stable-Diffusion-Prompt\footnote{\href{https://huggingface.co/datasets/Gustavosta/Stable-Diffusion-Prompts}{
Stable-Diffusion-Prompts}}. See \cref{fig:quality_2}, with the increase in watermark length, the model maintains a good generation quality. Moreover, the diversity and randomness of watermarked images indirectly reflect the performance-lossless characteristic of Gaussian Shading.

\newpage

\newpage

\newpage

\end{document}